\documentclass[10pt]{article} 
\usepackage[accepted]{tmlr}

\usepackage{amsmath,amsfonts,bm}

\def\eqref#1{equation~\ref{#1}}

\def\1{\bm{1}}

\DeclareMathAlphabet{\mathsfit}{\encodingdefault}{\sfdefault}{m}{sl}
\SetMathAlphabet{\mathsfit}{bold}{\encodingdefault}{\sfdefault}{bx}{n}

\usepackage{latexsym}
\usepackage{amsmath}
\usepackage{amssymb}
\usepackage{hyperref}
\usepackage{url}
\usepackage{graphicx}
\usepackage{bbding}
\usepackage{arydshln}
\usepackage{booktabs}
\usepackage{multirow}
\usepackage{subcaption}
\usepackage{enumitem}
\usepackage[T1]{fontenc}
\usepackage[utf8]{inputenc}
\usepackage{microtype}
\usepackage{bbding}
\usepackage{wrapfig}
\usepackage{soul}
\setlist{topsep=0pt, partopsep=0pt, parsep=0pt, itemsep=0pt}
\usepackage{soul}
\soulregister{\cite}{7}  
\soulregister{\citep}{7} 
\soulregister{\citet}{7} 
\soulregister{\citealp}{7} 
\soulregister{\S}{7} 
\soulregister{\footnote}{7}  
\soulregister{\ref}{7}   

\usepackage{csquotes}
\usepackage{array}
\newcommand{\PreserveBackslash}[1]{\let\temp=\\#1\let\\=\temp}
\newcolumntype{C}[1]{>{\PreserveBackslash\centering}p{#1}}
\newcolumntype{R}[1]{>{\PreserveBackslash\raggedleft}p{#1}}
\newcolumntype{L}[1]{>{\PreserveBackslash\raggedright}p{#1}}
\usepackage{xcolor}         
\definecolor{ao}{rgb}{0.0, 0.5, 0.0}
\definecolor{forestgreen}{RGB}{0, 150, 0}

\newcommand{\jl}[1]{\textcolor{ao}{\textbf{Ji-Ung:} #1}}

\title{B-cos LM: Efficiently Transforming Pre-trained Language Models for Improved Explainability}

\author{\name Yifan Wang \email yifwang@lst.uni-saarland.de \\
      \addr Saarland University, Saarbrücken, Germany
      \AND
      \name Sukrut Rao \email sukrut.rao@mpi-inf.mpg.de \\
      \addr Max Planck Institute for Informatics, Saarland Informatics Campus, Saarbrücken, Germany 
      \AND
      \name Ji-Ung Lee \email ji-ung.lee@uni-saarland.de\\
      \addr Saarland University, Saarbrücken, Germany 
      \AND
      \name Mayank Jobanputra \email mayank@lst.uni-saarland.de \\
      \addr Saarland University, Saarbrücken, Germany 
      \AND
      \name Vera Demberg \email vera@lst.uni-saarland.de \\
      \addr Saarland University, Saarbrücken, Germany \\
      Max Planck Institute for Informatics, Saarland Informatics Campus, Saarbrücken, Germany 
      }

\def\month{12}  
\def\year{2025} 
\def\openreview{\url{https://openreview.net/forum?id=c180UH8Dg8}} 

\begin{document}

\maketitle

\begin{abstract}
Post-hoc explanation methods for black-box models often struggle with faithfulness and human interpretability due to the lack of explainability in current neural architectures. 
Meanwhile, B-cos networks have been introduced to improve model explainability by proposing an architecture that removes bias terms and promotes input-weight alignment.
Although B-cos networks have shown success in building explainable systems, their application has so far been limited to computer vision models and their associated training pipelines.
In this work, we introduce B-cos LMs, i.e., B-cos Language Models (LMs) empowered for natural language processing (NLP) tasks. 
Our approach directly transforms pre-trained language models into B-cos LMs by combining B-cos conversion and task fine-tuning, improving efficiency compared to previous methods.
Automatic and human evaluation results demonstrate that B-cos LMs produce more faithful and human interpretable explanations than post-hoc methods, while maintaining task performance comparable to conventional fine-tuning.
Our in-depth analysis explores how B-cos LMs differ from conventionally fine-tuned models in their learning processes and explanation patterns. 
Finally, we present a first exploration of transforming decoder-only models to B-cos LMs for generation tasks. 
Our code is available at \href{https://github.com/Ewanwong/bcos_lm}{https://github.com/Ewanwong/bcos\_lm}.
\end{abstract}


\section{Introduction}

Pre-trained language models (PLMs) such as BERT~\citep{Devlin-2019-BERT} and GPT~\citep{Radford-2019-GPT, NEURIPS2020_1457c0d6, DBLP:journals/corr/abs-2303-08774} have significantly advanced performance across a plethora of NLP tasks~\citep{wang-etal-2018-glue, Eval-2023-Harness}. 
However, their complex architectures and black-box nature make understanding their behavior a persistent challenge~\citep{Bommasani-2021-FoundationModels}. 
To address this, research has increasingly focused on understanding model predictions in various natural language understanding and generation tasks using different forms of explanations, such as input-based explanations~\citep{zijian-etal-2024-unveiling, wei-jie-etal-2024-plausible, jiang-etal-2024-mare, madsen-etal-2024-self, yin-neubig-2022-interpreting, deiseroth-etal-2023-atman, abbasi-etal-2025-normxlogit}, natural language explanations~\citep{sahana-etal-2024-tailoring, wang-etal-2025-cross}, and concept-based explanations~\citep{yu-etal-2024-latent, raman-etal-2024-understanding}. Among others, input-based explanations, often referred to as rationales, aim to reveal how specific inputs influence a model's prediction~\citep{arras-etal-2019-evaluating, atanasova-etal-2020-diagnostic, lyu-etal-2024-towards}. In this work, we focus on input-based explanations, as they offer the most direct insight into model behavior and are often mandated by laws, such as the EU Artificial Intelligence Act.


\begin{figure}
    \centering
    \includegraphics[width=0.7\linewidth]{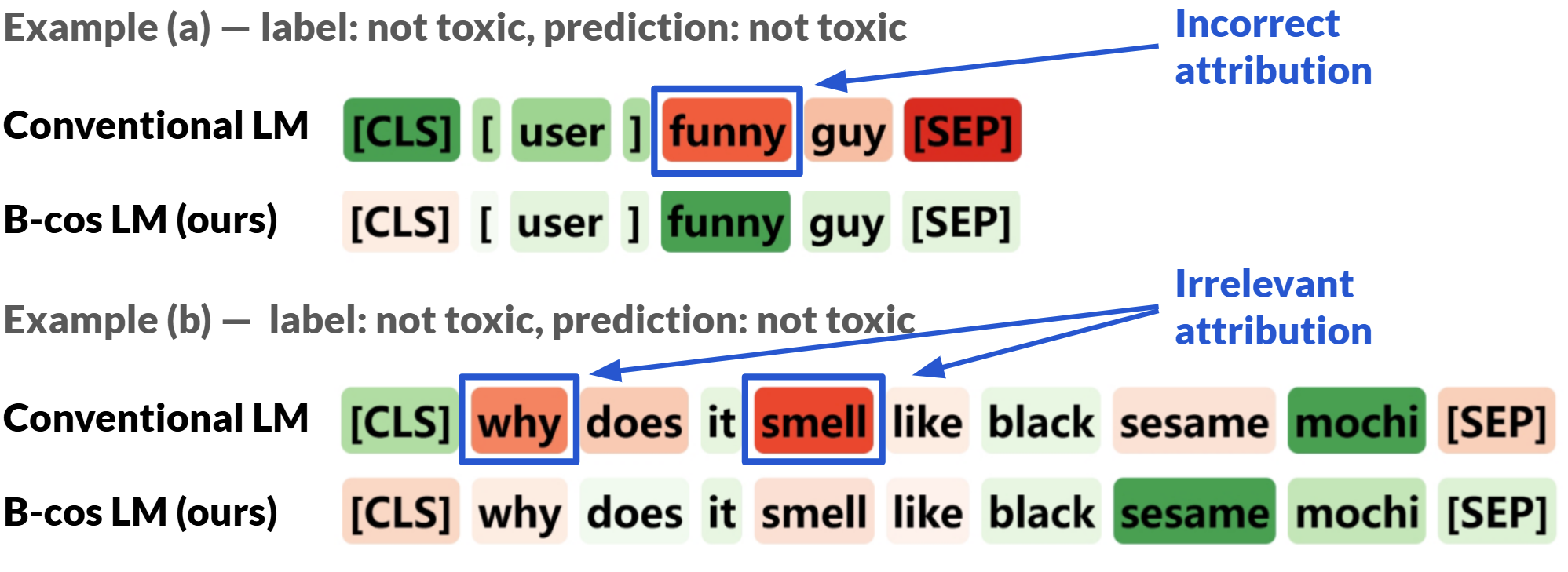}
    \caption{\textbf{Visualization of $\mathbf{W(x)x}$ in a conventionally fine-tuned model (Conventional LM) and a B-cos LM.} \textcolor{forestgreen}{Green} (\textcolor{red}{red}) indicates the \textcolor{forestgreen}{positive} (\textcolor{red}{negative}) impact of tokens on the prediction. In both examples, both models correctly predict \textit{not toxic}. In the Conventional LM, ``funny'' is \textcolor{blue}{incorrectly} assigned a negative attribution in example (a), and in example (b), \textcolor{blue}{irrelevant} words like ``why'' and ``smell'' are highlighted, making the explanations unfaithful and less interpretable. Examples and explanations are drawn from HateXplain. See ~\S\ref{sec:methodology} for details on how $\mathbf{W(x)x}$ is computed.}
    \label{fig:bcos_vs_baseline}
\end{figure}

Most input-based explanation methods for neural models are post-hoc, meaning that they attempt to explain a model's behavior only after it has been trained and deployed~\citep{lyu-etal-2024-towards}. 
While these methods are widely used and easy to apply, they have been shown to produce unfaithful explanations that do not accurately reflect the model's actual reasoning process~\citep{Kindermans-2019-Reliability, Slack-2020-FoolingLimeSHAP, pruthi-etal-2020-learning, ye-etal-2025-input}.
They also struggle with human interpretability, making it difficult for users to understand the model's reasoning~\citep{Smilkov-2017-SmoothGrad, NEURIPS2021_e0cd3f16}. 
Prior research suggests that these limitations stem from a lack of inherent explainability in current models, that is, the model's ability to generate faithful and interpretable explanations by design~\citep{Kindermans-2018-PatternNet, NEURIPS2018_3e9f0fc9, rudin2019stop}.
As a result, improving model explainability is crucial for producing explanations that are both reliable and useful to users.\footnote{Considering the evolving definition of these terms in past literature, we provide a detailed definition in Appendix~\ref{appendix:definitions}.}
Figure \ref{fig:bcos_vs_baseline} provides examples illustrating this issue.

To overcome these limitations, we introduce \textbf{B-cos LM}, a dynamic linear model that learns the most task-relevant patterns through increased input-weight alignment pressure. 
Building upon B-cos networks that were first introduced by~\citet{Bohle_2022_CVPR} for computer vision, we ensure the explainability of B-cos LMs through mathematically grounded architectural and computational adaptations, with specialized architectural modifications and training pipelines tailored for NLP tasks.

We conduct comprehensive empirical experiments using encoder-only models on classification tasks. Our focus on classification is motivated by its prevalence in high-stakes applications, such as loan approvals, hiring decisions, and hate speech detection, where explainability is crucial and often legally mandated. Encoder-only models have also seen renewed interest in the research community~\citep{modernbert, neobert, Reason-ModernColBERT}, and they remain the standard architecture for text classification tasks and continue to perform competitively compared to large language models (LLMs)~\citep{hang-etal-2024-advancing}.
Beyond that, we also explore applying B-cos LMs to decoder-only models for generation tasks and show that B-cos LMs can be extended to a variety of tasks and the latest model architectures.
Our contributions are as follows:

\begin{enumerate}
    \item We propose B-cos LM, a novel model with enhanced explainability. 
    Automatic and human evaluations demonstrate that B-cos LMs generate more faithful and human interpretable explanations than post-hoc explanations while maintaining a strong task performance.
    \item We investigate different strategies for transforming PLMs into task-specific B-cos LMs. 
    Our findings show that combining task fine-tuning and B-cos conversion is the most efficient approach, leading to faster convergence than previous B-cos methods and conventional fine-tuning.
    \item We thoroughly investigate how B-cos LMs differ from conventionally fine-tuned models and examine how alignment pressure influences their behavior.
    \item We are also the first to explore the transformation of decoder-only models to B-cos LMs for generation tasks, providing a step towards a broader application of B-cos LMs in the era of LLMs.
\end{enumerate}

\section{Related Work}

\paragraph{Post-hoc Explanation Methods} 
Various methods have been proposed to provide post-hoc explanations for neural model predictions~\citep{atanasova-etal-2020-diagnostic}. 
These methods can be broadly categorized based on how they generate explanations: gradient-based~\citep{DBLP:journals/corr/SimonyanVZ13, DBLP:journals/corr/KindermansSMD16, Sundararajan-2017-IntegratedGrad, enguehard-2023-sequential}, propagation-based~\citep{bach2015pixel, shrikumar2017learning, DBLP:journals/corr/SpringenbergDBR14, ferrando-etal-2023-explaining, modarressi-etal-2022-globenc, modarressi-etal-2023-decompx}, and perturbation-based methods~\citep{DBLP:journals/corr/LiMJ16a, Ribeiro-2016-LIME, Lundberg-2017-SHAP, deiseroth-etal-2023-atman}. 
Besides, the attention mechanism~\citep{DBLP:journals/corr/BahdanauCB14} is often viewed as an explanation, particularly in transformer-based models~\citep{DBLP:conf/nips/VaswaniSPUJGKP17}. 
While most existing work focuses on understanding model predictions in classification settings, recent efforts have also aimed to explain model behavior in generation tasks, including sentence completion~\citep{yin-neubig-2022-interpreting, ferrando-etal-2023-explaining}, question answering~\citep{enouen-etal-2024-textgenshap}, and summarization~\citep{cohen-etal-2024-contextcite}.

Although post-hoc methods have been widely used, numerous studies have shown that they lack faithfulness, often failing to capture the true decision-making process of the model~\citep{Kindermans-2019-Reliability, jain-wallace-2019-attention, Slack-2020-FoolingLimeSHAP, pruthi-etal-2020-learning, ye-etal-2025-input}. 
Furthermore, they are noisy and may select irrelevant information, leading to explanations that cannot be interpreted by humans~\citep{Smilkov-2017-SmoothGrad, NEURIPS2021_e0cd3f16}.

\paragraph{From Post-hoc Explanations to Explainable Models}

Prior research suggests that the lack of faithfulness and human interpretability in post-hoc explanations arises from the fundamental lack of explainability in modern neural models, which are typically optimized solely for task performance~\citep{Kindermans-2018-PatternNet, rudin2019stop, atanasova2022diagnostics}.
In response, various efforts have been made to enhance model explainability. 
Some works have introduced constraints that improve specific explanation properties, such as faithfulness~\citep{tutek2022toward, moradi-etal-2020-training, moradi-etal-2021-measuring, barkan-etal-2024-llm}, consistency~\citep{atanasova2022diagnostics}, locality~\citep{NEURIPS2018_3e9f0fc9}, and plausibility~\citep{NEURIPS2021_e0cd3f16}. 
However, as these constraints are typically imposed as regularizers, their effectiveness in improving explanation quality is not guaranteed~\citep{pruthi-etal-2020-learning}. 
Others have proposed self-explanatory model architectures such as rationale-based models that utilize an ``explain-then-predict'' pipeline, where one module selects rationales for another to make predictions based on them~\citep{lei-etal-2016-rationalizing}.
Although seemingly transparent, both components rely on neural networks, making the rationale extraction and utilization processes opaque~\citep{zheng-etal-2022-irrationality, jacovi-goldberg-2021-aligning}. 
Besides, such models may face optimization challenges that limit their practicality in real-world tasks~\citep{lyu-etal-2024-towards}.
\looseness=-1


To tackle these shortcomings, \citet{Bohle_2022_CVPR} proposed B-cos networks. 
Unlike methods that impose external constraints, B-cos networks improve explainability through mathematically grounded architectural and computational adaptations. 
Moreover, these adaptations are designed as drop-in replacements for conventional model components, making B-cos networks easy to train with minimal performance loss. Most recently, \citet{arya24bcosification} explored B-cosification techniques to convert existing models into B-cos models, which reduces the training costs of adopting B-cos architectures. 

Despite their successful application in vision tasks, B-cos networks have yet to be explored in NLP, where input modalities and training paradigms differ significantly. 
In this work, we adapt B-cos models for the language domain, integrating them efficiently into NLP pipelines. 


\section{Methodology}
\label{sec:methodology}
In this section, we outline the architecture and training process of B-cos LMs and how their design ensures faithful and human interpretable explanations.
We first introduce B-cos networks (\S~\ref{bcos background}) and then describe how we transform PLMs to task-specific B-cos LMs (\S~\ref{b-cosification in nlp}). 
Finally, we demonstrate how to generate explanations from B-cos LMs (\S~\ref{generate explanation}).
Notations used in the work are detailed in Appendix~\ref{appendix:notation}.
\looseness=-1

\subsection{B-cos Networks}
\label{bcos background}

Complex neural networks can be interpreted as generalized linear models~\citep{DBLP:conf/icml/NairH10, NEURIPS2018_3e9f0fc9, NEURIPS2019_80537a94}. 
For each input $\mathbf{x}$, the network effectively applies a linear transformation: $\mathbf{f(x) = W(x)x + b(x)}$, where both the weight $\mathbf{W(x)}$ and bias $\mathbf{b(x)}$ depend on $\mathbf{x}$. 
Given that many activation functions are (approximately) piecewise linear, the overall network can be viewed as (approximately) piecewise affine~\citep{NEURIPS2018_3e9f0fc9}. 
Earlier work refers to such models as dynamic linear models~\citep{bohle2021convolutional, Bohle_2022_CVPR}, highlighting the fact that the weight and bias terms dynamically change according to $\mathbf{x}$. 

Under this dynamic linear perspective, the linear mapping $\mathbf{W(x)}$ can be seen as attributing model predictions to individual input features, and $\mathbf{W(x_i)x_i}$ can be seen as the contribution of feature $\mathbf{x_i}$ to the model prediction.
However, two challenges hinder the direct use of this interpretation. 
First, $\mathbf{W(x)}$ alone provides an incomplete and unfaithful model summary, since $\mathbf{f(x) \neq W(x)x}$ due to the presence of the bias term $\mathbf{b(x)}$, and incorporating $\mathbf{b(x)}$ into explanations is highly non-trivial~\citep{pmlr-v97-wang19p}. 
Second, $\mathbf{W(x_i)x_i}$ is often difficult for humans to interpret, as $\mathbf{W(x)}$ does not necessarily align only with task-relevant input patterns~\citep{Smilkov-2017-SmoothGrad} and therefore yields noisy and irrelevant explanations.
Figure~\ref{fig:bcos_vs_baseline} illustrates these challenges. 
To address these issues, \citet{Bohle_2022_CVPR} introduced B-cos networks by replacing the conventional linear transformation:
\begin{equation}
\mathbf{f(x;w, \text{b}) = w^T x+\text{b} = \|w\|\|x\|\text{cos}(x,w)+\text{b}}
\end{equation}
with a B-cos transformation: 
\begin{align}
\text{B-cos}&\mathbf{(x;w)}=\mathbf{\hat{w}^T x} \times |\text{cos}\mathbf{(x, \hat{w})|^\text{B-1}} \label{eq:b-cos}\\
&=\|\mathbf{\hat{w}\|\|x\|}|\text{cos}\mathbf{(x,\hat{w})|^\text{B}}\nonumber \times \mathbf{\text{sgn}(\text{cos}(x, \hat{w}))} \nonumber
\end{align}
where $\mathbf{\hat{w}}$ is a scaled version of $\mathbf{w}$ with unit norm and $\text{sgn}$ denotes the sign function.

B-cos$\mathbf{(x;w)}$ can be seen as a linear transformation of $\mathbf{x}$ with the dynamic linear weight $\mathbf{w(x)=|\text{cos}(x,\hat{w})|^\text{B-1} \times \hat{w}}$. 
The absence of $\mathbf{b(x)}$ ensures the completeness of summary $\mathbf{w(x)}$.
We show that all components of transformer models can be viewed as or easily converted to bias-free, dynamic linear modules in Appendix~\ref{sec:dynamic linear components}, and demonstrate that this completeness extends to an entire network composed of such modules in \S~\ref{generate explanation}.
Moreover, since the B-cos module output is bounded by $\mathbf{\|x\|}$, the weight $\mathbf{w}$ must align closely with task-relevant patterns to achieve a high cosine similarity and strong activation, especially under additional alignment pressure (B>1). 
This drives the model to assign greater weight to the most relevant features when optimizing target output probabilities, promoting the learning of representative patterns during training.
Consequently, during explanation generation, task-relevant features $\mathbf{x_i}$ receive higher attribution $\mathbf{W(x_i)x_i}$ due to stronger alignment, while irrelevant features receive lower attribution, suppressed by weaker alignment and the exponential scaling.
For a more detailed discussion of how the B-cos transformation enhances faithfulness and human interpretability, please see~\citet{Bohle_2022_CVPR, bohle2024b}.

While early B-cos models were trained from scratch, \citet{arya24bcosification} recently introduced B-cosification, an efficient method to obtain B-cos models. 
This approach first modifies conventional models with task capacities to adopt the B-cos architecture, followed by fine-tuning on downstream datasets for B-cos conversion. 
B-cosified models generate explanations as faithful and interpretable as B-cos models trained from scratch but at a much lower training cost. 
However, directly applying B-cosification to LMs is non-trivial and inefficient due to the significant differences in model architectures and training pipelines.


\subsection{B-cosification for LMs}
\label{b-cosification in nlp}
In this section, we present our B-cosification approach for LMs.
We summarize the differences between B-cosification for LMs, its counterpart for vision models, and conventional fine-tuning in Table \ref{tab:comparison}. 
\begin{table*}[h]
    \centering
    \resizebox{0.9\linewidth}{!}{
    \tabcolsep=2pt
    \begin{tabular}{lC{5cm}C{7.5cm}C{6cm}}
        \toprule
        \textbf{Property} & \textbf{Conventional Fine-tuning} & \textbf{B-cosification for vision}~\citep{arya24bcosification} & \textbf{B-cos LM (ours)} \\
        \midrule
        \textbf{Bias terms} & yes & no & no \\
        \textbf{B (alignment pressure)} & 1 & 2 & 1.25 / 1.5 \\
        \textbf{Pred. Head Activations} & tanh  & n/a\footnotemark & identity \\
        \midrule
       \textbf{Prior task abilities} & no & yes  & no  \\
        \textbf{Training objectives} & Task fine-tuning & B-cos conversion & Task fine-tuning \& B-cos conversion \\
        \bottomrule
    \end{tabular}
    }
    \caption{Comparison between conventional fine-tuning, B-cosification for vision models and B-cosification for language models (B-cos LM). 
    Conventional fine-tuning and B-cosification for vision follow the configuration of BERT for sequence classification and CLIP~\citep{radford2021learning}, respectively (cf. \S~\ref{sec:methodology} for details).}
    \label{tab:comparison}
\end{table*}

\subsubsection{B-cos Adaptations}
\label{b-cosification-architecture}
Given a conventional model, we first modify its architecture and computation to integrate the B-cos framework.
\footnotetext{\citet{arya24bcosification} used a single linear layer on top of CLIP so the prediction head activation is not applicable in their setup.}
\paragraph{Architectural Adaptations} 
For completeness and faithfulness of explanations, we follow \citet{arya24bcosification} and remove all bias terms in models, including those in the affine transformations of layer normalization and attention blocks. 
Additionally, a prediction head is typically added on top of transformers before fine-tuning for downstream tasks in the NLP pipeline. This head often includes activation functions that are not (approximately) piecewise linear, such as sigmoid and tanh.
To accommodate the unique architecture of LMs, we remove all activation functions in the prediction heads, as they generate explanations that are not locally difference-bounded~\citep{NEURIPS2018_3e9f0fc9} and introduce numerical instability during explanation generation. 
Our experiments show that the added non-linearity from B>1 could compensate for this removal. 

\paragraph{Introducing B-cos Computation} 
To promote input-weight alignment and improve human interpretability of explanations, we replace all linear transformations with B-cos transformations in \S~\ref{bcos background}. 
For a more efficient B-cosification, B-cos layers are initialized with the corresponding weights of the original model.

\subsubsection{Fine-tuning}
\label{b-cosification-finetuning}
The B-cos adaptations above modify the architecture and computation of models, requiring fine-tuning to restore their capabilities and adapt to alignment pressure.
Following the ``pre-train then fine-tune'' paradigm, which is frequently utilized in NLP tasks, we directly transform PLMs to B-cos LMs, rather than adapting task-specific models 
as done in previous work~\citep{arya24bcosification}. 
This fundamental difference in the training pipeline adds complexity to B-cosification for LMs, as the objective involves both B-cos conversion and task fine-tuning. 
While there are multiple ways to conjoin these two steps (cf. \S~\ref{sec: different setups}), we find that the most efficient way is to combine them by first applying B-cos adaptations to a PLM and then fine-tuning it on a downstream task. 
Following \citet{Bohle_2022_CVPR}, we use the binary cross-entropy (BCE) loss instead of the conventional cross-entropy loss, as it explicitly maximizes the absolute target logits and strengthens the alignment pressure.
We provide an extensive comparison of different B-cosification setups in \S~\ref{sec: different setups}. 



\subsection{Computing B-cos Explanations}
\label{generate explanation}
Once trained, the B-cos LM can generate explanations that faithfully summarize its decision-making process during inference. 
As all components are dynamic linear with no bias terms (cf. Appendix~\ref{sec:dynamic linear components}), the entire model computation can be expressed as a sequence of matrix multiplications, which can be completely summarized as a single dynamic linear function:
\begin{equation}
\mathbf{\hat{W}}_L(\mathbf{A}_L)\mathbf{\hat{W}}_{L-1}(\mathbf{A}_{L-1})...\mathbf{\hat{W}}_1(\mathbf{A}_1=\mathbf{X})\mathbf{X} = \Pi_{j=1}^{L}\mathbf{\hat{W}}_j(\mathbf{A}_j)
\end{equation}
Note that a residual connection of $\mathbf{W(x)x+x}$ with $\mathbf{x} \in \mathbb{R}^n$ and $\mathbf{W(x)} \in \mathbb{R}^{n \times n}$ is mathematically equivalent to a single dynamic linear transformation of $(\mathbf{W(x)+I}_n)\mathbf{x}$. 
Considering the textual inputs specific to LMs, we attribute the model's predictions to the embedding representations. Specifically, to quantify the contribution of a token $i$ to a model prediction, we compute the dot product $\mathbf{W(x}_i)\mathbf{x}_i$ between its embedding $\mathbf{x}_i$ and the corresponding dynamic linear weight $\mathbf{W(x}_i)$ for the target class logit. 
For the remainder of the paper, we will refer to such explanations as \textit{B-cos explanations}.


\section{Experiments}

We evaluate the task performance of B-cos LMs and faithfulness of B-cos explanations with automatic evaluation across various tasks and PLMs. In addition, we conduct a human evaluation study to compare the human interpretability of B-cos explanations.
\S\ref{sec:experimental-setup}--\ref{sec:human-evaluation} describe our automatic evaluation setup, results, as well as human evaluation study, respectively.
\S\ref{sec:qualitative-analysis} provides a qualitative analysis. 
Finally, we conduct an ablation study in \S\ref{sec:ablation-study}. More details on the experimental setup and baseline methods are provided in Appendix~\ref{appendix: implementation details} and a comparison of computational efficiency is provided in Appendix~\ref{sec:explanation-efficiency}.

\subsection{Experimental Setup}\label{sec:experimental-setup}

\paragraph{Datasets and Models} Our experiments use three datasets: AG News (topic classification, \citealp{NIPS2015_250cf8b5}), IMDB (sentiment analysis, \citealp{maas-etal-2011-learning}), and HateXplain (hate speech detection, \citealp{mathew2021HateXplain}). 
BERT \citep{Devlin-2019-BERT}, DistilBERT \citep{DBLP:journals/corr/abs-1910-01108}, and RoBERTa \citep{DBLP:journals/corr/abs-1907-11692} are used as the basis for conventional fine-tuning and for obtaining B-cos LMs. 
We set B=1.25 for IMDB and B=1.5 for AG News and HateXplain datasets.

\paragraph{Baselines} We compare B-cos explanations against a diverse set of post-hoc explanation methods: Attention \citep{DBLP:journals/corr/BahdanauCB14}, InputXGradient (IxG, \citealp{DBLP:journals/corr/KindermansSMD16}), Sequential Integrated Gradients (SIG, \citealp{enguehard-2023-sequential}), DecompX \citep{modarressi-etal-2023-decompx}, Shapley Value Sampling (ShapSampl, \citealp{strumbelj2010efficient}), and LIME \citep{Ribeiro-2016-LIME}. 
We also apply these methods to a model trained with Saloss~\citep{chrysostomou-aletras-2021-enjoy}, which introduces faithfulness regularization. This setup enables a direct comparison between B-cos LMs and models specifically optimized for explainability.
For embedding-level explanation methods, we aggregate attributions by summing across all embedding dimensions.

\paragraph{Faithfulness Metrics} For a more comprehensive evaluation, we employ two different methods to assess faithfulness. 
First, we report two perturbation-based metrics~\citep{deyoung-etal-2020-eraser}:
\begin{itemize}
    \item \textbf{Comprehensiveness} (Comp) measures the average drop in predicted class probability after masking out the top $k\%$ most important tokens in the explanation. A higher score indicates better faithfulness.
    \item \textbf{Sufficiency} (Suff) measures the average drop in predicted class probability after keeping only the top $k\%$ tokens. A lower score indicates better faithfulness.
\end{itemize}

To avoid arbitrary choices of $k$, we compute Comp and Suff for multiple values ($k=10, 20, ..., 90$) and summarize them using the \underline{A}rea \underline{O}ver the \underline{P}erturbation \underline{C}urve (AOPC, \citealp{deyoung-etal-2020-eraser}).

In addition, we introduce a new faithfulness metric called \underline{Seq}uence \underline{P}ointing \underline{G}ame (Seq\-PG), inspired by the grid pointing game in vision tasks~\citep{bohle2021convolutional}:
\begin{itemize}
    \item \textbf{Sequence Pointing Game} (SeqPG). We evaluate models on synthetic sequences composed of segments associated with different classes. To assess faithfulness, we measure the proportion of positive attribution assigned to the corresponding segment of each class and compute their average. A higher score indicates better faithfulness.
\end{itemize}

Compared to perturbation-based metrics, SeqPG does not rely on perturbations and thus avoids the potential distortions introduced by token masking. 
When constructing SeqPG examples, we truncate each segment to a fixed length and randomize segment order to control for length and position effects. 
We generate synthetic examples using correctly and most confidently classified test instances.
SeqPG can be seen as a standardized version of hybrid document evaluation~\citep{poerner-etal-2018-evaluating}. 
We provide an example of SeqPG in Figure~\ref{fig:seqpg} and more details in Appendix~\ref{appendix:seqpg example}.

\subsection{Automatic Evaluation Results}\label{sec:automatic-evaluation-results}

\paragraph{Task Performance}
\label{sec:task performance}

\begin{wrapfigure}{r}{0.48\linewidth}
    \centering
    \vspace{-5mm}
    \includegraphics[width=\linewidth]{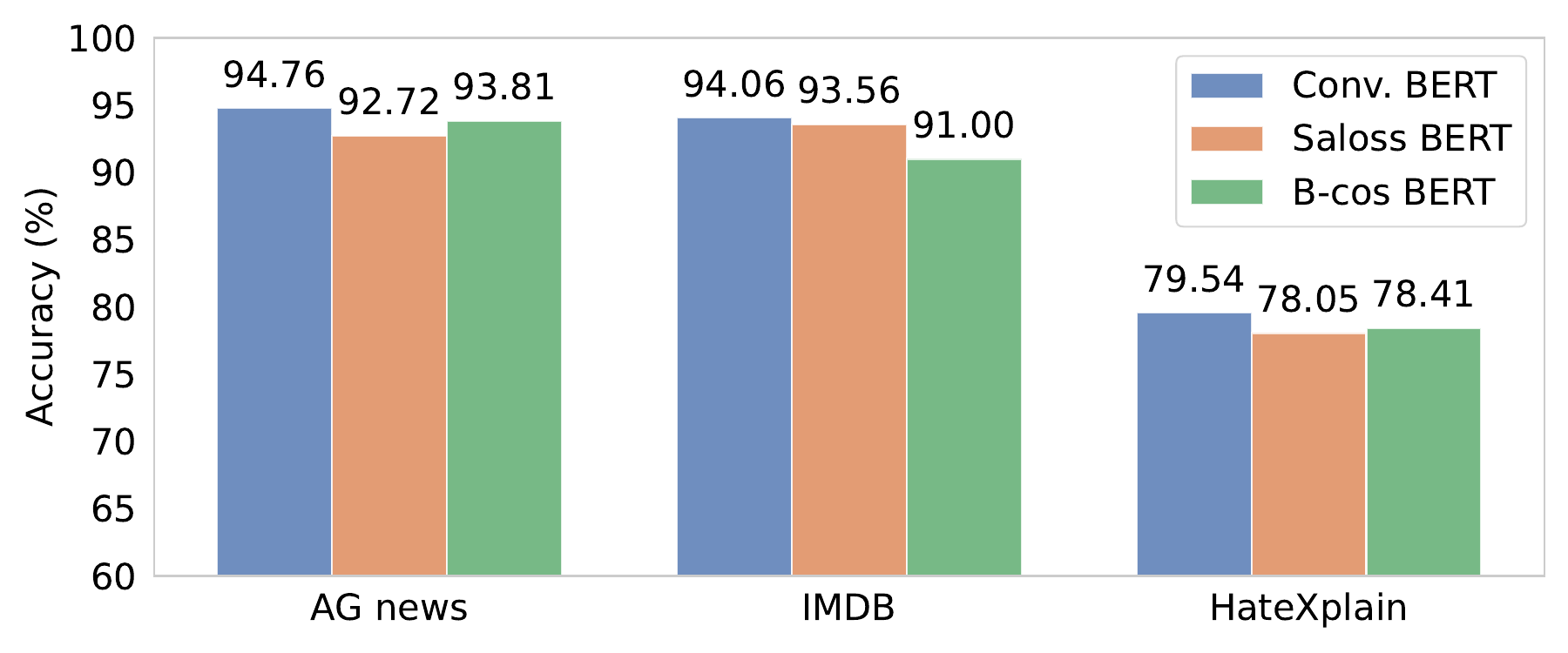}
    \caption{Mean accuracy of conventionally fine-tuned, Saloss and B-cos BERT averaged over three runs. 
    B-cos models perform comparably to conventional models on most tasks.}
    \label{fig:bert performance}
    \vspace{-3mm}
\end{wrapfigure}

Figure~\ref{fig:bert performance} shows the accuracy of conventionally fine-tuned, Saloss and B-cos BERT across three datasets (we provide results for DistilBERT and RoBERTa in Appendix~\ref{appendix:accuracy other plms}).
On AG News and HateXplain, B-cos LMs performs on par with conventional models, with only a minor drop ($\sim$1\%) in accuracy. They also outperform Saloss models on these datasets.
Only for IMDB, we find a slightly larger drop of 3.06\% compared to conventional BERT, though the performance remains strong overall. 

\paragraph{Faithfulness Results}

Table~\ref{tab:faithfulness} shows the faithfulness scores for post-hoc explanation methods on conventionally fine-tuned and Saloss BERT models, as well as B-cos explanations from B-cos BERT.
The results show that B-cos explanations are consistently and substantially more faithful than post-hoc methods across all models and datasets. 
On average, B-cos explanations outperform the strongest post-hoc methods on conventional models by 14.63 points in Comp and achieve negative Suff scores, indicating that the identified important tokens alone enable even more confident predictions. 
B-cos also shows significant gains in SeqPG.
While Saloss improves faithfulness for some post-hoc methods over conventional models, it still lags markedly behind B-cos LMs.
Similar trends are observed for other PLMs (Appendix~\ref{appendix:other PLMs}) as well. Although we do not include rationale-based models in the main experiments because they typically require additional supervision, a supplementary comparison in Appendix~\ref{appendix:rationale-based} shows that B-cos BERT still outperforms a rationale-based model on HateXplain.

\begin{table*}[ht]
  \centering
  \resizebox{\textwidth}{!}{
  \begin{tabular}{llccccccccc}
     \toprule
      \multirow{2}{*}{\textbf{Model}}&\multirow{2}{*}{\textbf{Method}} & \multicolumn{3}{c}{\textbf{AG News}} & \multicolumn{3}{c}{\textbf{IMDB}} & \multicolumn{3}{c}{\textbf{HateXplain}} \\
      \cmidrule(lr){3-5}\cmidrule(lr){6-8} \cmidrule(lr){9-11}
      & & Comp ($\uparrow$) & Suff ($\downarrow$) & SeqPG ($\uparrow$) & Comp ($\uparrow$) & Suff ($\downarrow$) & SeqPG ($\uparrow$) & Comp ($\uparrow$) & Suff ($\downarrow$) & SeqPG ($\uparrow$) \\
      \midrule

      \multirow{6}{*}{Conv. BERT}&Attention &24.40 & 8.09 & 50 & 26.84 & 14.56 & 50 & 27.64 & 13.83 & 50 \\
      &IxG & 15.28 & 10.19 & 45.41 & 18.29 & 16.96 & 49.42 & 19.16 & 18.90
      & 47.24 \\
      &SIG & 27.02 & 3.40 & 64.77 & 29.34 & 14.05 & 59.09 & 37.31 & 5.10 & 66.38 \\
      &DecompX & 52.16 & 0.92 & 84.48 & 57.94 & 2.41 & 63.27 & 44.86 & 2.72 & 66.76 \\
      &ShapSampl & 43.96 & 0.46 & 82.87 & 58.29 & 2.44 & \textbf{71.29} & 44.86 & 2.43 & 67.17 \\
      &LIME & 44.95 & 0.06 & 80.28 & 51.45 & 6.07 & 60.15 & 22.64 & 14.30 & 57.61 \\
    \midrule
      \multirow{6}{*}{Saloss BERT}&Attention & 34.73 & 3.65 & 50 & 27.59 & 13.64 & 50 & 34.95 & 26.26 & 50 \\
      &IxG & 14.98 & 12.66 & 51.01 & 24.19 & 16.30 & 49.02 & 26.61 & 30.94 & 50.74 \\
      &SIG & 16.70 & 8.22 & 63.74 & 45.44 & 8.48 & 54.96 & 44.53 & 21.50 & 54.70 \\
      &DecompX & 59.37 & 0.30 & 75.34 & 59.42 & 5.38 & 62.02 & 58.71 & 13.23 & 65.17 \\
      &ShapSampl & 37.73 & 0.77 & 73.96 & 65.38 & 3.17 & 70.23 & 57.05 & 15.10 & 72.36 \\
      &LIME & 53.18 & 2.37 & 76.16 & 53.31 & 6.32 & 58.65 & 21.73 & 21.96 & 55.71 \\

      \midrule
B-cos BERT & B-cos & \textbf{64.22} & \textbf{-1.26} & \textbf{87.92} & \textbf{74.18} & \textbf{-2.87} & 
      70.43 & \textbf{59.66} & \textbf{-4.89} & \textbf{77.57} \\
    \bottomrule
  \end{tabular}
  }
  \caption{Faithfulness evaluation for conventionally fine-tuned, Saloss, and B-cos BERT across three datasets. 
  The best results are in \textbf{bold}. 
B-cos explanations are consistently more faithful than post-hoc explanations from both baseline models.}
  \label{tab:faithfulness}
\end{table*}

\subsection{Human Evaluation}\label{sec:human-evaluation}

Contrary to previous B-cos studies that rely solely on automatic evaluations to assess explanations, we conduct the first human study to better evaluate the human interpretability and agreement of B-cos explanations. We compare them against three strong post-hoc explanation methods on conventional BERT models.

Following the practice in~\citet{enguehard-2023-sequential} and~\citet{dare}, we randomly select 50 instances, respectively, from AG News and HateXplain where the B-cos and conventional models make the same prediction.
Five annotators then rate the explanations in terms of human interpretability (how well they understand them) and human agreement (how much they agree with them) on a scale of 1-5. 
Further details on the evaluation criteria, rating scales, annotator instructions, and example annotations can be found in Appendix~\ref{appendix:human evaluation}.
\begin{wrapfigure}{r}{0.45\textwidth}  
  \vspace{-0mm}                        
  \centering
  \includegraphics[width=\linewidth]{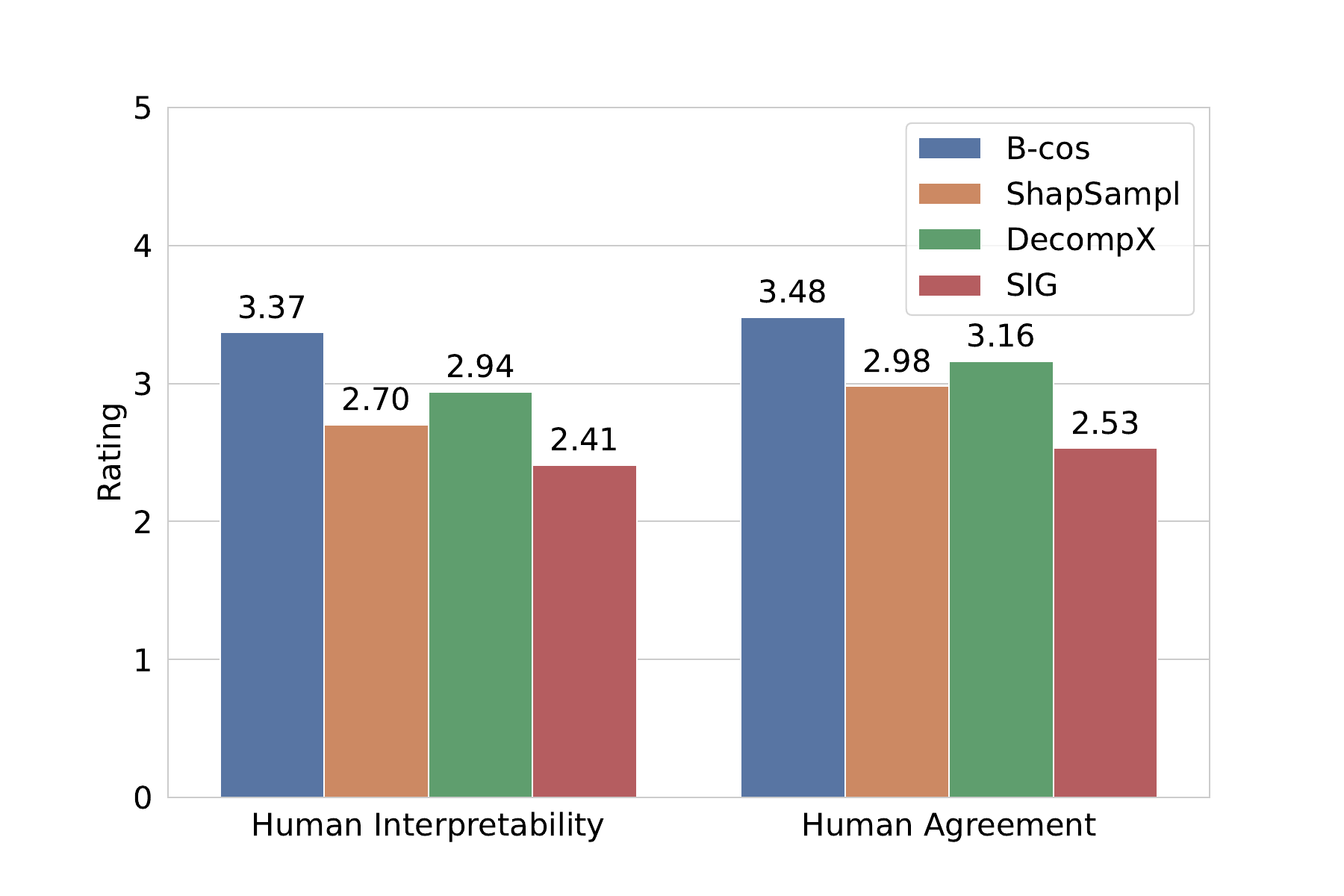}
  \caption{Human evaluation reveals that B-cos explanations have better human interpretability and human agreement than baseline methods.}
  \label{fig:human}
  \vspace{-15mm}
\end{wrapfigure}

Figure~\ref{fig:human} shows that B-cos explanations have a better human interpretability and exhibit greater alignment with human reasoning than post-hoc methods, even though they are not directly optimized for human agreement.
Paired t-tests with a Bonferroni-corrected significance level $\alpha=\frac{0.05}{6}=0.008\overline{3}$~\citep{bonferroni1936teoria} shows that the improvements of B-cos explanations are statistically significant ($p<\alpha$) for both metrics.

\subsection{Qualitative Analysis}\label{sec:qualitative-analysis}
\label{sec:qualitative}

\begin{figure*}[h]
    \centering
    \includegraphics[width=0.8\textwidth]{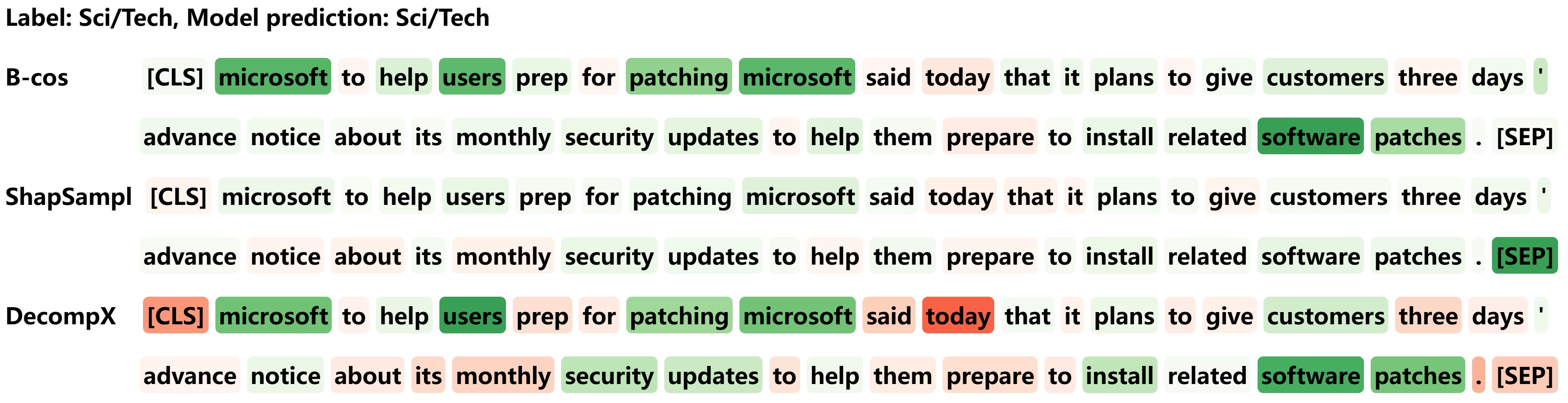}
    \caption{Examples of B-cos explanations (B-cos BERT) as well as ShapSampl and DecompX explanations (conv. BERT) from AG News.
    \textcolor{forestgreen}{Green} (\textcolor{red}{red}) indicates the \textcolor{forestgreen}{positive} (\textcolor{red}{negative}) impact of tokens on the prediction. 
    The B-cos explanation highlights only relevant tokens and is more interpretable to humans.}
    \label{fig:examples}
\end{figure*}

Figure~\ref{fig:examples} provides an example of B-cos and other (post-hoc) explanations.
It can be seen that the B-cos explanation highlights important tokens correctly with little focus on irrelevant ones. 
In contrast, ShapSampl attributes the highest importance to the [SEP] token and provides only little useful information. 
Meanwhile, DecompX extracts a significant amount of irrelevant information. Overall, the B-cos explanation is more interpretable to humans by providing clearer and more relevant attributions compared to the post-hoc explanations. We provide more examples and analyze two undesired explanations from B-cos LMs in Appendix~\ref{appendix:more examples}.

\subsection{Ablation Study}\label{sec:ablation-study}
\label{appendix:ablation}

To better understand B-cos LMs, we conduct an ablation study evaluating the impact of key design choices on task performance and explanation faithfulness. 
\begin{table*}[h]
    \centering
    \resizebox{0.8\textwidth}{!}{
    \begin{tabular}{lcccc}
    \toprule
     & Acc ($\uparrow$) & Comp ($\uparrow$) & Suff ($\downarrow$) & SeqPG ($\uparrow$) \\
     \midrule
    Full system     & 78.64 & 59.66 & -4.89 & 77.57 \\
    \midrule
    w/o alignment pressure (B=1) &  78.07 (\textcolor{red}{0.57}) & 57.19 (\textcolor{red}{2.44}) & -2.57 (\textcolor{red}{2.32}) & 70.18 (\textcolor{red}{7.39}) \\
    w/o BCE training & 79.00 (\textcolor{ao}{0.36}) & 49.22 (\textcolor{red}{10.44}) & -7.91 (\textcolor{ao}{3.02}) & 79.21 (\textcolor{ao}{1.64}) \\
    w/o architectural adaptations & 77.65 (\textcolor{red}{0.99}) & 52.23 (\textcolor{red}{7.43}) & -3.80 (\textcolor{red}{1.09}) & 74.30 (\textcolor{red}{3.27}) \\
    \midrule
    w/o dynamic linear weights (IxG) & 78.64 (0.00) & 44.93 (\textcolor{red}{14.73}) & -0.60 \textcolor{red}{(4.29}) & 53.57 (\textcolor{red}{24.00}) \\
    $\mathbf{W(x)x}$ from conv. model & 80.77 (\textcolor{ao}{2.13}) & 44.92 (\textcolor{red}{14.74}) & 2.80 (\textcolor{red}{7.69}) & 70.20 (\textcolor{red}{7.37})\\
    \bottomrule
    
    \end{tabular}
    }
    
    \caption{Ablation study of key designs in the B-cos BERT model on HateXplain. 
    \textcolor{forestgreen}{Green} (\textcolor{red}{red}) values in parentheses indicate the results are \textcolor{forestgreen}{better} (\textcolor{red}{worse}) than the full system.}
    \label{tab:ablation}
\end{table*}

In Table~\ref{tab:ablation}, we find that removing alignment pressure (using B=1) degrades both task performance and explanation faithfulness. 
Replacing cross-entropy with BCE loss has little effect on classification accuracy, but improves faithfulness in perturbation-based evaluations. 
Architectural adaptations, including removing bias terms and eliminating activation functions in prediction heads, are also critical for enhancing task performance and explainability. 
Besides, we observe numerical instability when generating explanations without these architectural adaptations, as the dynamic linear weights for sigmoid and tanh ($\text{sigmoid}(\mathbf{x})\times\mathbf{x^{-1}}$ and $\text{tanh}(\mathbf{x})\times\mathbf{x^{-1}}$) become unstable when $\mathbf{x}$ is close to zero.

In addition to ablations of model design and training components, we also evaluate alternative explanation methods. Replacing the dynamic linear weights $\mathbf{W(x)}$ with gradients (equivalent to IxG) yields less faithful explanations on B-cos LMs. Besides, directly extracting B-cos-like explanations, $\mathbf{W(x)x}$, from a conventional model results in worse faithfulness compared to those from B-cos LMs.

\section{Comparison of B-cosification Setups}
\label{sec: different setups}

Transforming PLMs into task-specific B-cos LMs involves two key objectives: task fine-tuning and B-cos conversion. While our main experiments combine these two phases, they can also be performed separately. To assess their effects, we compare two alternative training setups:
\begin{itemize}
    \item Task then B-cos: PLMs are first fine-tuned on a downstream task. 
    B-cos adaptations are then applied, followed by further fine-tuning on the same task for B-cos conversion. 
    This setup is equivalent to \citet{arya24bcosification} who apply B-cosification to models with downstream task capabilities.
    \item B-cos then task: B-cos adaptations are applied to PLMs first, followed by unsupervised pre-training to enhance B-cosification.
    The pre-trained B-cos models are then fine-tuned on the downstream task. 
\end{itemize}
\begin{wraptable}{r}{0.45\textwidth} 
\vspace{-3mm}
    \centering
    \resizebox{\linewidth}{!}{
    \tabcolsep=2pt
    \begin{tabular}{@{}lcccc@{}}
    \toprule 
    \textbf{Setup} & \textbf{Epochs} & \textbf{Acc ($\uparrow$)} & \textbf{SeqPG ($\uparrow$)} & \textbf{Steps (K)} \\
    \midrule
    Conv. LM & 5 & 94.06 & - & 6.67 \\
    \midrule
    B-cos LM & 5 & 91.00 & 70.66 & 4.33 \\
    \midrule
    B-cos from scratch & 5 & 88.25 & 60.92 & 4.33 \\
    \midrule
    \multirow{5}{*}{Task then B-cos} & 1+4 & 91.17 & 70.01 & 1+5 \\
     & 2+3 & 91.30 & 70.48 & 3+3.33 \\
     & 3+2 & 91.38 & 70.83 & 4+3 \\
     & 4+1 & 89.56 & 70.66 & 5+1 \\
     & 5+5* & 91.27 & 70.78 & 6.67+3.33 \\
     \midrule
    \multirow{7}{*}{B-cos then task} & 1+4 & 90.64 & 67.07 & 1+5 \\
    & 2+3 & 91.04 & 68.97 & 3+4 \\
    & 3+2 & 90.50 & 68.48 & 4+3 \\
    & 4+1 & 89.18 & 69.92 & 6+1 \\
    & 5+5* & 91.45 & 71.86 & 7+5.33 \\
    & 10+5* & 92.19 & 73.44 & 15+6.33 \\
    & 20+5* & 92.87 & 75.01 & 31+6 \\
    \bottomrule
    \end{tabular}
    }
    \caption{Training epochs, accuracy, explanation faithfulness, and convergence steps for different B-cosification setups. 
    For two-phase methods, we report epoch distribution and convergence steps per phase. 
    * marks additional training epochs.}
    \label{tab:different setup}
    \vspace{-5mm}
\end{wraptable}

We evaluate these setups against the B-cosification approach used in our main experiments (B-cos LM) and compare task performance, faithfulness, and training efficiency (cf. Appendix~\ref{appendix: implementation details} for B-cos pre-training details). 
We also report results for conventional fine-tuning (Conv. LM) and randomly initialized B-cos models (B-cos from scratch).
All experiments are run on IMDB with B=1.25 for B-cos models, with results averaged over three runs.

Table~\ref{tab:different setup} shows that B-cos LM requires fewer training steps to reach optimal validation performance than conventional fine-tuning. 
Training B-cos LM from scratch results in worse accuracy and faithfulness, showing the importance of good parameter initialization.
Among the two setups that separate task fine-tuning and B-cos conversion, \textit{Task then B-cos} achieves results similar to B-cos LM but requires more total training steps. 
\textit{B-cos then task} initially performs worse under the same training budget. 
However, with additional pre-training epochs, it surpasses other B-cosification setups in both task performance and faithfulness.
Overall, we find that combining task fine-tuning and B-cos conversion is the most efficient approach. 
However, with sufficient pre-training, \textit{B-cos then task} can produce more performant and explainable models. 

\section{Impact of B-cosification and B Values}
\label{sec:B}

For a deeper understanding of how B-cosification and alignment pressure parameter B affect model performance and behavior, we compare conventional and B-cos BERT trained on HateXplain across different B values. 
We also provide an empirical analysis of the impact of B on input-weight alignment in Appendix~\ref{appendix:alignment}.

\paragraph{Model Performance}

\begin{wrapfigure}{r}{0.45\textwidth}
\vspace{-5mm}
    \centering
    \includegraphics[width=\linewidth]{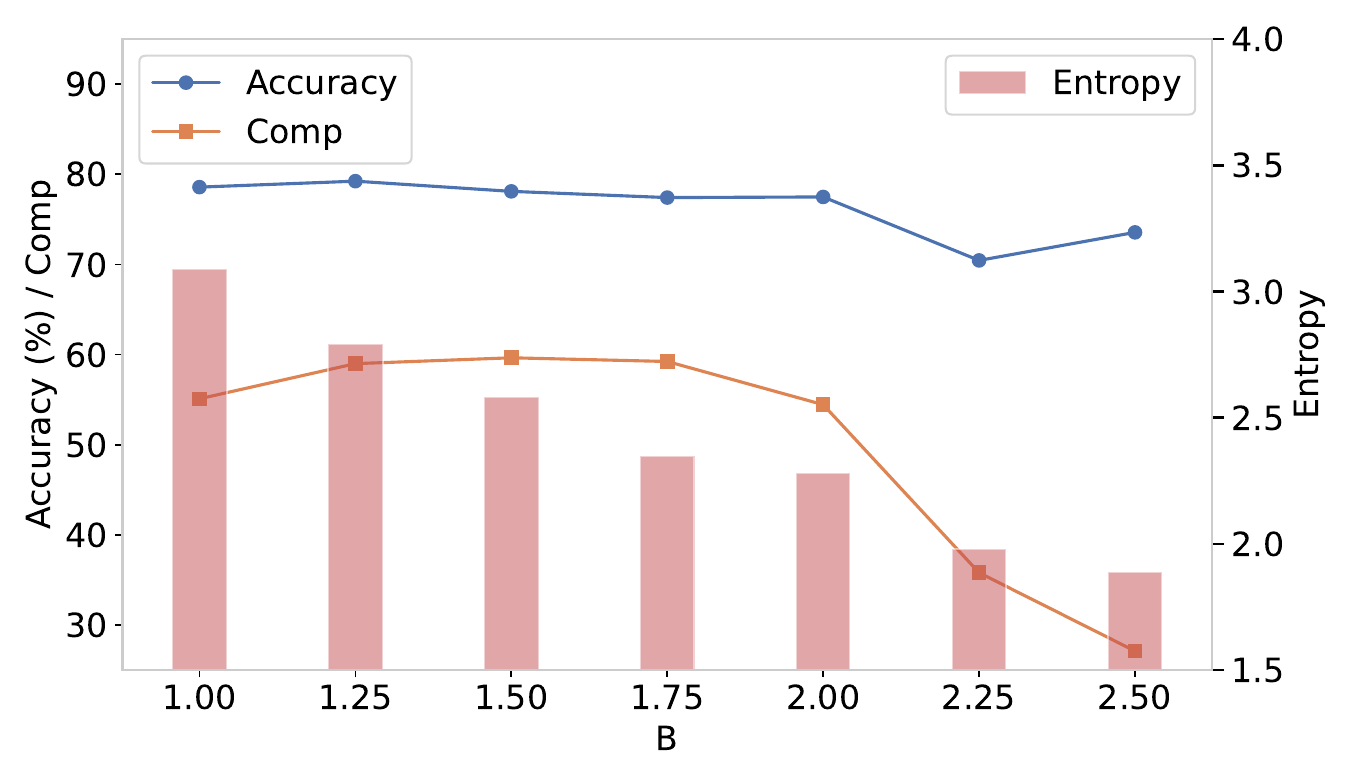}
    \caption{Varying B for B-cos BERT (HateXplain). Accuracy and Comp both peak around B=1.5, while explanation entropy negatively correlates with B.}
    \label{fig:b analysis}
\vspace{-7mm}
\end{wrapfigure}

Figure~\ref{fig:b analysis} shows the effects of varying B on the task performance and explanation faithfulness. 
Classification accuracy initially improves slightly as B increases from 1 to 1.25, benefiting from the extra non-linearity introduced by B>1. 
However, beyond this point, accuracy declines as higher alignment pressure reduces model flexibility. 
A similar trend is observed for Comp, which peaks around B=1.5 before decreasing. This differs from previous findings in vision models~\citep{Bohle_2022_CVPR}, which we attribute to the high sparsity of explanations at larger B values. 
As alignment pressure increases, fewer tokens receive attribution scores that are not close to zero, leading to poor token importance calibration and worse Comp scores. 
The effects of B on other metrics are similar and can be found in Appendix~\ref{appendix:different b all metrics}.

\paragraph{Explanation Entropy}
Figure~\ref{fig:b analysis} also reveals a negative correlation between explanation entropy and B, indicating that higher alignment pressure leads to sparser explanations.
This aligns with our expectations: a larger B amplifies the differences between dimensions in $\mathbf{|\text{cos}(x,\hat{W})|^\text{B-1}}$ of B-cos layers (Equation~\ref{eq:b-cos}) and the dynamic linear weight assigns more distinct attributions to input features. As a result, explanations become more concentrated, where only a few tokens receive high attributions, while most remain close to zero (cf. Appendix~\ref{appendix:examples different b} for an example).

\paragraph{Model Bias}
\label{sec:bias}

\begin{wrapfigure}{r}{0.45\textwidth}
\vspace{-5mm}
    \centering
    \includegraphics[width=\linewidth]{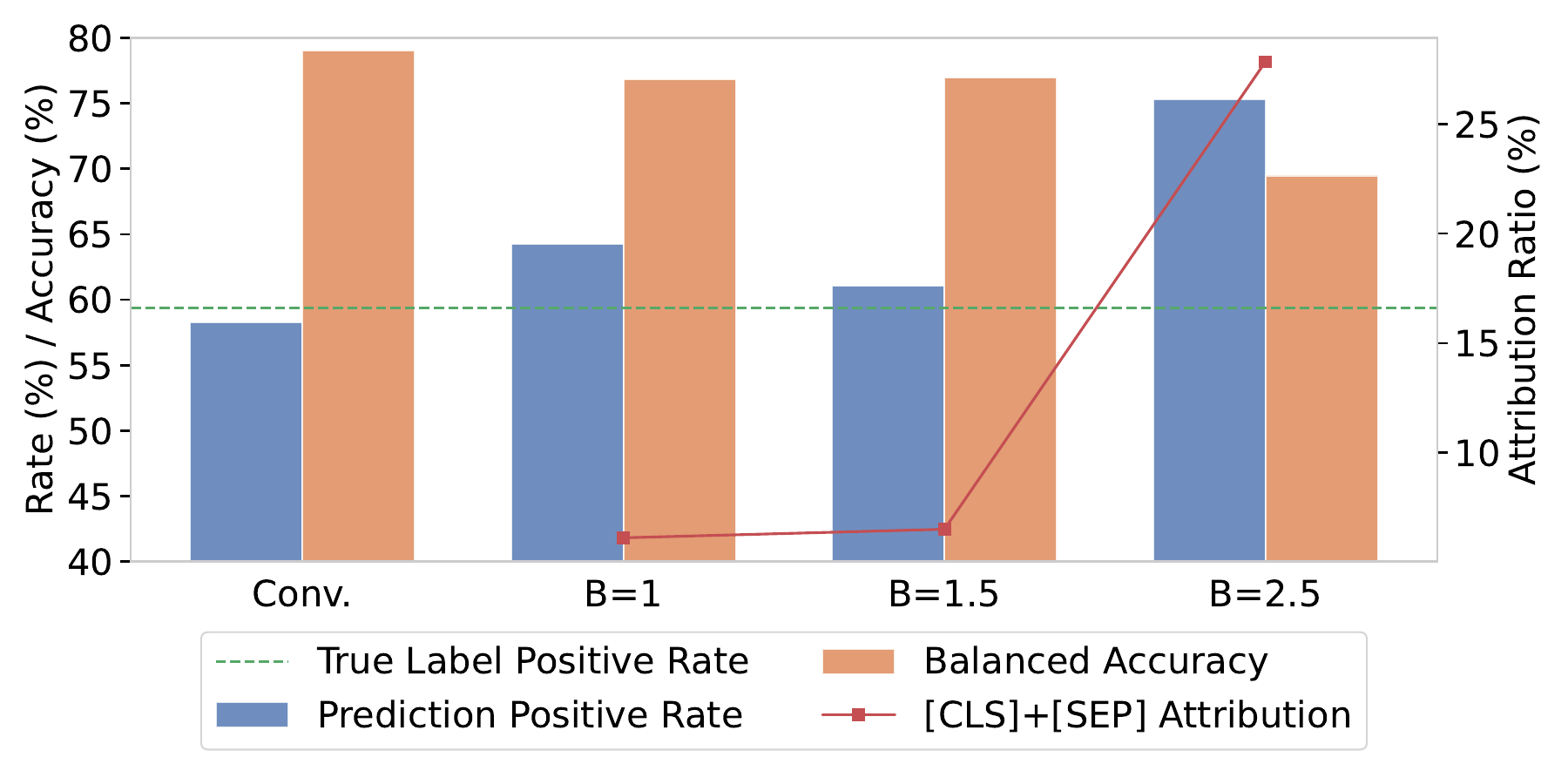}
    \caption{Comparison of conv. BERT and B-cos BERT with different B values. 
    The attributions to [CLS] and [SEP] tokens ($\color{red}\blacksquare$) indicate that B-cos LMs with large B overfit to the non-semantic label distribution.}
    \label{fig:label bias}
\vspace{-7mm}
\end{wrapfigure}

PLMs often acquire biased patterns present in their training data~\citep{wang-demberg-2024-parameter, wang-demberg-2024-rsa}.
Since B-cos LMs with larger B values rely on fewer tokens for prediction, we investigate whether this may cause them to overfit and learn biases in the data. 
For this, we examine label bias and word-level spurious correlations using HateXplain, where approximately 60\% of training and test examples have positive labels and societal biases are present.
Figure~\ref{fig:label bias} shows that a larger B value (B=2.5) reduces the model capacity, leading to a substantially higher positive rate in predictions and therefore lower class-balanced accuracy. 
Moreover, the B=2.5 model assigns higher attributions to non-semantic [CLS] and [SEP] tokens, indicating a reduced reliance on meaningful content. 
Notably, this label bias is not observed in the balanced AG News and IMDB datasets.

We also find that B-cosification, particularly with large B, amplifies reliance on spurious correlations. 
For example, the prediction positive rate for examples with the word ``black'' rises from 49.02\% in the test set and 52.94\% in the conventional model to 59.80\%, 56.86\%, and 73.53\% in B-cos LMs with B=1, 1.5, and, 2.5, respectively (we provide an example in Appendix~\ref{appendix:bias}).
However, the faithfulness and interpretability of B-cos explanations facilitate the detection of spurious correlations and can effectively guide models toward reducing them~\citep{Rao_2023_ICCV, balkir-etal-2022-challenges, bridge-wang-2025}. 
We leave the exploration of B-cos LMs for bias detection and mitigation to future work.

\sloppy
\section{B-cosifying Decoder-Only Models for Generation Tasks}
\label{sec:decoder}

LLMs are increasingly used as general-purpose assistants, with most based on decoder-only architectures~\citep{llm-survey-2023, llm-survey-2024}.
While our primary focus is on classification tasks using encoder-only models, we also extend B-cosification to decoder-only models for generation tasks to demonstrate the broader applicability of B-cos LMs.
Specifically, we apply B-cosification to two decoder-only models, GPT-2 small (\citealp{Radford-2019-GPT}, referred to as GPT-2 afterwards) and Llama-3.2-1B (\citealp{Llama3}, referred to as Llama-3.2 afterwards), and evaluate their language modeling performance and explanation quality on two generation tasks. For more details on the datasets, experimental setup and baseline models, see Appendix~\ref{appendix: implementation details}.

\begin{wraptable}{r}{0.45\textwidth} 
\vspace{-0mm}
    \centering
    \resizebox{\linewidth}{!}{
    \begin{tabular}{@{}lccc@{}}
         \toprule
    \multirow{2}{*}{\textbf{Model}} & \multicolumn{2}{c}{\textbf{Probability Gap ($\uparrow$)}} & \multirow{2}{*}{\textbf{PPL ($\downarrow$)}} \\
    \cmidrule{2-3}
     & BLiMP & IOI & \\
     \midrule
     GPT-2 & 0.0055 & 0.3351 & 3.10 \\
     B-cos GPT-2 & \textbf{0.0059} & 0.3265 & \textbf{3.04} \\
    
     \midrule
      Llama-3.2 & 0.0058 & 0.4652 & 2.51 \\
     B-cos Llama-3.2 & \textbf{0.0065} & \textbf{0.5021} & 2.64 \\
     \bottomrule
    \end{tabular}}
    \caption{Language ability results for vanilla and B-cos decoder-only models. Scores where B-cos LM outperforms their vanilla counterparts are in \textbf{bold}. 
    B-cos LMs show language modeling ability comparable to vanilla models. 
    Results for each subset can be found in Table~\ref{tab:blimp probability difference} in Appendix~\ref{appendix:blimp}.}
    \label{tab:decoder language ability}
    \vspace{-7mm}
\end{wraptable}

\paragraph{B-cosification Setup} Given the complexity of modeling natural language, we use a small B value of 1.1. We do not B-cosify the language head, as its parameters are tied with the embedding layer. We use the standard cross-entropy loss instead of BCE, since the unnormalized language head weights could otherwise grow arbitrarily large to minimize the loss. To convert GPT-2 and Llama-3.2 to B-cos LMs, we apply B-cos adaptations and further train them on 500,000 and 4,000,000 sentences from OpenWebText\footnote{\url{https://huggingface.co/datasets/Skylion007/openwebtext}}, respectively.

\paragraph{Datasets} For explanation evaluation, we use the BLiMP dataset~\citep{warstadt2020blimp} to assess explanations for linguistic phenomena, and the Indirect Object Identification (IOI) dataset~\citep{fahamu_2022} to test models' reasoning about object identification. Following~\citet{ferrando-etal-2023-explaining}, we use nine subsets of BLiMP. 
Each example in both datasets consists of a sentence prefix followed by a target and a foil next word prediction, differing in whether they align with the phenomenon or ability of interest. 
Ground truth evidence is provided to support either grammatical correctness or correct object identification. Examples of these datasets can be found in Table~\ref{tab:decoder dataset} in Appendix~\ref{appendix: implementation details}.
\begin{wraptable}{r}{0.45\textwidth} 
\vspace{-4mm}
    \centering
    \resizebox{\linewidth}{!}{
    \begin{tabular}{@{}lcccc@{}}
    \toprule
       \multirow{2}{*}{\textbf{Method}} & \multicolumn{2}{c}{\textbf{GPT-2 MRR ($\uparrow$)}} & \multicolumn{2}{c}{\textbf{Llama-3.2 MRR ($\uparrow$)}} \\
       \cmidrule{2-3} \cmidrule{4-5} 
       & BLiMP & IOI & BLiMP & IOI \\
       \midrule
       Random & 0.5130 & 0.2360 & 0.5132 & 0.2328 \\
       Grad Norm & 0.5465& 0.8599 & 0.5504 & 0.3637 \\
       IxG & 0.4750 & 0.1112 & 0.5303 & 0.1034 \\
       Occlusion & 0.6365 & 0.8517 & 0.6201 & 0.4767 \\
       Logit & 0.7307 & \textbf{1.0} & - & - \\
       ALTI Logit & 0.7391 & \textbf{1.0} & - & - \\
       \midrule
       B-cos & \textbf{0.7561} & \textbf{1.0} & \textbf{0.6969} & \textbf{0.9913} \\
       \bottomrule
         
    \end{tabular}}
    \caption{Alignment results (MRR) on BLiMP and IOI. Logit and ALTI Logit results are replicated from the original paper~\citep{ferrando-etal-2023-explaining}. Best scores are marked in \textbf{bold}. B-cos explanations achieve the best alignment with ground truth evidence. Results for each subset can be found in Table~\ref{tab:blimp gpt} and Table~\ref{tab:blimp Llama} in Appendix~\ref{appendix:blimp}.}
    \label{tab:decoder alignment}
    \vspace{-7mm}
\end{wraptable}

\paragraph{Metrics and Baselines} We evaluate explanation quality using Mean Reciprocal Rank (MRR), where higher scores indicate stronger alignment with the ground truth evidence. To assess language modeling abilities of models, we report two metrics: (1) the probability gap between target and foil predictions, and (2) perplexity (PPL) on a held-out corpus.
Following~\citet{yin-neubig-2022-interpreting}, we generate contrastive explanations that explain why the model predicts target tokens instead of foil tokens, and compare B-cos explanations against several baseline methods: L1 gradient norm (Grad Norm), IxG, Occlusion, and two propagation-based methods Logit and ALTI Logit from~\citet{ferrando-etal-2023-explaining}. 

\paragraph{Results}

Table~\ref{tab:decoder language ability} shows that B-cos GPT-2 and B-cos Llama-3.2 models achieve strong language modeling performance comparable to their vanilla counterparts. Besides, Table~\ref{tab:decoder alignment} demonstrates that B-cos explanations exhibit better alignment with ground truth across tasks and models, indicating improved explainability of B-cos decoder-only LMs. Although the current B-cosification pipeline requires additional training, future work could explore more efficient approaches that reduce training overhead or integrate B-cosification into the pre-training phase. Overall, we believe B-cos decoder-only models are well-suited for tasks where explainability is critical and represent a promising direction for building more transparent and reliable LLMs.

\section{Practical Guidance for Training B-cos LMs}

Based on our experiments and analyses, we provide the following guidance on configuring B-cos LMs:

\paragraph{B-cosification Setup} For efficient B-cosification of encoder-only models, we recommend combining B-cos conversion and task fine-tuning. However, if resources permit, an additional B-cos pre-training stage can further enhance both task performance and explanation faithfulness. For decoder-only models, B-cosification can be applied to pre-trained models with unsupervised training data, enabling their conversion into B-cos LMs with improved explainability.

\paragraph{Choice of B Value} Several factors influence the selection of an appropriate B value: 

(1) Model capacity and explanation sparsity: Excessively large B values can reduce model capacity and produce overly sparse explanations. We therefore recommend setting B within the range of 1–2. 

(2) Task complexity and language diversity: For complex tasks with varied language use (e.g., online forum data with diverse language styles), smaller B values are preferable, as they preserve model flexibility and capture more useful features. For other tasks, larger B values can improve explainability.

(3) Bias considerations: Be mindful that B-cosification may amplify biases that the model learns from biased training data. When this risk is present, consider choosing smaller B values and applying bias mitigation methods if necessary.


\section{Conclusion}

In this work, we introduce B-cos LM, a bias-free dynamic linear model that learns task-relevant patterns through increased input-weight alignment pressure. 
B-cos LMs generate more faithful and human interpretable explanations while maintaining strong task performance and fast convergence. 
We further explore adapting decoder-only models into B-cos LMs for generation tasks and show that, with additional training, they match the language modeling performance of conventional models while providing better explanations. 
Finally, based on our systematic analysis, we provide practical guidelines for effectively transforming PLMs into B-cos LMs.


\section{Limitations}

This study has certain limitations that should be acknowledged.
First, the automatic evaluation metrics we use may not fully capture the faithfulness of different explanation methods~\citep{feng-etal-2018-pathologies, lapuschkin2019unmasking}. 
However, since there is no universal consensus on the most reliable evaluation metrics, this remains an open challenge in explainability research.

Second, we find that B-cos explanations do not consistently capture token interactions within multi-token phrases. 
For example, a negation phrase like \textit{not good} tends to receive an overall attribution score that aligns with its meaning (e.g., a negative score for positive sentiment), but the individual token scores within the phrase vary across contexts. In some cases, the word \textit{good} may receive either positive or negative scores across different examples, even when the overall sentiment remains the same.
Similar issues arise in other methods, suggesting a broader limitation of token-level rationales in capturing compositional semantics.

Finally, B-cos explanations, like all input-based approaches, are not suitable for all tasks. 
They are particularly suitable for NLP tasks where predictions can be directly attributed to specific words or phrases. 
Examples include various text classification tasks, such as sentiment analysis and toxicity detection, where the presence or absence of certain tokens often provides clear evidence for the model's decision. 
Conversely, for tasks requiring multi-step reasoning or external knowledge (e.g., commonsense reasoning, fact verification), token-level rationales become less informative. 
Higher-level explanations, such as concept-based or natural language explanations, may instead offer more meaningful insight into the model's reasoning.

\section*{Acknowledgements}
This work was funded in part by the Deutsche Forschungsgemeinschaft (DFG, German Research Foundation) -- GRK 2853/1 "Neuroexplicit Models of Language, Vision, and Action" - project number 471607914. 
We also thank all those who provided valuable feedback and suggestions during the development of this work. Their input has been instrumental in improving its quality.

\bibliography{main}

@article{lyu-etal-2024-towards,
    title = "Towards Faithful Model Explanation in {NLP}: A Survey",
    author = "Lyu, Qing  and
      Apidianaki, Marianna  and
      Callison-Burch, Chris",
    journal = "Computational Linguistics",
    volume = "50",
    number = "2",
    month = jun,
    year = "2024",
    address = "Cambridge, MA",
    publisher = "MIT Press",
    url = "https://aclanthology.org/2024.cl-2.6/",
    doi = "10.1162/coli_a_00511",
    pages = "657--723"
}

@inproceedings{wolf-etal-2020-transformers,
    title = "Transformers: State-of-the-Art Natural Language Processing",
    author = "Thomas Wolf and Lysandre Debut and Victor Sanh and Julien Chaumond and Clement Delangue and Anthony Moi and Pierric Cistac and Tim Rault and Rémi Louf and Morgan Funtowicz and Joe Davison and Sam Shleifer and Patrick von Platen and Clara Ma and Yacine Jernite and Julien Plu and Canwen Xu and Teven Le Scao and Sylvain Gugger and Mariama Drame and Quentin Lhoest and Alexander M. Rush",
    booktitle = "Proceedings of the 2020 Conference on Empirical Methods in Natural Language Processing: System Demonstrations",
    month = oct,
    year = "2020",
    address = "Online",
    publisher = "Association for Computational Linguistics",
    url = "https://www.aclweb.org/anthology/2020.emnlp-demos.6",
    pages = "38--45"
}

@misc{Eval-2023-Harness,
	author       = {Gao, Leo and Tow, Jonathan and Abbasi, Baber and Biderman, Stella and Black, Sid and DiPofi, Anthony and Foster, Charles and Golding, Laurence and Hsu, Jeffrey and Le Noac'h, Alain and Li, Haonan and McDonell, Kyle and Muennighoff, Niklas and Ociepa, Chris and Phang, Jason and Reynolds, Laria and Schoelkopf, Hailey and Skowron, Aviya and Sutawika, Lintang and Tang, Eric and Thite, Anish and Wang, Ben and Wang, Kevin and Zou, Andy},
	title        = {A framework for few-shot language model evaluation},
	month        = 12,
	year         = 2023,
	publisher    = {Zenodo},
	version      = {v0.4.0},
	doi          = {10.5281/zenodo.10256836},
	url          = {https://zenodo.org/records/10256836}
}

@InProceedings{Sundararajan-2017-IntegratedGrad,
  title = 	 {Axiomatic Attribution for Deep Networks},
  author =       {Mukund Sundararajan and Ankur Taly and Qiqi Yan},
  booktitle = 	 {Proceedings of the 34th International Conference on Machine Learning},
  pages = 	 {3319--3328},
  year = 	 {2017},
  editor = 	 {Precup, Doina and Teh, Yee Whye},
  volume = 	 {70},
  series = 	 {Proceedings of Machine Learning Research},
  month = 	 {06--11 Aug},
  publisher =    {PMLR},
  url = 	 {https://proceedings.mlr.press/v70/sundararajan17a.html},
}

@inproceedings{DBLP:journals/corr/BahdanauCB14,
  author       = {Dzmitry Bahdanau and
                  Kyunghyun Cho and
                  Yoshua Bengio},
  editor       = {Yoshua Bengio and
                  Yann LeCun},
  title        = {Neural Machine Translation by Jointly Learning to Align and Translate},
  booktitle    = {3rd International Conference on Learning Representations, {ICLR} 2015,
                  San Diego, CA, USA, May 7-9, 2015, Conference Track Proceedings},
  year         = {2015},
  url          = {http://arxiv.org/abs/1409.0473},
  bibsource    = {dblp computer science bibliography, https://dblp.org}
}

@inproceedings{Ribeiro-2016-LIME,
  title={" Why should i trust you?" Explaining the predictions of any classifier},
  author={Ribeiro, Marco Tulio and Singh, Sameer and Guestrin, Carlos},
  booktitle={Proceedings of the 22nd ACM SIGKDD international conference on knowledge discovery and data mining},
  pages={1135--1144},
  year={2016}
}

@inproceedings{DBLP:journals/corr/SimonyanVZ13,
  author       = {Karen Simonyan and
                  Andrea Vedaldi and
                  Andrew Zisserman},
  editor       = {Yoshua Bengio and
                  Yann LeCun},
  title        = {Deep Inside Convolutional Networks: Visualising Image Classification
                  Models and Saliency Maps},
  booktitle    = {2nd International Conference on Learning Representations, {ICLR} 2014,
                  Banff, AB, Canada, April 14-16, 2014, Workshop Track Proceedings},
  year         = {2014},
  url          = {http://arxiv.org/abs/1312.6034},
  bibsource    = {dblp computer science bibliography, https://dblp.org}
}

@article{DBLP:journals/corr/KindermansSMD16,
  author       = {Pieter{-}Jan Kindermans and
                  Kristof Sch{\"{u}}tt and
                  Klaus{-}Robert M{\"{u}}ller and
                  Sven D{\"{a}}hne},
  title        = {Investigating the influence of noise and distractors on the interpretation
                  of neural networks},
  journal      = {CoRR},
  volume       = {abs/1611.07270},
  year         = {2016},
  url          = {http://arxiv.org/abs/1611.07270},
  eprinttype    = {arXiv},
  eprint       = {1611.07270},
  bibsource    = {dblp computer science bibliography, https://dblp.org}
}

@article{bach2015pixel,
  title={On pixel-wise explanations for non-linear classifier decisions by layer-wise relevance propagation},
  author={Bach, Sebastian and Binder, Alexander and Montavon, Gr{\'e}goire and Klauschen, Frederick and M{\"u}ller, Klaus-Robert and Samek, Wojciech},
  journal={PloS one},
  volume={10},
  number={7},
  pages={e0130140},
  year={2015},
  publisher={Public Library of Science San Francisco, CA USA}
}

@inproceedings{shrikumar2017learning,
  title={Learning important features through propagating activation differences},
  author={Shrikumar, Avanti and Greenside, Peyton and Kundaje, Anshul},
  booktitle={International conference on machine learning},
  pages={3145--3153},
  year={2017},
  organization={PMlR}
}

@inproceedings{DBLP:journals/corr/SpringenbergDBR14,
  author       = {Jost Tobias Springenberg and
                  Alexey Dosovitskiy and
                  Thomas Brox and
                  Martin A. Riedmiller},
  editor       = {Yoshua Bengio and
                  Yann LeCun},
  title        = {Striving for Simplicity: The All Convolutional Net},
  booktitle    = {3rd International Conference on Learning Representations, {ICLR} 2015,
                  San Diego, CA, USA, May 7-9, 2015, Workshop Track Proceedings},
  year         = {2015},
  url          = {http://arxiv.org/abs/1412.6806},
  bibsource    = {dblp computer science bibliography, https://dblp.org}
}

@inproceedings{atanasova-etal-2020-diagnostic,
    title = "A Diagnostic Study of Explainability Techniques for Text Classification",
    author = "Atanasova, Pepa  and
      Simonsen, Jakob Grue  and
      Lioma, Christina  and
      Augenstein, Isabelle",
    editor = "Webber, Bonnie  and
      Cohn, Trevor  and
      He, Yulan  and
      Liu, Yang",
    booktitle = "Proceedings of the 2020 Conference on Empirical Methods in Natural Language Processing (EMNLP)",
    month = nov,
    year = "2020",
    address = "Online",
    publisher = "Association for Computational Linguistics",
    url = "https://aclanthology.org/2020.emnlp-main.263/",
    doi = "10.18653/v1/2020.emnlp-main.263",
    pages = "3256--3274"
}

@article{bonferroni1936teoria,
  title={Teoria statistica delle classi e calcolo delle probabilita},
  author={Bonferroni, Carlo},
  journal={Pubblicazioni del R Istituto Superiore di Scienze Economiche e Commericiali di Firenze},
  volume={8},
  pages={3--62},
  year={1936},
  doi = "http://dx.doi.org/10.4135/9781412961288.n455"
}

@inproceedings{arras-etal-2019-evaluating,
    title = "Evaluating Recurrent Neural Network Explanations",
    author = {Arras, Leila  and
      Osman, Ahmed  and
      M\"uller, Klaus-Robert  and
      Samek, Wojciech},
    editor = "Linzen, Tal  and
      Chrupa\l a, Grzegorz  and
      Belinkov, Yonatan  and
      Hupkes, Dieuwke",
    booktitle = "Proceedings of the 2019 ACL Workshop BlackboxNLP: Analyzing and Interpreting Neural Networks for NLP",
    month = aug,
    year = "2019",
    address = "Florence, Italy",
    publisher = "Association for Computational Linguistics",
    url = "https://aclanthology.org/W19-4813/",
    doi = "10.18653/v1/W19-4813",
    pages = "113--126"
}

@article{strumbelj2010efficient,
  title={An efficient explanation of individual classifications using game theory},
  author={Strumbelj, Erik and Kononenko, Igor},
  journal={The Journal of Machine Learning Research},
  volume={11},
  pages={1--18},
  year={2010},
  publisher={JMLR. org}
}

@inproceedings{pruthi-etal-2020-learning,
    title = "Learning to Deceive with Attention-Based Explanations",
    author = "Pruthi, Danish  and
      Gupta, Mansi  and
      Dhingra, Bhuwan  and
      Neubig, Graham  and
      Lipton, Zachary C.",
    editor = "Jurafsky, Dan  and
      Chai, Joyce  and
      Schluter, Natalie  and
      Tetreault, Joel",
    booktitle = "Proceedings of the 58th Annual Meeting of the Association for Computational Linguistics",
    month = jul,
    year = "2020",
    address = "Online",
    publisher = "Association for Computational Linguistics",
    url = "https://aclanthology.org/2020.acl-main.432/",
    doi = "10.18653/v1/2020.acl-main.432",
    pages = "4782--4793",
}

@inproceedings{Slack-2020-FoolingLimeSHAP,
  title={Fooling lime and shap: Adversarial attacks on post hoc explanation methods},
  author={Slack, Dylan and Hilgard, Sophie and Jia, Emily and Singh, Sameer and Lakkaraju, Himabindu},
  booktitle={Proceedings of the AAAI/ACM Conference on AI, Ethics, and Society},
  pages={180--186},
  year={2020}
}

@inproceedings{jain-wallace-2019-attention,
    title = "{A}ttention is not {E}xplanation",
    author = "Jain, Sarthak  and
      Wallace, Byron C.",
    editor = "Burstein, Jill  and
      Doran, Christy  and
      Solorio, Thamar",
    booktitle = "Proceedings of the 2019 Conference of the North {A}merican Chapter of the Association for Computational Linguistics: Human Language Technologies, Volume 1 (Long and Short Papers)",
    month = jun,
    year = "2019",
    address = "Minneapolis, Minnesota",
    publisher = "Association for Computational Linguistics",
    url = "https://aclanthology.org/N19-1357/",
    doi = "10.18653/v1/N19-1357",
    pages = "3543--3556"
}

@article{Kindermans-2019-Reliability,
  title={The (un) reliability of saliency methods},
  author={Kindermans, Pieter-Jan and Hooker, Sara and Adebayo, Julius and Alber, Maximilian and Sch{\"u}tt, Kristof T and D{\"a}hne, Sven and Erhan, Dumitru and Kim, Been},
  journal={Explainable AI: Interpreting, explaining and visualizing deep learning},
  pages={267--280},
  year={2019},
  publisher={Springer}
}

@inproceedings{zheng-etal-2022-irrationality,
    title = "The Irrationality of Neural Rationale Models",
    author = "Zheng, Yiming  and
      Booth, Serena  and
      Shah, Julie  and
      Zhou, Yilun",
    editor = "Verma, Apurv  and
      Pruksachatkun, Yada  and
      Chang, Kai-Wei  and
      Galstyan, Aram  and
      Dhamala, Jwala  and
      Cao, Yang Trista",
    booktitle = "Proceedings of the 2nd Workshop on Trustworthy Natural Language Processing (TrustNLP 2022)",
    month = jul,
    year = "2022",
    address = "Seattle, U.S.A.",
    publisher = "Association for Computational Linguistics",
    url = "https://aclanthology.org/2022.trustnlp-1.6/",
    doi = "10.18653/v1/2022.trustnlp-1.6",
    pages = "64--73"
}

@inproceedings{DBLP:conf/nips/VaswaniSPUJGKP17,
  author       = {Ashish Vaswani and
                  Noam Shazeer and
                  Niki Parmar and
                  Jakob Uszkoreit and
                  Llion Jones and
                  Aidan N. Gomez and
                  Lukasz Kaiser and
                  Illia Polosukhin},
  editor       = {Isabelle Guyon and
                  Ulrike von Luxburg and
                  Samy Bengio and
                  Hanna M. Wallach and
                  Rob Fergus and
                  S. V. N. Vishwanathan and
                  Roman Garnett},
  title        = {Attention is All you Need},
  booktitle    = {Advances in Neural Information Processing Systems 30: Annual Conference
                  on Neural Information Processing Systems 2017, December 4-9, 2017,
                  Long Beach, CA, {USA}},
  pages        = {5998--6008},
  year         = {2017},
  url          = {https://proceedings.neurips.cc/paper/2017/hash/3f5ee243547dee91fbd053c1c4a845aa-Abstract.html},
  bibsource    = {dblp computer science bibliography, https://dblp.org}
}

@article{tutek2022toward,
  title={Toward practical usage of the attention mechanism as a tool for interpretability},
  author={Tutek, Martin and {\v{S}}najder, Jan},
  journal={IEEE access},
  volume={10},
  pages={47011--47030},
  year={2022},
  publisher={IEEE}
}

@inproceedings{atanasova2022diagnostics,
  title={Diagnostics-guided explanation generation},
  author={Atanasova, Pepa and Simonsen, Jakob Grue and Lioma, Christina and Augenstein, Isabelle},
  booktitle={Proceedings of the AAAI Conference on Artificial Intelligence},
  volume={36},
  pages={10445--10453},
  year={2022}
}

@inproceedings{moradi-etal-2020-training,
    title = "Training with Adversaries to Improve Faithfulness of Attention in Neural Machine Translation",
    author = "Moradi, Pooya  and
      Kambhatla, Nishant  and
      Sarkar, Anoop",
    editor = "Shmueli, Boaz  and
      Huang, Yin Jou",
    booktitle = "Proceedings of the 1st Conference of the Asia-Pacific Chapter of the Association for Computational Linguistics and the 10th International Joint Conference on Natural Language Processing: Student Research Workshop",
    month = dec,
    year = "2020",
    address = "Suzhou, China",
    publisher = "Association for Computational Linguistics",
    url = "https://aclanthology.org/2020.aacl-srw.14/",
    doi = "10.18653/v1/2020.aacl-srw.14",
    pages = "93--100"
}

@inproceedings{moradi-etal-2021-measuring,
    title = "Measuring and Improving Faithfulness of Attention in Neural Machine Translation",
    author = "Moradi, Pooya  and
      Kambhatla, Nishant  and
      Sarkar, Anoop",
    editor = "Merlo, Paola  and
      Tiedemann, Jorg  and
      Tsarfaty, Reut",
    booktitle = "Proceedings of the 16th Conference of the European Chapter of the Association for Computational Linguistics: Main Volume",
    month = apr,
    year = "2021",
    address = "Online",
    publisher = "Association for Computational Linguistics",
    url = "https://aclanthology.org/2021.eacl-main.243/",
    doi = "10.18653/v1/2021.eacl-main.243",
    pages = "2791--2802"
}

@inproceedings{NEURIPS2021_e0cd3f16,
 author = {Ismail, Aya Abdelsalam and Corrada Bravo, Hector and Feizi, Soheil},
 booktitle = {Advances in Neural Information Processing Systems},
 editor = {M. Ranzato and A. Beygelzimer and Y. Dauphin and P.S. Liang and J. Wortman Vaughan},
 pages = {26726--26739},
 publisher = {Curran Associates, Inc.},
 title = {Improving Deep Learning Interpretability by Saliency Guided Training},
 url = {https://proceedings.neurips.cc/paper_files/paper/2021/file/e0cd3f16f9e883ca91c2a4c24f47b3d9-Paper.pdf},
 volume = {34},
 year = {2021}
}

@inproceedings{NEURIPS2018_3e9f0fc9,
 author = {Alvarez Melis, David and Jaakkola, Tommi},
 booktitle = {Advances in Neural Information Processing Systems},
 editor = {S. Bengio and H. Wallach and H. Larochelle and K. Grauman and N. Cesa-Bianchi and R. Garnett},
 pages = {},
 publisher = {Curran Associates, Inc.},
 title = {Towards Robust Interpretability with Self-Explaining Neural Networks},
 url = {https://proceedings.neurips.cc/paper_files/paper/2018/file/3e9f0fc9b2f89e043bc6233994dfcf76-Paper.pdf},
 volume = {31},
 year = {2018}
}

@inproceedings{lei-etal-2016-rationalizing,
    title = "Rationalizing Neural Predictions",
    author = "Lei, Tao  and
      Barzilay, Regina  and
      Jaakkola, Tommi",
    editor = "Su, Jian  and
      Duh, Kevin  and
      Carreras, Xavier",
    booktitle = "Proceedings of the 2016 Conference on Empirical Methods in Natural Language Processing",
    month = nov,
    year = "2016",
    address = "Austin, Texas",
    publisher = "Association for Computational Linguistics",
    url = "https://aclanthology.org/D16-1011/",
    doi = "10.18653/v1/D16-1011",
    pages = "107--117"
}

@InProceedings{Bohle_2022_CVPR,
    author    = {B\"ohle, Moritz and Fritz, Mario and Schiele, Bernt},
    title     = {B-Cos Networks: Alignment Is All We Need for Interpretability},
    booktitle = {Proceedings of the IEEE/CVF Conference on Computer Vision and Pattern Recognition (CVPR)},
    month     = {June},
    year      = {2022},
    pages     = {10329-10338}
}

@article{bohle2024b,
  title={B-cos Alignment for Inherently Interpretable CNNs and Vision Transformers},
  author={B\"ohle, Moritz and Singh, Navdeeppal and Fritz, Mario and Schiele, Bernt},
  journal={IEEE Transactions on Pattern Analysis and Machine Intelligence},
  year={2024},
  publisher={IEEE}
}

@inproceedings{modarressi-etal-2023-decompx,
    title = "{D}ecomp{X}: Explaining Transformers Decisions by Propagating Token Decomposition",
    author = "Modarressi, Ali  and
      Fayyaz, Mohsen  and
      Aghazadeh, Ehsan  and
      Yaghoobzadeh, Yadollah  and
      Pilehvar, Mohammad Taher",
    editor = "Rogers, Anna  and
      Boyd-Graber, Jordan  and
      Okazaki, Naoaki",
    booktitle = "Proceedings of the 61st Annual Meeting of the Association for Computational Linguistics (Volume 1: Long Papers)",
    month = jul,
    year = "2023",
    address = "Toronto, Canada",
    publisher = "Association for Computational Linguistics",
    url = "https://aclanthology.org/2023.acl-long.149/",
    doi = "10.18653/v1/2023.acl-long.149",
    pages = "2649--2664"
}

@inproceedings{deyoung-etal-2020-eraser,
    title = "{ERASER}: {A} Benchmark to Evaluate Rationalized {NLP} Models",
    author = "DeYoung, Jay  and
      Jain, Sarthak  and
      Rajani, Nazneen Fatema  and
      Lehman, Eric  and
      Xiong, Caiming  and
      Socher, Richard  and
      Wallace, Byron C.",
    editor = "Jurafsky, Dan  and
      Chai, Joyce  and
      Schluter, Natalie  and
      Tetreault, Joel",
    booktitle = "Proceedings of the 58th Annual Meeting of the Association for Computational Linguistics",
    month = jul,
    year = "2020",
    address = "Online",
    publisher = "Association for Computational Linguistics",
    url = "https://aclanthology.org/2020.acl-main.408/",
    doi = "10.18653/v1/2020.acl-main.408",
    pages = "4443--4458"
}

@inproceedings{poerner-etal-2018-evaluating,
    title = "Evaluating neural network explanation methods using hybrid documents and morphosyntactic agreement",
    author = {Poerner, Nina  and
      Sch\"utze, Hinrich  and
      Roth, Benjamin},
    editor = "Gurevych, Iryna  and
      Miyao, Yusuke",
    booktitle = "Proceedings of the 56th Annual Meeting of the Association for Computational Linguistics (Volume 1: Long Papers)",
    month = jul,
    year = "2018",
    address = "Melbourne, Australia",
    publisher = "Association for Computational Linguistics",
    url = "https://aclanthology.org/P18-1032/",
    doi = "10.18653/v1/P18-1032",
    pages = "340--350"
}

@inproceedings{enguehard-2023-sequential,
    title = "Sequential Integrated Gradients: a simple but effective method for explaining language models",
    author = "Enguehard, Joseph",
    editor = "Rogers, Anna  and
      Boyd-Graber, Jordan  and
      Okazaki, Naoaki",
    booktitle = "Findings of the Association for Computational Linguistics: ACL 2023",
    month = jul,
    year = "2023",
    address = "Toronto, Canada",
    publisher = "Association for Computational Linguistics",
    url = "https://aclanthology.org/2023.findings-acl.477/",
    doi = "10.18653/v1/2023.findings-acl.477",
    pages = "7555--7565"
}

@inproceedings{jacovi-goldberg-2020-towards,
    title = "Towards Faithfully Interpretable {NLP} Systems: How Should We Define and Evaluate Faithfulness?",
    author = "Jacovi, Alon  and
      Goldberg, Yoav",
    editor = "Jurafsky, Dan  and
      Chai, Joyce  and
      Schluter, Natalie  and
      Tetreault, Joel",
    booktitle = "Proceedings of the 58th Annual Meeting of the Association for Computational Linguistics",
    month = jul,
    year = "2020",
    address = "Online",
    publisher = "Association for Computational Linguistics",
    url = "https://aclanthology.org/2020.acl-main.386/",
    doi = "10.18653/v1/2020.acl-main.386",
    pages = "4198--4205"
}

@inproceedings{Lundberg-2017-SHAP,
 author = {Lundberg, Scott M and Lee, Su-In},
 booktitle = {Advances in Neural Information Processing Systems},
 editor = {I. Guyon and U. Von Luxburg and S. Bengio and H. Wallach and R. Fergus and S. Vishwanathan and R. Garnett},
 pages = {1--10},
 publisher = {Curran Associates, Inc.},
 title = {A Unified Approach to Interpreting Model Predictions},
 url = {https://proceedings.neurips.cc/paper_files/paper/2017/file/8a20a8621978632d76c43dfd28b67767-Paper.pdf},
 volume = {30},
 year = {2017}
}

@article{jacovi-goldberg-2021-aligning,
    title = "Aligning Faithful Interpretations with their Social Attribution",
    author = "Jacovi, Alon  and
      Goldberg, Yoav",
    editor = "Roark, Brian  and
      Nenkova, Ani",
    journal = "Transactions of the Association for Computational Linguistics",
    volume = "9",
    year = "2021",
    address = "Cambridge, MA",
    publisher = "MIT Press",
    url = "https://aclanthology.org/2021.tacl-1.18/",
    doi = "10.1162/tacl_a_00367",
    pages = "294--310"
}

@inproceedings{Devlin-2019-BERT,
  author       = {Jacob Devlin and
                  Ming{-}Wei Chang and
                  Kenton Lee and
                  Kristina Toutanova},
  editor       = {Jill Burstein and
                  Christy Doran and
                  Thamar Solorio},
  title        = {{BERT:} Pre-training of Deep Bidirectional Transformers for Language
                  Understanding},
  booktitle    = {Proceedings of the 2019 Conference of the North American Chapter of
                  the Association for Computational Linguistics: Human Language Technologies,
                  {NAACL-HLT} 2019, Minneapolis, MN, USA, June 2-7, 2019, Volume 1 (Long
                  and Short Papers)},
  pages        = {4171--4186},
  publisher    = {Association for Computational Linguistics},
  year         = {2019},
  url          = {https://doi.org/10.18653/v1/n19-1423},
  doi          = {10.18653/V1/N19-1423},
  bibsource    = {dblp computer science bibliography, https://dblp.org}
}

@article{Radford-2019-GPT,
  title={Language models are unsupervised multitask learners},
  author={Radford, Alec and Wu, Jeffrey and Child, Rewon and Luan, David and Amodei, Dario and Sutskever, Ilya and others},
  journal={OpenAI blog},
  volume={1},
  number={8},
  pages={9},
  year={2019}
}

@inproceedings{NEURIPS2020_1457c0d6,
 author = {Brown, Tom and Mann, Benjamin and Ryder, Nick and Subbiah, Melanie and Kaplan, Jared D and Dhariwal, Prafulla and Neelakantan, Arvind and Shyam, Pranav and Sastry, Girish and Askell, Amanda and Agarwal, Sandhini and Herbert-Voss, Ariel and Krueger, Gretchen and Henighan, Tom and Child, Rewon and Ramesh, Aditya and Ziegler, Daniel and Wu, Jeffrey and Winter, Clemens and Hesse, Chris and Chen, Mark and Sigler, Eric and Litwin, Mateusz and Gray, Scott and Chess, Benjamin and Clark, Jack and Berner, Christopher and McCandlish, Sam and Radford, Alec and Sutskever, Ilya and Amodei, Dario},
 booktitle = {Advances in Neural Information Processing Systems},
 editor = {H. Larochelle and M. Ranzato and R. Hadsell and M.F. Balcan and H. Lin},
 pages = {1877--1901},
 publisher = {Curran Associates, Inc.},
 title = {Language Models are Few-Shot Learners},
 url = {https://proceedings.neurips.cc/paper_files/paper/2020/file/1457c0d6bfcb4967418bfb8ac142f64a-Paper.pdf},
 volume = {33},
 year = {2020}
}

@article{DBLP:journals/corr/abs-2303-08774,
  author       = {OpenAI},
  title        = {{GPT-4} Technical Report},
  journal      = {CoRR},
  volume       = {abs/2303.08774},
  year         = {2023},
  url          = {https://doi.org/10.48550/arXiv.2303.08774},
  doi          = {10.48550/ARXIV.2303.08774},
  eprinttype    = {arXiv},
  eprint       = {2303.08774},
  bibsource    = {dblp computer science bibliography, https://dblp.org}
}

@inproceedings{wang-etal-2018-glue,
    title = "{GLUE}: A Multi-Task Benchmark and Analysis Platform for Natural Language Understanding",
    author = "Wang, Alex  and
      Singh, Amanpreet  and
      Michael, Julian  and
      Hill, Felix  and
      Levy, Omer  and
      Bowman, Samuel",
    editor = "Linzen, Tal  and
      Chrupa\l a, Grzegorz  and
      Alishahi, Afra",
    booktitle = "Proceedings of the 2018 {EMNLP} Workshop {B}lackbox{NLP}: Analyzing and Interpreting Neural Networks for {NLP}",
    month = nov,
    year = "2018",
    address = "Brussels, Belgium",
    publisher = "Association for Computational Linguistics",
    url = "https://aclanthology.org/W18-5446/",
    doi = "10.18653/v1/W18-5446",
    pages = "353--355",
}

@article{Bommasani-2021-FoundationModels,
  author       = {Rishi Bommasani and
                  Drew A. Hudson and
                  Ehsan Adeli and
                  Russ B. Altman and
                  Simran Arora and
                  Sydney von Arx and
                  Michael S. Bernstein and
                  Jeannette Bohg and
                  Antoine Bosselut and
                  Emma Brunskill and
                  Erik Brynjolfsson and
                  Shyamal Buch and
                  Dallas Card and
                  Rodrigo Castellon and
                  Niladri S. Chatterji and
                  Annie S. Chen and
                  Kathleen Creel and
                  Jared Quincy Davis and
                  Dorottya Demszky and
                  Chris Donahue and
                  Moussa Doumbouya and
                  Esin Durmus and
                  Stefano Ermon and
                  John Etchemendy and
                  Kawin Ethayarajh and
                  Li Fei{-}Fei and
                  Chelsea Finn and
                  Trevor Gale and
                  Lauren E. Gillespie and
                  Karan Goel and
                  Noah D. Goodman and
                  Shelby Grossman and
                  Neel Guha and
                  Tatsunori Hashimoto and
                  Peter Henderson and
                  John Hewitt and
                  Daniel E. Ho and
                  Jenny Hong and
                  Kyle Hsu and
                  Jing Huang and
                  Thomas Icard and
                  Saahil Jain and
                  Dan Jurafsky and
                  Pratyusha Kalluri and
                  Siddharth Karamcheti and
                  Geoff Keeling and
                  Fereshte Khani and
                  Omar Khattab and
                  Pang Wei Koh and
                  Mark S. Krass and
                  Ranjay Krishna and
                  Rohith Kuditipudi and
                  et al.},
  title        = {On the Opportunities and Risks of Foundation Models},
  journal      = {CoRR},
  volume       = {abs/2108.07258},
  year         = {2021},
  url          = {https://arxiv.org/abs/2108.07258},
  eprinttype    = {arXiv},
  eprint       = {2108.07258},
  bibsource    = {dblp computer science bibliography, https://dblp.org}
}

@article{DBLP:journals/corr/abs-1907-11692,
  author       = {Yinhan Liu and
                  Myle Ott and
                  Naman Goyal and
                  Jingfei Du and
                  Mandar Joshi and
                  Danqi Chen and
                  Omer Levy and
                  Mike Lewis and
                  Luke Zettlemoyer and
                  Veselin Stoyanov},
  title        = {RoBERTa: {A} Robustly Optimized {BERT} Pretraining Approach},
  journal      = {CoRR},
  volume       = {abs/1907.11692},
  year         = {2019},
  url          = {http://arxiv.org/abs/1907.11692},
  eprinttype    = {arXiv},
  eprint       = {1907.11692},
  bibsource    = {dblp computer science bibliography, https://dblp.org}
}

@article{DBLP:journals/corr/abs-1910-01108,
  author       = {Victor Sanh and
                  Lysandre Debut and
                  Julien Chaumond and
                  Thomas Wolf},
  title        = {DistilBERT, a distilled version of {BERT:} smaller, faster, cheaper
                  and lighter},
  journal      = {CoRR},
  volume       = {abs/1910.01108},
  year         = {2019},
  url          = {http://arxiv.org/abs/1910.01108},
  eprinttype    = {arXiv},
  eprint       = {1910.01108},
  bibsource    = {dblp computer science bibliography, https://dblp.org}
}

@inproceedings{DBLP:conf/icml/NairH10,
  author       = {Vinod Nair and
                  Geoffrey E. Hinton},
  editor       = {Johannes F{\"{u}}rnkranz and
                  Thorsten Joachims},
  title        = {Rectified Linear Units Improve Restricted Boltzmann Machines},
  booktitle    = {Proceedings of the 27th International Conference on Machine Learning
                  (ICML-10), June 21-24, 2010, Haifa, Israel},
  pages        = {807--814},
  publisher    = {Omnipress},
  year         = {2010},
  url          = {https://icml.cc/Conferences/2010/papers/432.pdf},
  bibsource    = {dblp computer science bibliography, https://dblp.org}
}

@inproceedings{NEURIPS2019_80537a94,
 author = {Srinivas, Suraj and Fleuret, Fran\c{c}ois},
 booktitle = {Advances in Neural Information Processing Systems},
 editor = {H. Wallach and H. Larochelle and A. Beygelzimer and F. d\textquotesingle Alch\'{e}-Buc and E. Fox and R. Garnett},
 pages = {},
 publisher = {Curran Associates, Inc.},
 title = {Full-Gradient Representation for Neural Network Visualization},
 url = {https://proceedings.neurips.cc/paper_files/paper/2019/file/80537a945c7aaa788ccfcdf1b99b5d8f-Paper.pdf},
 volume = {32},
 year = {2019}
}

@inproceedings{bohle2021convolutional,
  title={Convolutional dynamic alignment networks for interpretable classifications},
  author={B\"ohle, Moritz and Fritz, Mario and Schiele, Bernt},
  booktitle={Proceedings of the IEEE/CVF Conference on Computer Vision and Pattern Recognition},
  pages={10029--10038},
  year={2021}
}

@inproceedings{Kindermans-2018-PatternNet,
  author       = {Pieter{-}Jan Kindermans and
                  Kristof T. Sch{\"{u}}tt and
                  Maximilian Alber and
                  Klaus{-}Robert M{\"{u}}ller and
                  Dumitru Erhan and
                  Been Kim and
                  Sven D{\"{a}}hne},
  title        = {Learning how to explain neural networks: PatternNet and PatternAttribution},
  booktitle    = {6th International Conference on Learning Representations, {ICLR} 2018,
                  Vancouver, BC, Canada, April 30 - May 3, 2018, Conference Track Proceedings},
  publisher    = {OpenReview.net},
  year         = {2018},
  url          = {https://openreview.net/forum?id=Hkn7CBaTW},
  bibsource    = {dblp computer science bibliography, https://dblp.org}
}

@InProceedings{pmlr-v97-wang19p,
  title = 	 {Bias Also Matters: Bias Attribution for Deep Neural Network Explanation},
  author =       {Wang, Shengjie and Zhou, Tianyi and Bilmes, Jeff},
  booktitle = 	 {Proceedings of the 36th International Conference on Machine Learning},
  pages = 	 {6659--6667},
  year = 	 {2019},
  editor = 	 {Chaudhuri, Kamalika and Salakhutdinov, Ruslan},
  volume = 	 {97},
  series = 	 {Proceedings of Machine Learning Research},
  month = 	 {09--15 Jun},
  publisher =    {PMLR},
  url = 	 {https://proceedings.mlr.press/v97/wang19p.html}
}

@inproceedings{NIPS2015_250cf8b5,
 author = {Zhang, Xiang and Zhao, Junbo and LeCun, Yann},
 booktitle = {Advances in Neural Information Processing Systems},
 editor = {C. Cortes and N. Lawrence and D. Lee and M. Sugiyama and R. Garnett},
 pages = {},
 publisher = {Curran Associates, Inc.},
 title = {Character-level Convolutional Networks for Text Classification},
 url = {https://proceedings.neurips.cc/paper_files/paper/2015/file/250cf8b51c773f3f8dc8b4be867a9a02-Paper.pdf},
 volume = {28},
 year = {2015}
}

@inproceedings{maas-etal-2011-learning,
    title = "Learning Word Vectors for Sentiment Analysis",
    author = "Maas, Andrew L.  and
      Daly, Raymond E.  and
      Pham, Peter T.  and
      Huang, Dan  and
      Ng, Andrew Y.  and
      Potts, Christopher",
    editor = "Lin, Dekang  and
      Matsumoto, Yuji  and
      Mihalcea, Rada",
    booktitle = "Proceedings of the 49th Annual Meeting of the Association for Computational Linguistics: Human Language Technologies",
    month = jun,
    year = "2011",
    address = "Portland, Oregon, USA",
    publisher = "Association for Computational Linguistics",
    url = "https://aclanthology.org/P11-1015/",
    pages = "142--150"
}

@inproceedings{mathew2021hatexplain,
  title={Hatexplain: A benchmark dataset for explainable hate speech detection},
  author={Mathew, Binny and Saha, Punyajoy and Yimam, Seid Muhie and Biemann, Chris and Goyal, Pawan and Mukherjee, Animesh},
  booktitle={Proceedings of the AAAI conference on artificial intelligence},
  volume={35},
  pages={14867--14875},
  year={2021}
}

@inproceedings{ethayarajh-2019-contextual,
    title = "How Contextual are Contextualized Word Representations? {C}omparing the Geometry of {BERT}, {ELM}o, and {GPT}-2 Embeddings",
    author = "Ethayarajh, Kawin",
    editor = "Inui, Kentaro  and
      Jiang, Jing  and
      Ng, Vincent  and
      Wan, Xiaojun",
    booktitle = "Proceedings of the 2019 Conference on Empirical Methods in Natural Language Processing and the 9th International Joint Conference on Natural Language Processing (EMNLP-IJCNLP)",
    month = nov,
    year = "2019",
    address = "Hong Kong, China",
    publisher = "Association for Computational Linguistics",
    url = "https://aclanthology.org/D19-1006/",
    doi = "10.18653/v1/D19-1006",
    pages = "55--65"
}

@inproceedings{li-etal-2020-sentence,
    title = "On the Sentence Embeddings from Pre-trained Language Models",
    author = "Li, Bohan  and
      Zhou, Hao  and
      He, Junxian  and
      Wang, Mingxuan  and
      Yang, Yiming  and
      Li, Lei",
    editor = "Webber, Bonnie  and
      Cohn, Trevor  and
      He, Yulan  and
      Liu, Yang",
    booktitle = "Proceedings of the 2020 Conference on Empirical Methods in Natural Language Processing (EMNLP)",
    month = nov,
    year = "2020",
    address = "Online",
    publisher = "Association for Computational Linguistics",
    url = "https://aclanthology.org/2020.emnlp-main.733/",
    doi = "10.18653/v1/2020.emnlp-main.733",
    pages = "9119--9130"
}

@article{DBLP:journals/corr/HendrycksG16,
  author       = {Dan Hendrycks and
                  Kevin Gimpel},
  title        = {Bridging Nonlinearities and Stochastic Regularizers with Gaussian
                  Error Linear Units},
  journal      = {CoRR},
  volume       = {abs/1606.08415},
  year         = {2016},
  url          = {http://arxiv.org/abs/1606.08415},
  eprinttype    = {arXiv},
  eprint       = {1606.08415},
  bibsource    = {dblp computer science bibliography, https://dblp.org}
}

@article{lapuschkin2019unmasking,
  title={Unmasking Clever Hans predictors and assessing what machines really learn},
  author={Lapuschkin, Sebastian and W{\"a}ldchen, Stephan and Binder, Alexander and Montavon, Gr{\'e}goire and Samek, Wojciech and M{\"u}ller, Klaus-Robert},
  journal={Nature communications},
  volume={10},
  number={1},
  pages={1096},
  year={2019},
  publisher={Nature Publishing Group UK London}
}

@inproceedings{feng-etal-2018-pathologies,
    title = "Pathologies of Neural Models Make Interpretations Difficult",
    author = "Feng, Shi  and
      Wallace, Eric  and
      Grissom II, Alvin  and
      Iyyer, Mohit  and
      Rodriguez, Pedro  and
      Boyd-Graber, Jordan",
    editor = "Riloff, Ellen  and
      Chiang, David  and
      Hockenmaier, Julia  and
      Tsujii, Jun{'}ichi",
    booktitle = "Proceedings of the 2018 Conference on Empirical Methods in Natural Language Processing",
    month = oct # "-" # nov,
    year = "2018",
    address = "Brussels, Belgium",
    publisher = "Association for Computational Linguistics",
    url = "https://aclanthology.org/D18-1407/",
    doi = "10.18653/v1/D18-1407",
    pages = "3719--3728"
}

@article{Smilkov-2017-SmoothGrad,
  author       = {Daniel Smilkov and
                  Nikhil Thorat and
                  Been Kim and
                  Fernanda B. Vi{\'{e}}gas and
                  Martin Wattenberg},
  title        = {SmoothGrad: removing noise by adding noise},
  journal      = {CoRR},
  volume       = {abs/1706.03825},
  year         = {2017},
  url          = {http://arxiv.org/abs/1706.03825},
  eprinttype    = {arXiv},
  eprint       = {1706.03825},
  bibsource    = {dblp computer science bibliography, https://dblp.org}
}

@article{DBLP:journals/corr/LiMJ16a,
  author       = {Jiwei Li and
                  Will Monroe and
                  Dan Jurafsky},
  title        = {Understanding Neural Networks through Representation Erasure},
  journal      = {CoRR},
  volume       = {abs/1612.08220},
  year         = {2016},
  url          = {http://arxiv.org/abs/1612.08220},
  eprinttype    = {arXiv},
  eprint       = {1612.08220},
  bibsource    = {dblp computer science bibliography, https://dblp.org}
}

@inproceedings{chrysostomou-aletras-2021-enjoy,
    title = "Enjoy the Salience: Towards Better Transformer-based Faithful Explanations with Word Salience",
    author = "Chrysostomou, George  and
      Aletras, Nikolaos",
    editor = "Moens, Marie-Francine  and
      Huang, Xuanjing  and
      Specia, Lucia  and
      Yih, Scott Wen-tau",
    booktitle = "Proceedings of the 2021 Conference on Empirical Methods in Natural Language Processing",
    month = nov,
    year = "2021",
    address = "Online and Punta Cana, Dominican Republic",
    publisher = "Association for Computational Linguistics",
    url = "https://aclanthology.org/2021.emnlp-main.645/",
    doi = "10.18653/v1/2021.emnlp-main.645",
    pages = "8189--8200"
}

@article{rudin2019stop,
  title={Stop explaining black box machine learning models for high stakes decisions and use interpretable models instead},
  author={Rudin, Cynthia},
  journal={Nature machine intelligence},
  volume={1},
  number={5},
  pages={206--215},
  year={2019},
  publisher={Nature Publishing Group UK London}
}

@inproceedings{arya24bcosification,
 author = {Arya, Shreyash and Rao, Sukrut and B\"{o}hle, Moritz and Schiele, Bernt},
 booktitle = {Advances in Neural Information Processing Systems},
 editor = {A. Globerson and L. Mackey and D. Belgrave and A. Fan and U. Paquet and J. Tomczak and C. Zhang},
 pages = {62756--62786},
 publisher = {Curran Associates, Inc.},
 title = {B-cosification: Transforming Deep Neural Networks to be Inherently LInterpretable},
 url = {https://proceedings.neurips.cc/paper_files/paper/2024/file/72d50a87b218d84c175d16f4557f7e12-Paper-Conference.pdf},
 volume = {37},
 year = {2024}
}

@InProceedings{Rao_2023_ICCV,
    author    = {Rao, Sukrut and B\"ohle, Moritz and Parchami-Araghi, Amin and Schiele, Bernt},
    title     = {Studying How to Efficiently and Effectively Guide Models with Explanations},
    booktitle = {Proceedings of the IEEE/CVF International Conference on Computer Vision (ICCV)},
    month     = {October},
    year      = {2023},
    pages     = {1922-1933}
}

@article{DBLP:journals/corr/abs-1902-00006,
  author       = {Isaac Lage and
                  Emily Chen and
                  Jeffrey He and
                  Menaka Narayanan and
                  Been Kim and
                  Sam Gershman and
                  Finale Doshi{-}Velez},
  title        = {An Evaluation of the Human-Interpretability of Explanation},
  journal      = {CoRR},
  volume       = {abs/1902.00006},
  year         = {2019},
  url          = {http://arxiv.org/abs/1902.00006},
  eprinttype    = {arXiv},
  eprint       = {1902.00006},
  bibsource    = {dblp computer science bibliography, https://dblp.org}
}

@article{DBLP:journals/corr/abs-2009-07896,
  author       = {Narine Kokhlikyan and
                  Vivek Miglani and
                  Miguel Martin and
                  Edward Wang and
                  Bilal Alsallakh and
                  Jonathan Reynolds and
                  Alexander Melnikov and
                  Natalia Kliushkina and
                  Carlos Araya and
                  Siqi Yan and
                  Orion Reblitz{-}Richardson},
  title        = {Captum: {A} unified and generic model interpretability library for
                  PyTorch},
  journal      = {CoRR},
  volume       = {abs/2009.07896},
  year         = {2020},
  url          = {https://arxiv.org/abs/2009.07896},
  eprinttype    = {arXiv},
  eprint       = {2009.07896},
  bibsource    = {dblp computer science bibliography, https://dblp.org}
}

@inproceedings{radford2021learning,
  title={Learning transferable visual models from natural language supervision},
  author={Radford, Alec and Kim, Jong Wook and Hallacy, Chris and Ramesh, Aditya and Goh, Gabriel and Agarwal, Sandhini and Sastry, Girish and Askell, Amanda and Mishkin, Pamela and Clark, Jack and others},
  booktitle={International Conference on Machine Learning},
  pages={8748--8763},
  year={2021},
  organization={PMLR}
}

@inproceedings{enouen-etal-2024-textgenshap,
    title = "{T}ext{G}en{SHAP}: Scalable Post-Hoc Explanations in Text Generation with Long Documents",
    author = "Enouen, James  and
      Nakhost, Hootan  and
      Ebrahimi, Sayna  and
      Arik, Sercan  and
      Liu, Yan  and
      Pfister, Tomas",
    editor = "Ku, Lun-Wei  and
      Martins, Andre  and
      Srikumar, Vivek",
    booktitle = "Findings of the Association for Computational Linguistics: ACL 2024",
    month = aug,
    year = "2024",
    address = "Bangkok, Thailand",
    publisher = "Association for Computational Linguistics",
    url = "https://aclanthology.org/2024.findings-acl.832/",
    doi = "10.18653/v1/2024.findings-acl.832",
    pages = "13984--14011"
}

@inproceedings{ferrando-etal-2023-explaining,
    title = "Explaining How Transformers Use Context to Build Predictions",
    author = "Ferrando, Javier  and
      G{\'a}llego, Gerard I.  and
      Tsiamas, Ioannis  and
      Costa-juss{\`a}, Marta R.",
    editor = "Rogers, Anna  and
      Boyd-Graber, Jordan  and
      Okazaki, Naoaki",
    booktitle = "Proceedings of the 61st Annual Meeting of the Association for Computational Linguistics (Volume 1: Long Papers)",
    month = jul,
    year = "2023",
    address = "Toronto, Canada",
    publisher = "Association for Computational Linguistics",
    url = "https://aclanthology.org/2023.acl-long.301/",
    doi = "10.18653/v1/2023.acl-long.301",
    pages = "5486--5513"
}

@inproceedings{barkan-etal-2024-llm,
    title = "{LLM} Explainability via Attributive Masking Learning",
    author = "Barkan, Oren  and
      Toib, Yonatan  and
      Elisha, Yehonatan  and
      Weill, Jonathan  and
      Koenigstein, Noam",
    editor = "Al-Onaizan, Yaser  and
      Bansal, Mohit  and
      Chen, Yun-Nung",
    booktitle = "Findings of the Association for Computational Linguistics: EMNLP 2024",
    month = nov,
    year = "2024",
    address = "Miami, Florida, USA",
    publisher = "Association for Computational Linguistics",
    url = "https://aclanthology.org/2024.findings-emnlp.556/",
    doi = "10.18653/v1/2024.findings-emnlp.556",
    pages = "9522--9537"
}

@inproceedings{cohen-etal-2024-contextcite,
 author = {Cohen-Wang, Benjamin and Shah, Harshay and Georgiev, Kristian and M\k{a}dry, Aleksander},
 booktitle = {Advances in Neural Information Processing Systems},
 editor = {A. Globerson and L. Mackey and D. Belgrave and A. Fan and U. Paquet and J. Tomczak and C. Zhang},
 pages = {95764--95807},
 publisher = {Curran Associates, Inc.},
 title = {ContextCite: Attributing Model Generation to Context},
 url = {https://proceedings.neurips.cc/paper_files/paper/2024/file/adbea136219b64db96a9941e4249a857-Paper-Conference.pdf},
 volume = {37},
 year = {2024}
}

@inproceedings{zijian-etal-2024-unveiling,
  author       = {Zijian Feng and
                  Hanzhang Zhou and
                  Zixiao Zhu and
                  Junlang Qian and
                  Kezhi Mao},
  title        = {Unveiling and Manipulating Prompt Influence in Large Language Models},
  booktitle    = {The Twelfth International Conference on Learning Representations,
                  {ICLR} 2024, Vienna, Austria, May 7-11, 2024},
  publisher    = {OpenReview.net},
  year         = {2024},
  url          = {https://openreview.net/forum?id=ap1ByuwQrX},
  bibsource    = {dblp computer science bibliography, https://dblp.org}
}

@article{hang-etal-2024-advancing,
  author       = {Hang Zhao and
                  Qile P. Chen and
                  Yijing Barry Zhang and
                  Gang Yang},
  title        = {Advancing Single- and Multi-task Text Classification through Large
                  Language Model Fine-tuning},
  journal      = {CoRR},
  volume       = {abs/2412.08587},
  year         = {2024},
  url          = {https://doi.org/10.48550/arXiv.2412.08587},
  doi          = {10.48550/ARXIV.2412.08587},
  eprinttype    = {arXiv},
  eprint       = {2412.08587},
  bibsource    = {dblp computer science bibliography, https://dblp.org}
}

@inproceedings{wei-jie-etal-2024-plausible,
    title = "Plausible Extractive Rationalization through Semi-Supervised Entailment Signal",
    author = "Wei Jie, Yeo  and
      Satapathy, Ranjan  and
      Cambria, Erik",
    editor = "Ku, Lun-Wei  and
      Martins, Andre  and
      Srikumar, Vivek",
    booktitle = "Findings of the Association for Computational Linguistics: ACL 2024",
    month = aug,
    year = "2024",
    address = "Bangkok, Thailand",
    publisher = "Association for Computational Linguistics",
    url = "https://aclanthology.org/2024.findings-acl.307/",
    doi = "10.18653/v1/2024.findings-acl.307",
    pages = "5182--5192"
}

@inproceedings{jiang-etal-2024-mare,
    title = "{MARE}: Multi-Aspect Rationale Extractor on Unsupervised Rationale Extraction",
    author = "Jiang, Han  and
      Duan, Junwen  and
      Qu, Zhe  and
      Wang, Jianxin",
    editor = "Al-Onaizan, Yaser  and
      Bansal, Mohit  and
      Chen, Yun-Nung",
    booktitle = "Proceedings of the 2024 Conference on Empirical Methods in Natural Language Processing",
    month = nov,
    year = "2024",
    address = "Miami, Florida, USA",
    publisher = "Association for Computational Linguistics",
    url = "https://aclanthology.org/2024.emnlp-main.655/",
    doi = "10.18653/v1/2024.emnlp-main.655",
    pages = "11734--11745"
}

@inproceedings{madsen-etal-2024-self,
    title = "Are self-explanations from Large Language Models faithful?",
    author = "Madsen, Andreas  and
      Chandar, Sarath  and
      Reddy, Siva",
    editor = "Ku, Lun-Wei  and
      Martins, Andre  and
      Srikumar, Vivek",
    booktitle = "Findings of the Association for Computational Linguistics: ACL 2024",
    month = aug,
    year = "2024",
    address = "Bangkok, Thailand",
    publisher = "Association for Computational Linguistics",
    url = "https://aclanthology.org/2024.findings-acl.19/",
    doi = "10.18653/v1/2024.findings-acl.19",
    pages = "295--337"
}

@inproceedings{yin-neubig-2022-interpreting,
    title = "Interpreting Language Models with Contrastive Explanations",
    author = "Yin, Kayo  and
      Neubig, Graham",
    editor = "Goldberg, Yoav  and
      Kozareva, Zornitsa  and
      Zhang, Yue",
    booktitle = "Proceedings of the 2022 Conference on Empirical Methods in Natural Language Processing",
    month = dec,
    year = "2022",
    address = "Abu Dhabi, United Arab Emirates",
    publisher = "Association for Computational Linguistics",
    url = "https://aclanthology.org/2022.emnlp-main.14/",
    doi = "10.18653/v1/2022.emnlp-main.14",
    pages = "184--198"
}

@inproceedings{deiseroth-etal-2023-atman,
 author = {Deiseroth, Bj\"{o}rn and Deb, Mayukh and Weinbach, Samuel and Brack, Manuel and Schramowski, Patrick and Kersting, Kristian},
 booktitle = {Advances in Neural Information Processing Systems},
 editor = {A. Oh and T. Naumann and A. Globerson and K. Saenko and M. Hardt and S. Levine},
 pages = {63437--63460},
 publisher = {Curran Associates, Inc.},
 title = {ATMAN: Understanding Transformer Predictions Through Memory Efficient Attention Manipulation},
 url = {https://proceedings.neurips.cc/paper_files/paper/2023/file/c83bc020a020cdeb966ed10804619664-Paper-Conference.pdf},
 volume = {36},
 year = {2023}
}

@inproceedings{sahana-etal-2024-tailoring,
  author       = {Sahana Ramnath and
                  Brihi Joshi and
                  Skyler Hallinan and
                  Ximing Lu and
                  Liunian Harold Li and
                  Aaron Chan and
                  Jack Hessel and
                  Yejin Choi and
                  Xiang Ren},
  title        = {Tailoring Self-Rationalizers with Multi-Reward Distillation},
  booktitle    = {The Twelfth International Conference on Learning Representations,
                  {ICLR} 2024, Vienna, Austria, May 7-11, 2024},
  publisher    = {OpenReview.net},
  year         = {2024},
  url          = {https://openreview.net/forum?id=t8eO0CiZJV},
  bibsource    = {dblp computer science bibliography, https://dblp.org}
}

@inproceedings{wang-etal-2025-cross,
    title = "Cross-Refine: Improving Natural Language Explanation Generation by Learning in Tandem",
    author = {Wang, Qianli  and
      Anikina, Tatiana  and
      Feldhus, Nils  and
      Ostermann, Simon  and
      M{\"o}ller, Sebastian  and
      Schmitt, Vera},
    editor = "Rambow, Owen  and
      Wanner, Leo  and
      Apidianaki, Marianna  and
      Al-Khalifa, Hend  and
      Eugenio, Barbara Di  and
      Schockaert, Steven",
    booktitle = "Proceedings of the 31st International Conference on Computational Linguistics",
    month = jan,
    year = "2025",
    address = "Abu Dhabi, UAE",
    publisher = "Association for Computational Linguistics",
    url = "https://aclanthology.org/2025.coling-main.77/",
    pages = "1150--1167"
}

@inproceedings{yu-etal-2024-latent,
    title = "Latent Concept-based Explanation of {NLP} Models",
    author = "Yu, Xuemin  and
      Dalvi, Fahim  and
      Durrani, Nadir  and
      Nouri, Marzia  and
      Sajjad, Hassan",
    editor = "Al-Onaizan, Yaser  and
      Bansal, Mohit  and
      Chen, Yun-Nung",
    booktitle = "Proceedings of the 2024 Conference on Empirical Methods in Natural Language Processing",
    month = nov,
    year = "2024",
    address = "Miami, Florida, USA",
    publisher = "Association for Computational Linguistics",
    url = "https://aclanthology.org/2024.emnlp-main.692/",
    doi = "10.18653/v1/2024.emnlp-main.692",
    pages = "12435--12459"
}

@InProceedings{raman-etal-2024-understanding,
  title = 	 {Understanding Inter-Concept Relationships in Concept-Based Models},
  author =       {Raman, Naveen Janaki and Espinosa Zarlenga, Mateo and Jamnik, Mateja},
  booktitle = 	 {Proceedings of the 41st International Conference on Machine Learning},
  pages = 	 {42009--42025},
  year = 	 {2024},
  editor = 	 {Salakhutdinov, Ruslan and Kolter, Zico and Heller, Katherine and Weller, Adrian and Oliver, Nuria and Scarlett, Jonathan and Berkenkamp, Felix},
  volume = 	 {235},
  series = 	 {Proceedings of Machine Learning Research},
  month = 	 {21--27 Jul},
  publisher =    {PMLR},
  url = 	 {https://proceedings.mlr.press/v235/raman24a.html}
}

@inproceedings{punyajoy-etal-2023-rationale,
  author       = {Punyajoy Saha and
                  Divyanshu Sheth and
                  Kushal Kedia and
                  Binny Mathew and
                  Animesh Mukherjee},
  editor       = {Kobi Gal and
                  Ann Now{\'{e}} and
                  Grzegorz J. Nalepa and
                  Roy Fairstein and
                  Roxana Radulescu},
  title        = {Rationale-Guided Few-Shot Classification to Detect Abusive Language},
  booktitle    = {{ECAI} 2023 - 26th European Conference on Artificial Intelligence,
                  September 30 - October 4, 2023, Krak{\'{o}}w, Poland - Including
                  12th Conference on Prestigious Applications of Intelligent Systems
                  {(PAIS} 2023)},
  series       = {Frontiers in Artificial Intelligence and Applications},
  volume       = {372},
  pages        = {2041--2048},
  publisher    = {{IOS} Press},
  year         = {2023},
  url          = {https://doi.org/10.3233/FAIA230497},
  doi          = {10.3233/FAIA230497},
  bibsource    = {dblp computer science bibliography, https://dblp.org}
}

@inproceedings{modarressi-etal-2022-globenc,
    title = "{G}lob{E}nc: Quantifying Global Token Attribution by Incorporating the Whole Encoder Layer in Transformers",
    author = "Modarressi, Ali  and
      Fayyaz, Mohsen  and
      Yaghoobzadeh, Yadollah  and
      Pilehvar, Mohammad Taher",
    editor = "Carpuat, Marine  and
      de Marneffe, Marie-Catherine  and
      Meza Ruiz, Ivan Vladimir",
    booktitle = "Proceedings of the 2022 Conference of the North American Chapter of the Association for Computational Linguistics: Human Language Technologies",
    month = jul,
    year = "2022",
    address = "Seattle, United States",
    publisher = "Association for Computational Linguistics",
    url = "https://aclanthology.org/2022.naacl-main.19/",
    doi = "10.18653/v1/2022.naacl-main.19",
    pages = "258--271"
}

@article{llm-survey-2023,
  author       = {Wayne Xin Zhao and
                  Kun Zhou and
                  Junyi Li and
                  Tianyi Tang and
                  Xiaolei Wang and
                  Yupeng Hou and
                  Yingqian Min and
                  Beichen Zhang and
                  Junjie Zhang and
                  Zican Dong and
                  Yifan Du and
                  Chen Yang and
                  Yushuo Chen and
                  Zhipeng Chen and
                  Jinhao Jiang and
                  Ruiyang Ren and
                  Yifan Li and
                  Xinyu Tang and
                  Zikang Liu and
                  Peiyu Liu and
                  Jian{-}Yun Nie and
                  Ji{-}Rong Wen},
  title        = {A Survey of Large Language Models},
  journal      = {CoRR},
  volume       = {abs/2303.18223},
  year         = {2023},
  url          = {https://doi.org/10.48550/arXiv.2303.18223},
  doi          = {10.48550/ARXIV.2303.18223},
  eprinttype    = {arXiv},
  eprint       = {2303.18223},
  bibsource    = {dblp computer science bibliography, https://dblp.org}
}

@article{llm-survey-2024,
  author       = {Shervin Minaee and
                  Tom{\'{a}}s Mikolov and
                  Narjes Nikzad and
                  Meysam Chenaghlu and
                  Richard Socher and
                  Xavier Amatriain and
                  Jianfeng Gao},
  title        = {Large Language Models: {A} Survey},
  journal      = {CoRR},
  volume       = {abs/2402.06196},
  year         = {2024},
  url          = {https://doi.org/10.48550/arXiv.2402.06196},
  doi          = {10.48550/ARXIV.2402.06196},
  eprinttype    = {arXiv},
  eprint       = {2402.06196},
  bibsource    = {dblp computer science bibliography, https://dblp.org}
}

@article{llama3,
  author       = {Abhimanyu Dubey and
                  Abhinav Jauhri and
                  Abhinav Pandey and
                  Abhishek Kadian and
                  Ahmad Al{-}Dahle and
                  Aiesha Letman and
                  Akhil Mathur and
                  Alan Schelten and
                  Amy Yang and
                  Angela Fan and
                  Anirudh Goyal and
                  Anthony Hartshorn and
                  Aobo Yang and
                  Archi Mitra and
                  Archie Sravankumar and
                  Artem Korenev and
                  Arthur Hinsvark and
                  Arun Rao and
                  Aston Zhang and
                  Aur{\'{e}}lien Rodriguez and
                  Austen Gregerson and
                  Ava Spataru and
                  Baptiste Rozi{\`{e}}re and
                  Bethany Biron and
                  Binh Tang and
                  Bobbie Chern and
                  Charlotte Caucheteux and
                  Chaya Nayak and
                  Chloe Bi and
                  Chris Marra and
                  Chris McConnell and
                  Christian Keller and
                  Christophe Touret and
                  Chunyang Wu and
                  Corinne Wong and
                  Cristian Canton Ferrer and
                  Cyrus Nikolaidis and
                  Damien Allonsius and
                  Daniel Song and
                  Danielle Pintz and
                  Danny Livshits and
                  David Esiobu and
                  Dhruv Choudhary and
                  Dhruv Mahajan and
                  Diego Garcia{-}Olano and
                  Diego Perino and
                  Dieuwke Hupkes and
                  Egor Lakomkin and
                  Ehab AlBadawy and
                  Elina Lobanova and
                  Emily Dinan and
                  Eric Michael Smith and
                  Filip Radenovic and
                  Frank Zhang and
                  Gabriel Synnaeve and
                  Gabrielle Lee and
                  Georgia Lewis Anderson and
                  Graeme Nail and
                  Gr{\'{e}}goire Mialon and
                  Guan Pang and
                  Guillem Cucurell and
                  Hailey Nguyen and
                  Hannah Korevaar and
                  Hu Xu and
                  Hugo Touvron and
                  Iliyan Zarov and
                  Imanol Arrieta Ibarra and
                  Isabel M. Kloumann and
                  Ishan Misra and
                  Ivan Evtimov and
                  Jade Copet and
                  Jaewon Lee and
                  Jan Geffert and
                  Jana Vranes and
                  Jason Park and
                  Jay Mahadeokar and
                  Jeet Shah and
                  Jelmer van der Linde and
                  Jennifer Billock and
                  Jenny Hong and
                  Jenya Lee and
                  Jeremy Fu and
                  Jianfeng Chi and
                  Jianyu Huang and
                  Jiawen Liu and
                  Jie Wang and
                  Jiecao Yu and
                  Joanna Bitton and
                  Joe Spisak and
                  Jongsoo Park and
                  Joseph Rocca and
                  Joshua Johnstun and
                  Joshua Saxe and
                  Junteng Jia and
                  Kalyan Vasuden Alwala and
                  Kartikeya Upasani and
                  Kate Plawiak and
                  Ke Li and
                  Kenneth Heafield and
                  Kevin Stone and
                  et al.},
  title        = {The Llama 3 Herd of Models},
  journal      = {CoRR},
  volume       = {abs/2407.21783},
  year         = {2024},
  url          = {https://doi.org/10.48550/arXiv.2407.21783},
  doi          = {10.48550/ARXIV.2407.21783},
  eprinttype    = {arXiv},
  eprint       = {2407.21783},
  bibsource    = {dblp computer science bibliography, https://dblp.org}
}

@misc {fahamu_2022,
    author       = { {Brian Muhia} },
    title        = { ioi (Revision 223da8b) },
    year         = 2022,
    url          = { https://huggingface.co/datasets/fahamu/ioi },
    doi          = { 10.57967/hf/0142 },
    publisher    = { Hugging Face }
}

@article{warstadt2020blimp,
  title={BLiMP: The benchmark of linguistic minimal pairs for English},
  author={Warstadt, Alex and Parrish, Alicia and Liu, Haokun and Mohananey, Anhad and Peng, Wei and Wang, Sheng-Fu and Bowman, Samuel R},
  journal={Transactions of the Association for Computational Linguistics},
  volume={8},
  pages={377--392},
  year={2020},
  publisher={MIT Press One Rogers Street, Cambridge, MA 02142-1209, USA journals-info~…}
}

@inproceedings{dare,
 author = {Yue, Linan and Liu, Qi and Du, Yichao and An, Yanqing and Wang, Li and Chen, Enhong},
 booktitle = {Advances in Neural Information Processing Systems},
 editor = {S. Koyejo and S. Mohamed and A. Agarwal and D. Belgrave and K. Cho and A. Oh},
 pages = {26603--26617},
 publisher = {Curran Associates, Inc.},
 title = {DARE: Disentanglement-Augmented Rationale Extraction},
 url = {https://proceedings.neurips.cc/paper_files/paper/2022/file/a9a67d9309a28372dde3de2a1c837390-Paper-Conference.pdf},
 volume = {35},
 year = {2022}
}

@article{neobert,
  author       = {Lola Le Breton and
                  Quentin Fournier and
                  Mariam El Mezouar and
                  Sarath Chandar},
  title        = {NeoBERT: {A} Next-Generation {BERT}},
  journal      = {CoRR},
  volume       = {abs/2502.19587},
  year         = {2025},
  url          = {https://doi.org/10.48550/arXiv.2502.19587},
  doi          = {10.48550/ARXIV.2502.19587},
  eprinttype    = {arXiv},
  eprint       = {2502.19587},
  bibsource    = {dblp computer science bibliography, https://dblp.org}
}

@article{modernbert,
  author       = {Benjamin Warner and
                  Antoine Chaffin and
                  Benjamin Clavi{\'{e}} and
                  Orion Weller and
                  Oskar Hallstr{\"{o}}m and
                  Said Taghadouini and
                  Alexis Gallagher and
                  Raja Biswas and
                  Faisal Ladhak and
                  Tom Aarsen and
                  Nathan Cooper and
                  Griffin Adams and
                  Jeremy Howard and
                  Iacopo Poli},
  title        = {Smarter, Better, Faster, Longer: {A} Modern Bidirectional Encoder
                  for Fast, Memory Efficient, and Long Context Finetuning and Inference},
  journal      = {CoRR},
  volume       = {abs/2412.13663},
  year         = {2024},
  url          = {https://doi.org/10.48550/arXiv.2412.13663},
  doi          = {10.48550/ARXIV.2412.13663},
  eprinttype    = {arXiv},
  eprint       = {2412.13663},
  bibsource    = {dblp computer science bibliography, https://dblp.org}
}

@misc{Reason-ModernColBERT,
title={Reason-ModernColBERT},
author={Chaffin, Antoine},
url={https://huggingface.co/lightonai/Reason-ModernColBERT},
year={2025}
}

@article{bridge-wang-2025,
  author       = {Yifan Wang and
                  Mayank Jobanputra and
                  Ji{-}Ung Lee and
                  Soyoung Oh and
                  Isabel Valera and
                  Vera Demberg},
  title        = {Bridging Fairness and Explainability: Can Input-Based Explanations
                  Promote Fairness in Hate Speech Detection?},
  journal      = {CoRR},
  volume       = {abs/2509.22291},
  year         = {2025},
  url          = {https://doi.org/10.48550/arXiv.2509.22291},
  doi          = {10.48550/ARXIV.2509.22291},
  eprinttype    = {arXiv},
  eprint       = {2509.22291},
  bibsource    = {dblp computer science bibliography, https://dblp.org}
}

@inproceedings{wang-demberg-2024-rsa,
    title = "{RSA}-Control: A Pragmatics-Grounded Lightweight Controllable Text Generation Framework",
    author = "Wang, Yifan  and
      Demberg, Vera",
    editor = "Al-Onaizan, Yaser  and
      Bansal, Mohit  and
      Chen, Yun-Nung",
    booktitle = "Proceedings of the 2024 Conference on Empirical Methods in Natural Language Processing",
    month = nov,
    year = "2024",
    address = "Miami, Florida, USA",
    publisher = "Association for Computational Linguistics",
    url = "https://aclanthology.org/2024.emnlp-main.318/",
    doi = "10.18653/v1/2024.emnlp-main.318",
    pages = "5561--5582"
}

@inproceedings{wang-demberg-2024-parameter,
    title = "A Parameter-Efficient Multi-Objective Approach to Mitigate Stereotypical Bias in Language Models",
    author = "Wang, Yifan  and
      Demberg, Vera",
    editor = "Fale{\'n}ska, Agnieszka  and
      Basta, Christine  and
      Costa-juss{\`a}, Marta  and
      Goldfarb-Tarrant, Seraphina  and
      Nozza, Debora",
    booktitle = "Proceedings of the 5th Workshop on Gender Bias in Natural Language Processing (GeBNLP)",
    month = aug,
    year = "2024",
    address = "Bangkok, Thailand",
    publisher = "Association for Computational Linguistics",
    url = "https://aclanthology.org/2024.gebnlp-1.1/",
    doi = "10.18653/v1/2024.gebnlp-1.1",
    pages = "1--19"
}

@inproceedings{ye-etal-2025-input,
    title = "Can Input Attributions Explain Inductive Reasoning in In-Context Learning?",
    author = "Ye, Mengyu  and
      Kuribayashi, Tatsuki  and
      Kobayashi, Goro  and
      Suzuki, Jun",
    editor = "Che, Wanxiang  and
      Nabende, Joyce  and
      Shutova, Ekaterina  and
      Pilehvar, Mohammad Taher",
    booktitle = "Findings of the Association for Computational Linguistics: ACL 2025",
    month = jul,
    year = "2025",
    address = "Vienna, Austria",
    publisher = "Association for Computational Linguistics",
    url = "https://aclanthology.org/2025.findings-acl.1092/",
    doi = "10.18653/v1/2025.findings-acl.1092",
    pages = "21199--21225",
    ISBN = "979-8-89176-256-5"
}

@inproceedings{balkir-etal-2022-challenges,
    title = "Challenges in Applying Explainability Methods to Improve the Fairness of {NLP} Models",
    author = "Balkir, Esma  and
      Kiritchenko, Svetlana  and
      Nejadgholi, Isar  and
      Fraser, Kathleen",
    editor = "Verma, Apurv  and
      Pruksachatkun, Yada  and
      Chang, Kai-Wei  and
      Galstyan, Aram  and
      Dhamala, Jwala  and
      Cao, Yang Trista",
    booktitle = "Proceedings of the 2nd Workshop on Trustworthy Natural Language Processing (TrustNLP 2022)",
    month = jul,
    year = "2022",
    address = "Seattle, U.S.A.",
    publisher = "Association for Computational Linguistics",
    url = "https://aclanthology.org/2022.trustnlp-1.8/",
    doi = "10.18653/v1/2022.trustnlp-1.8",
    pages = "80--92"
}

@inproceedings{abbasi-etal-2025-normxlogit,
    title = "{N}orm{XL}ogit: The Head-on-Top Never Lies",
    author = "Abbasi, Sina  and
      Modarres, Mohammad Reza  and
      Pilehvar, Mohammad Taher",
    editor = "Christodoulopoulos, Christos  and
      Chakraborty, Tanmoy  and
      Rose, Carolyn  and
      Peng, Violet",
    booktitle = "Proceedings of the 2025 Conference on Empirical Methods in Natural Language Processing",
    month = nov,
    year = "2025",
    address = "Suzhou, China",
    publisher = "Association for Computational Linguistics",
    url = "https://aclanthology.org/2025.emnlp-main.1769/",
    doi = "10.18653/v1/2025.emnlp-main.1769",
    pages = "34914--34935",
    ISBN = "979-8-89176-332-6"
}
\bibliographystyle{tmlr}

\appendix
\section{Terminology}
\label{appendix:definitions}

To ensure clarity, we define key terms used in this work as follows:

\begin{itemize}
    \item \textbf{Faithfulness} The extent to which an explanation accurately reflects the model's actual reasoning process~\citep{jacovi-goldberg-2020-towards}. A faithful explanation should directly correspond to the internal mechanisms that led to the model's prediction.
    \item \textbf{Human Interpretability} The ease with which a person can understand the model's reasoning from the explanation~\citep{DBLP:journals/corr/abs-1902-00006}. A highly interpretable explanation should be clear, concise, and focused on relevant information while avoiding unnecessary or distracting information. However, an explanation that is easy for humans to interpret may not necessarily reflect the model's actual reasoning process or align with human reasoning patterns.
    \item \textbf{Human Agreement} The degree to which a model's explanation aligns with the reasoning a human would use for the same prediction. A high-agreement explanation should follow intuitive, logical reasoning patterns similar to human decision-making.
    \item \textbf{Explainability} The extent to which a model's computations can be faithfully explained and its learned patterns are understandable to humans. A highly explainable model should yield explanations that are both faithful to its actual reasoning process and interpretable to humans.
\end{itemize}

\section{Notation}
\label{appendix:notation}
In this paper, we use lowercase letters for scalars (e.g., $\text{b}$), bold lowercase letters for vectors (e.g., $\mathbf{w}$, $\mathbf{x}$), and bold uppercase letters ($\mathbf{W}$) for matrices.A special case is the alignment pressure parameter, denoted by the non-bold uppercase letter $\text{B}$, to distinguish it from the bias term $\text{b}$ in linear layers. We use bold uppercase letters $\mathbf{X}$ and $\mathbf{A}$ to denote a sequence of model inputs or hidden state activations. In \S~\ref{sec:methodology}, we use $\mathbf{x}$ to denote the input when a function is applied to each element of the input sequence separately. In contrast, we use $\mathbf{X}$ or $\mathbf{A}$ when the function involves interactions between elements, such as in the attention mechanism.

\section{Dynamic Linear Representation of Model Components}
\label{sec:dynamic linear components}
Here we describe how each model component in transformers can function as or be converted to a bias-free, dynamic linear module in B-cos LMs.

\paragraph{B-cos Layers}

B-cos layers are designed as bias-free, dynamic linear modules with a dynamic linear weight matrix $\mathbf{W(x)=|\text{cos}(x,\hat{W})|^\text{B-1}\otimes \hat{W}}$. Here, $\mathbf{\otimes}$ scales the rows of the matrix $\mathbf{\hat{W}}$ to its right by the scalar entries of the vector to its left.

\paragraph{Non-linear Activation Functions}

In transformer models, non-linearity is typically introduced using (approximately) piecewise linear activation functions, such as ReLU \citep{DBLP:conf/icml/NairH10} and GELU \citep{DBLP:journals/corr/HendrycksG16}. These functions can be easily interpreted as linear transformations with input-dependent weights. For example, $\text{GELU}(\mathbf{x})=\mathbf{x} \times (0.5 + 0.5 \times  \text{erf}(\mathbf{x}/\sqrt{2}))$ can be interpreted as a linear transformation where the second term acts as the dynamic linear weight.

\paragraph{Attention Blocks}

\citet{bohle2024b} showed that attention computations can be seamlessly integrated into B-cos networks as a dynamic linear module:
\begin{equation}
\text{Att}(\mathbf{X;Q,}\mathbf{K,V})=\text{softmax}(\mathbf{X^TQ^TKX)VX} =\mathbf{A(X)VX}=\mathbf{W(X)X}
\end{equation}

For multi-head self-attention (MSA), the output can be viewed as the concatenation of the outputs from $H$ attention heads, followed by a linear projection with matrix $\mathbf{U}$:
\begin{equation}
    \text{MSA}\mathbf{(X)=U[W}_1\mathbf{(X)X, ..., W}_H\mathbf{(X)X)]}
\end{equation}
Since this operation maintains a dynamic linear structure, the multi-head attention block remains a dynamic linear module.

\paragraph{Normalization Layers}

Following~\citet{bohle2024b}, we apply bias-free normalization layers to ensure completeness of explanations:
\begin{equation}
    \star \mathbf{Norm(x, \mathcal{X}; \gamma)} = \frac{\mathbf{x-\langle \mathcal{X} \rangle _{\star}}}{\sqrt{\text{var}_{\star}\mathbf{(\mathcal{X})}}} \times \mathbf{\gamma} 
\end{equation}
where $\mathbf{\mathcal{X}}$ represents a batch or sequence of inputs and $\star$ is the dimension along which the mean $\langle \cdot \rangle$ and variance $\text{var}(\cdot)$ are computed (e.g., across the batch or layer). Unlike standard normalization, this variant omits the bias term in the affine transformation to preserve explanation completeness. If a running mean estimate is used during inference, the centering term $\mathbf{\langle \mathcal{X} \rangle _{\star}}$ is also removed. This yields a bias-free, dynamic linear transformation with weight $\sqrt{\text{var}\mathbf{^{-1}_{\star}(\mathcal{X})}} \mathbf{\times \gamma}$.

\section{Implementation Details}
\label{appendix: implementation details}

\paragraph{Fine-tuning Setups} For all PLMs used in the experiments, we use the uncased base version from huggingface~\citep{wolf-etal-2020-transformers}. 
For both conventional models and B-cos LMs, we train them for 5 epochs with 10\% linear warm-up steps on the downstream task datasets. 
The learning rates are set to 2e-5 for IMDB and HateXplain, and 3e-5 for AG News. 
All models use a batch size of 16 and a maximum sequence length of 512. 
For validation, we randomly sample half of the test set from IMDB and AG News.

\paragraph{Baselines}
For IxG and ShapSampl, we use the Captum~\citep{DBLP:journals/corr/abs-2009-07896} implementations.\footnote{\url{https://captum.ai/api/}} We implement the Attention method ourselves, and LIME is sourced from the lit library\footnote{\url{https://github.com/PAIR-code/lit}}. For DecompX\footnote{\url{https://github.com/mohsenfayyaz/DecompX}} and SIG\footnote{\url{https://github.com/josephenguehard/time_interpret}}, we use their official implementations with default configurations. The number of samples is set to 25 for ShapSampl and 3,000 for LIME, with [MASK] as the baseline token. For all explanation methods at the embedding level, model predictions are attributed to the combined sum of word, position, and token type embeddings (if applicable). In the main experiments, we compute token attribution scores by summing over all embedding dimensions, as this approach demonstrates better faithfulness results than using the L2 norm.

For Saloss models, we use the official codebase\footnote{\url{https://github.com/GChrysostomou/saloss}} with default hyperparameters to train BERT and RoBERTa on AG News, IMDB, and HateXplain. DistilBERT is not included, as it is not supported by the codebase.

In Section~\ref{sec:decoder}, we follow~\citet{ferrando-etal-2023-explaining} to generate contrastive explanations that highlight why the models predicts the target token instead of the foil token. For Occlusion explanations, we use the [PAD] token to perform occlusion, instead of a zero vector as done in~\citet{yin-neubig-2022-interpreting} and~\citet{ferrando-etal-2023-explaining}. Using zero vectors distorts the input distribution and, in generative settings, can influence predictions differently depending on position. To avoid such positional effects, we instead occlude using the in-distribution embedding of the [PAD] token.

\paragraph{SeqPG Examples} When constructing examples for SeqPG, we set the sequence length to 50 for AG News, 256 for IMDB, and 25 for HateXplain, aligning with their median lengths. Only examples longer than these thresholds are selected, and they are truncated to construct synthetic examples. Additionally, we only use examples that are correctly predicted with a minimum confidence of 75\% after truncation. For a fair comparison, we evaluate Saloss models and B-cos LMs on the same sets of examples constructed based on the predictions of the corresponding conventional models.

\paragraph{Automatic Evaluation Setups}

For task performance evaluation, we use the complete test set for each task. For faithfulness evaluation, we conduct perturbation-based evaluations on 2000 test examples and SeqPG on 500 test examples for AG News and IMDB. For HateXplain, we use the full test set for perturbation-based evaluation (1,924 examples) and construct 269, 310, and 308 SeqPG examples from it using BERT, DistilBERT, and RoBERTa, respectively. In the perturbation-based evaluation, the [CLS] token is never perturbed because it is used directly to make predictions.

\paragraph{B-cos Pre-training}
For B-cos pre-training in \S~\ref{sec: different setups}, we set B=1.25 and further pre-train the model on 25,000 sentences from the Wikipedia dataset\footnote{\url{https://huggingface.co/datasets/wikimedia/wikipedia}} using masked language modeling loss with a learning rate of 1e-4 and a 15\% masking ratio. We do not B-cosify the language head, as its parameters are tied with the embedding layer. Pre-training uses the standard cross-entropy loss rather than binary cross-entropy loss, since the unnormalized language head weights could otherwise grow arbitrarily large to minimize the loss.

\paragraph{Decoder-only Models B-cosification}
We use the GPT-2 small and Llama-3.2-1B models from huggingface.
As with the encoder-based models, we do not B-cosify the language head and use cross-entropy loss for GPT-2 and Llama-3.2 training. 
B-cos adaptations are first applied and both models are then trained on 500,000 and 4,000,000 sentences, respectively, from the OpenWebText dataset for one epoch, using a learning rate of 5e-4. 
For GPT-2, we use a batch size of 16 and a maximum sequence length of 512; for Llama-3.2, we use a batch size of 128 and a sequence length of 1024. 
Perplexity is evaluated on a held-out OpenWebText subset of 10,000 sentences using a maximum sequence length of 512.

\paragraph{BLiMP Subsets}

We follow~\citet{ferrando-etal-2023-explaining} to use the following nine BLiMP subsets with corresponding IDs. 
aga: anaphor gender agreement; ana: anaphor number agreement; asp: animate subject passive; dna: determiner noun agreement 1; dnai: determiner noun agreement irregular 1; dnaa: determiner noun agreement with adj 1; dnaai: determiner noun agreement with adj irregular 1; npi: npi present 1; darn: distractor agreement relational noun. Examples of these datasets and the IOI dataset can be found in Table~\ref{tab:decoder dataset}.

\begin{table}[h]
    \centering
    \resizebox{\linewidth}{!}{
    \begin{tabular}{ccl}
    \toprule
      Dataset & ID   & Example  \\
      \midrule
      Anaphor gender agreement   & aga & \underline{Katherine} can't help \textcolor{forestgreen}{herself} / \textcolor{red}{himself}. \\
      Anaphor number agreement & ana & \underline{Susan} revealed \textcolor{forestgreen}{herself} / \textcolor{red}{themselves}. \\
      Animate subject passive & asp & Amanda was \underline{respected} by some \textcolor{forestgreen}{waitresses} / \textcolor{red}{pictures}. \\
      Determiner noun agreement 1 & dna & Raymond is selling \underline{this} \textcolor{forestgreen}{sketch} / \textcolor{red}{sketches}. \\
      Determiner noun agreement irregular 1 & dnai & Adam hadn't discussed \underline{these} \textcolor{forestgreen}{analyses} / \textcolor{red}{analysis}. \\
      Determiner noun agreement with adjectives 1 & dnaa & Rebecca was criticizing \underline{those} good \textcolor{forestgreen}{documentaries} / \textcolor{red}{documentary}. \\
      Determiner noun agreement with adjectives irregular 1 & dnaai &  Some waiters broke \underline{this} lost \textcolor{forestgreen}{foot} / \textcolor{red}{feet}. \\
      NPI present 1 & npi &  \underline{Even} Suzanne has \textcolor{forestgreen}{really} / \textcolor{red}{ever} joked around.\\
      Distractor agreement relational noun & darn & \underline{A niece of most senators} \textcolor{forestgreen}{hasn't} / \textcolor{red}{haven't} descended most slopes.\\
      \multirow{2}{*}{Indirect Object Identification} & \multirow{2}{*}{IOI} & Friends \underline{Juana} and Kristi found a mango at the bar. \\
       &  & Kristi gave it to \textcolor{forestgreen}{Juana} / \textcolor{red}{Kristi}. \\
    \bottomrule
      
    \end{tabular}
    }
    \caption{Examples from the BLiMP and IOI datasets. \textcolor{forestgreen}{Green} (\textcolor{red}{red}) indicates \textcolor{forestgreen}{target} (\textcolor{red}{foil}) predictions. Ground truth evidence for the correct continuations is \underline{underlined}.}
    \label{tab:decoder dataset}
\end{table}

\paragraph{Compute Infrastructure} 
Unless stated otherwise, all experiments are conducted on a single NVIDIA H100 GPU. 
Training one epoch of B-cos BERT takes approximately 40 minutes on AG News, 10 minutes on IMDB, and 5 minutes on HateXplain.

\section{Explanation Efficiency}\label{sec:explanation-efficiency}

\begin{wraptable}{r}{0.4\textwidth}  
  \vspace{-5mm}
    \centering
    \resizebox{0.9\linewidth}{!}{
    \tabcolsep=2pt
    \begin{tabular}{@{}lcc@{}}
    \toprule
    \textbf{Method} & \textbf{Time (s)} & \textbf{Memory (GB)}  \\
    \midrule
    ShapSampl & 37.22 & 21.95 \\
    LIME & 6.82 & 21.96 \\
    SIG & 67.46 & 29.09 \\
    DecompX & 0.76 & 48.38 \\
    B-cos & \textbf{0.08} & \textbf{2.82} \\
    \bottomrule
    \end{tabular}
    }
    \caption{Computational costs per example of generating explanations for 100 instances using an NVIDIA H100 GPU (batch size 1). 
    B-cos explanations (\textbf{bold}) are at least 9x faster and require at most $\frac{1}{8}$ of VRAM.
    }
    \label{tab:efficiency}
  \vspace{-13mm}
\end{wraptable}

Beyond improved faithfulness and human interpretability, B-cos explanations are also efficient to extract. 
Comparing their computational costs with strong post-hoc methods shows that B-cos explanations are the most efficient in both time and memory usage (Table~\ref{tab:efficiency}). Post-hoc and B-cos explanations are generated from the conventionally fine-tuned and B-cos BERT models on IMDB, respectively.

\section{SeqPG Example}
\label{appendix:seqpg example}
\begin{figure*}
    \centering
    \includegraphics[width=0.8\textwidth]{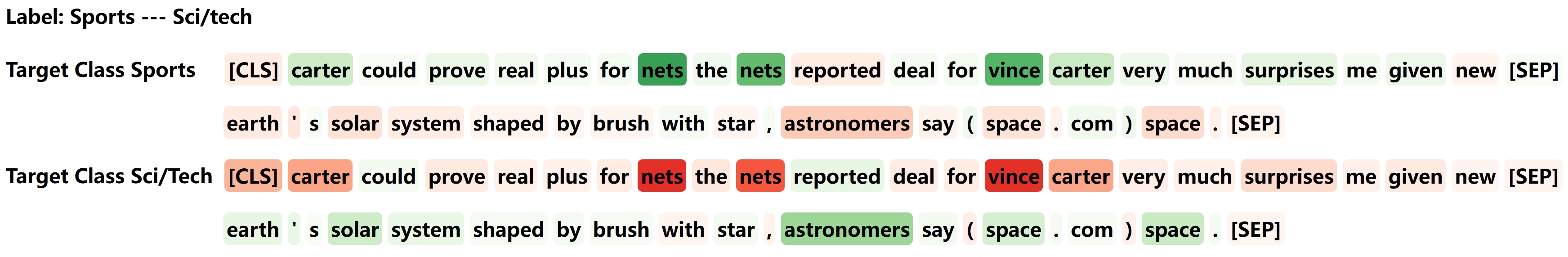}
    \caption{An example of SeqPG from AG News (using B-cos BERT). \textcolor{forestgreen}{Green} (\textcolor{red}{red}) indicates the \textcolor{forestgreen}{positive} (\textcolor{red}{negative}) impact of tokens on the prediction. The example consists of two sequences with different labels (Sports and Sci/tech), separated by the [SEP] token after the first sequence. Explanations are generated for each label, and the proportion of correctly attributed positive tokens is averaged across both labels to compute the SeqPG score for this example.}
    \label{fig:seqpg}
\end{figure*}

Figure~\ref{fig:seqpg} presents a SeqPG example from AG News using B-cos BERT. 
For better visualization, each segment is truncated to 20 tokens here instead of 50 used in the experiments. 
Unlike the hybrid document evaluation proposed by \citet{poerner-etal-2018-evaluating}, our approach explicitly controls segment length and position to ensure a fair comparison. 
Additionally, we measure the proportion of correctly assigned positive attributions rather than relying solely on the highest attribution value.

\section{Task Performance of Other B-cos LMs}
\label{appendix:accuracy other plms}

\begin{figure}[h]
    \centering
    \includegraphics[width=0.7\linewidth]{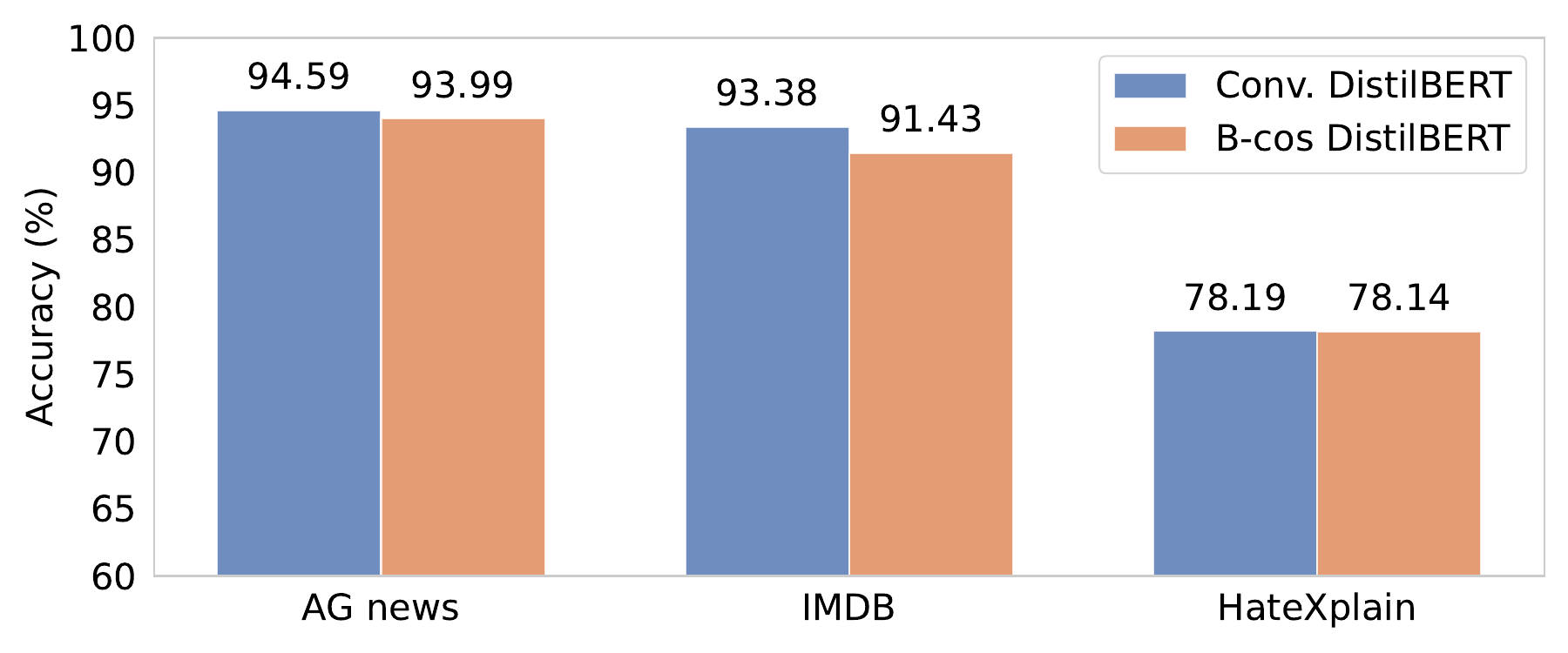}
    \caption{Mean accuracy of conventionally fine-tuned and B-cos DistilBERT models averaged over three runs. 
    B-cos models perform comparably to conventional models on most tasks.}
    \label{fig:distilbert performance}
\end{figure}

\begin{figure}[h]
    \centering
    \includegraphics[width=0.7\linewidth]{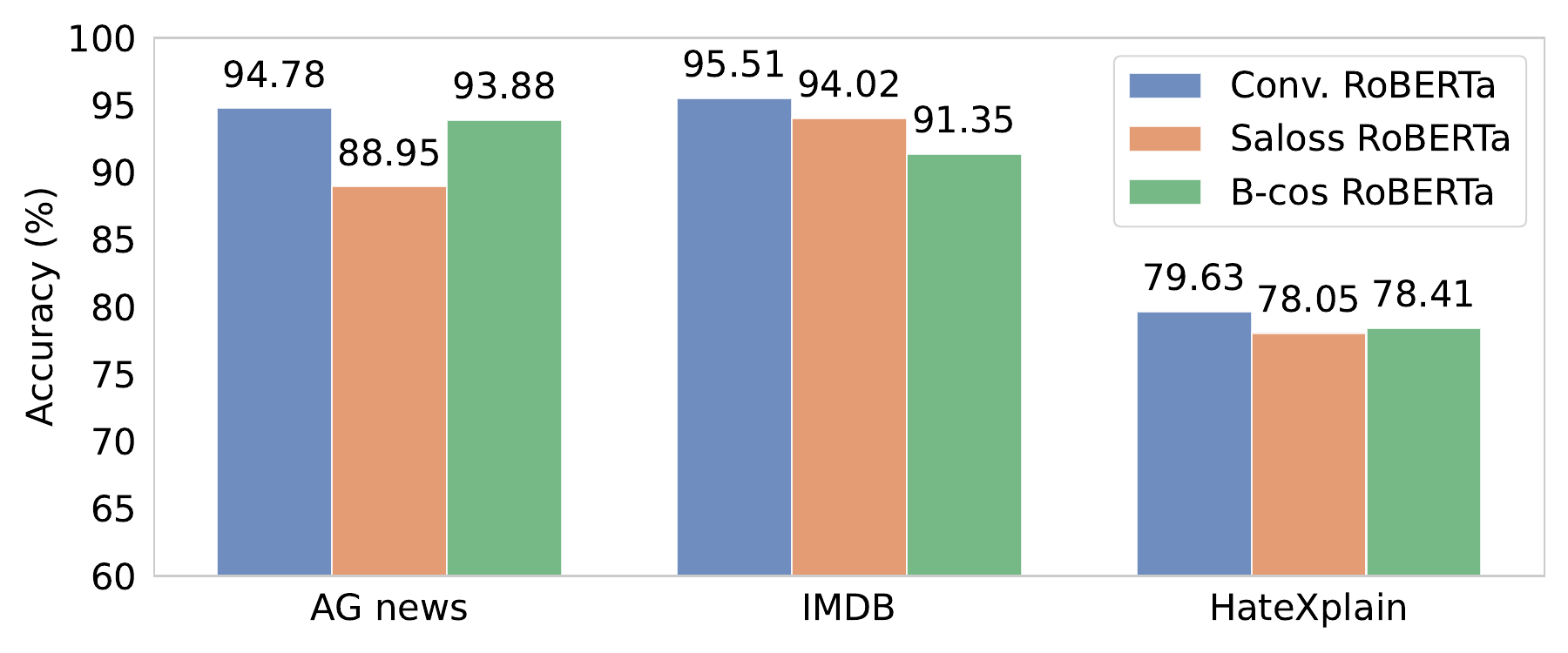}
    \caption{Mean accuracy of conventionally fine-tuned and B-cos RoBERTa models averaged over three runs. 
    B-cos models perform comparably to conventional models on most tasks.}
    \label{fig:roberta performance}
\end{figure}

Figures~\ref{fig:distilbert performance} and~\ref{fig:roberta performance} illustrate the task performance of conventional and B-cos DistilBERT and RoBERTa across datasets. 
Consistent with findings from BERT models (cf. Figure~\ref{fig:bert performance}), B-cos LMs exhibit strong performance comparable to conventionally fine-tuned models.

\section{Faithfulness Evaluation of Other B-cos LMs}
\label{appendix:other PLMs}
Tables~\ref{tab:faithfulness distilbert} and~\ref{tab:faithfulness roberta} present the faithfulness evaluation results for DistilBERT and RoBERTa. 
The findings are consistent with our main experiments (cf. Table~\ref{tab:faithfulness}), confirming that B-cos LMs produce more faithful explanations compared to post-hoc explanation methods.

\begin{table*}[ht]
  \centering
  \resizebox{\textwidth}{!}{
  \begin{tabular}{llccccccccc}
     \toprule
      \multirow{2}{*}{\textbf{Model}}&\multirow{2}{*}{\textbf{Method}} & \multicolumn{3}{c}{\textbf{AG News}} & \multicolumn{3}{c}{\textbf{IMDB}} & \multicolumn{3}{c}{\textbf{HateXplain}} \\
      \cmidrule(lr){3-5}\cmidrule(lr){6-8} \cmidrule(lr){9-11}
      & & Comp ($\uparrow$) & Suff ($\downarrow$) & SeqPG ($\uparrow$) & Comp ($\uparrow$) & Suff ($\downarrow$) & SeqPG ($\uparrow$) & Comp ($\uparrow$) & Suff ($\downarrow$) & SeqPG ($\uparrow$) \\
      \midrule
    \multirow{6}{*}{Conv. DistilBERT}&Attention & 26.36 & 5.37 & 50 & 31.62 & 10.46 & 50 & 30.56 & 14.67 & 50 \\
      &IxG & 19.29 & 6.21 & 53.71 & 23.78 & 12.38 & 49.23 & 25.13 & 18.08 & 46.60 \\
      &SIG & 30.78 & 1.63 & 67.87 & 47.16 & 5.48 & 60.66 & 41.11 & 4.23 & 58.55 \\
      &DecompX & - & - & - & - & - & - & - & - & - \\
      &ShapSampl & 52.56 & -0.56 & 82.64 & 63.29 & 2.91 & 70.27 & 48.73 & 0.87 & 64.44 \\
      &LIME & 52.59 & -0.56 & 77.64 & 58.6 & 5.12 & 61.11 & 31.61 & 12.94 & 56.49 \\

    \midrule

      B-cos DistilBERT  & B-cos & \textbf{61.93} & \textbf{-1.01} & \textbf{86.78} & \textbf{76.26} & \textbf{-1.28} & \textbf{72.68} & \textbf{57.2} & \textbf{-4.49} & \textbf{74.89} \\

    \bottomrule
  \end{tabular}
  }
  \caption{Faithfulness evaluation for conventionally fine-tuned DistilBERT and B-cos DistilBERT across three datasets. 
  The best results are in \textbf{bold}.
  We find that B-cos explanations are consistently more faithful than post-hoc explanations from both models.}
  \label{tab:faithfulness distilbert}
\end{table*}

\begin{table*}[h]
  \centering
  \resizebox{\textwidth}{!}{
  \begin{tabular}{llccccccccc}
     \toprule
      \multirow{2}{*}{\textbf{Model}}&\multirow{2}{*}{\textbf{Method}} & \multicolumn{3}{c}{\textbf{AG News}} & \multicolumn{3}{c}{\textbf{IMDB}} & \multicolumn{3}{c}{\textbf{HateXplain}} \\
      \cmidrule(lr){3-5}\cmidrule(lr){6-8} \cmidrule(lr){9-11}
      & & Comp ($\uparrow$) & Suff ($\downarrow$) & SeqPG ($\uparrow$) & Comp ($\uparrow$) & Suff ($\downarrow$) & SeqPG ($\uparrow$) & Comp ($\uparrow$) & Suff ($\downarrow$) & SeqPG ($\uparrow$) \\
      \midrule
      
      \multirow{6}{*}{Conv. RoBERTa}&Attention & 22.17 & 3.80 & 50 & 25.26 & 5.84 & 50 & 32.94 & 7.52 & 50 \\
      &IxG & 11.33 & 7.54 & 44.15 & 16.15 & 11.53 & 47.20 & 24.40 & 15.16 & 50.59 \\
      &SIG & 19.64 & 1.63 & 66.43 & 38.14 & 2.13 & 59.04 & 44.21 & -1.42 & 66.73 \\
      &DecompX & 50.00 & -0.84 & \textbf{90.38} & 49.24 & 0.65 & 72.80 & 46.94 & -1.42 & 70.16 \\
      &ShapSampl & 35.63 & -0.68 & 78.31 & 43.32 & 1.83 & 65.85 & 44.83 & -1.30 & 67.15 \\
      &LIME & 19.28 & 2.85 & 66.73 & 21.07 & 8.32 & 50.81 & 27.97 & 11.38 & 58.59 \\
    \midrule

\multirow{6}{*}{Saloss RoBERTa}&Attention & 40.69 & 2.77 & 50 & 24.51 & 4.33 & 50 & 47.04 & 7.83 & 50 \\
      &IxG & 6.19 & 27.30 & 52.46 & 10.98 & 12.30 & 47.92 & 22.78& 25.78 & 49.49 \\
      &SIG & 6.91 & 27.22 & 56.84 & 11.53 & 13.76 & 62.10 & 43.77 & 5.02 & 58.67 \\
      &DecompX & 61.46 & 0.16 & 74.20 & 65.50 & 0.10 & \textbf{74.41} & 54.94 & 2.47 & 65.63 \\
      &ShapSampl & 34.48 & 0.73 & 64.67 & 48.53 & 0.82 & 63.04 & \textbf{55.80} & 1.49 & 64.53 \\
      &LIME & 15.93 & 8.03 & 55.17 & 18.04 & 6.47 & 50.94 & 29.62 & 15.78 & 56.00 \\
      \midrule
B-cos RoBERTa & B-cos & \textbf{62.47} & \textbf{-1.18} & 86.63 & \textbf{73.87} & \textbf{-2.30} & 74.05 & 51.33 & \textbf{-5.18} & \textbf{74.01} \\
    \bottomrule
  \end{tabular}
  }
  \caption{Faithfulness evaluation for conventionally fine-tuned RoBERTa, Saloss RoBERTa and B-cos RoBERTa across three datasets. 
  The best results are in \textbf{bold}.
  We find that B-cos explanations are consistently more faithful than post-hoc explanations from both models.}
  \label{tab:faithfulness roberta}
\end{table*}

\section{Comparison to Rationale-Based Models}
\label{appendix:rationale-based}

We compare B-cos LMs to one rationale-based, explain-then-predict BERT model, RGFS-SA~\citep{punyajoy-etal-2023-rationale}\footnote{\url{https://huggingface.co/Hate-speech-CNERG/Rationale_predictor}} on HateXplain. This model leverages human rationales as additional supervision during training. As shown in Table~\ref{tab:rationale}, although the RGFS-SA model brings improvement over the conventional BERT model, it generates considerably less faithful rationales compared to B-cos explanations.

\begin{table}[h]
    \centering
    \begin{tabular}{lccccc}
    \toprule
    \textbf{Model} & \textbf{Method} & \textbf{Accuracy ($\uparrow$)} & \textbf{Comp ($\uparrow$)} & \textbf{Suff ($\downarrow$)} & \textbf{SeqPG ($\uparrow$)} \\
    \midrule
      Conv. BERT   & Attention & 80.77 &	22.64 &	13.83 &	50  \\
       RGFS-SA BERT &	Rationale &	80.09	& 36.11	& 16.54 & 50	 \\
       B-cos BERT &	B-cos &	78.64	& 59.66	& -4.89	& 77.57 \\
       \bottomrule
    \end{tabular}
    \caption{Performance of conventional, RGFS-SA and B-cos (B=1.5) BERT models on HateXplain. SeqPG is consistently 50 for rationale-based models, as their explanations are class-agnostic, similar to attention. The rationale-based RGFS-SA model generates less faithful explanations than B-cos BERT.}
    \label{tab:rationale}
\end{table}

\section{Human Evaluation Details}
\label{appendix:human evaluation}

In the human study, we select only examples shorter than 25 tokens for HateXplain and 40 tokens for AG News to improve visualization. Additionally, we replace [CLS] and [SEP] with \#\# to make the examples more understandable for lay users. Below, we provide the instructions along with a detailed description of the criteria and scoring used in our human evaluation. In our human study, 92\% of AG News examples and 80\% of HateXplain examples contain correct model predictions; in the remaining cases, explanations are supposed to support the wrong predictions.

\begin{displayquote}
\noindent \textbf{WARNING: SOME CONTENT IN THIS QUESTIONNAIRE IS HIGHLY OFFENSIVE.}

\noindent \textbf{Prerequisites:} Proficiency in English is required for this evaluation task. If you do not meet this criterion, please do not proceed.

\noindent We invite you to review 100 examples where LMs perform classification tasks and provide explanations for their predictions.
\begin{itemize}
    \item The first 50 examples come from a hate speech detection task, where the model predicts whether a text is toxic or not toxic.  
    \item The last 50 examples come from a topic classification task, where the model categorizes a text into one of four topics: sports, world, business, or sci/tech. 
\end{itemize}

\noindent For each example:
\begin{itemize}
    \item The model's prediction is shown along with four explanations justifying the prediction.
    \item The order of the explanations is randomized to prevent bias.
    \item Words highlighted in green indicate words that had a positive influence on the prediction, while words in red indicate words that had a negative influence. The intensity of the color reflects the strength of the impact.
    \item \textbf{Important:} The model's prediction may be incorrect. Your task is to evaluate the explanations based on how well they support the model's prediction, not the true labels. 
\end{itemize}

\noindent Evaluation Task:

\noindent After reviewing each example, please rate the the \textbf{human interpretability} and \textbf{human agreement} of the four explanations on a scale of 1 to 5. Refer to the definitions and rating scales provided below when making your assessments.

\paragraph{Human Interpretability:} How easily a person can \textbf{understand the model's reasoning} based on the explanation. A highly interpretable explanation should be clear and easy to follow, focus on relevant words and avoid unnecessary or distracting details.
\begin{enumerate}
    \item \textbf{Not Interpretable:} The explanation is unclear, noisy, or provides no meaningful insight.
    \item \textbf{Slightly Interpretable:} Some clues are present, but the explanation is too sparse, irrelevant, or confusing.
    \item \textbf{Moderately Interpretable:} The explanation contains useful information but is cluttered with noise or irrelevant details.
    \item \textbf{Highly Interpretable:} The explanation is mostly clear, with minimal irrelevant highlights.
    \item \textbf{Completely Interpretable:} The explanation is fully transparent, highlighting only the most relevant words, making the model's reasoning fully clear.
\end{enumerate}

\paragraph{Human Agreement:} How closely the model's explanation \textbf{aligns with the reasoning a human would use} for the same prediction. A high-agreement explanation should follow logical, intuitive reasoning and align with typical human decision-making patterns.
\begin{enumerate}
    \item \textbf{No Agreement:} The explanation contradicts human reasoning or lacks logic.
    \item \textbf{Low Agreement:} The explanation bears some resemblance to human reasoning but includes major inconsistencies.
    \item \textbf{Moderate Agreement:} The explanation partially aligns with human reasoning, yet contains notable differences.
    \item \textbf{High Agreement:} The explanation largely aligns with human reasoning, showing only minor discrepancies.
    \item \textbf{Complete Agreement:} The explanation fully matches human reasoning, following a logical and intuitive path that a human would naturally use.
\end{enumerate}
\end{displayquote}

\begin{figure}
    \centering
    \includegraphics[width=0.7\linewidth]{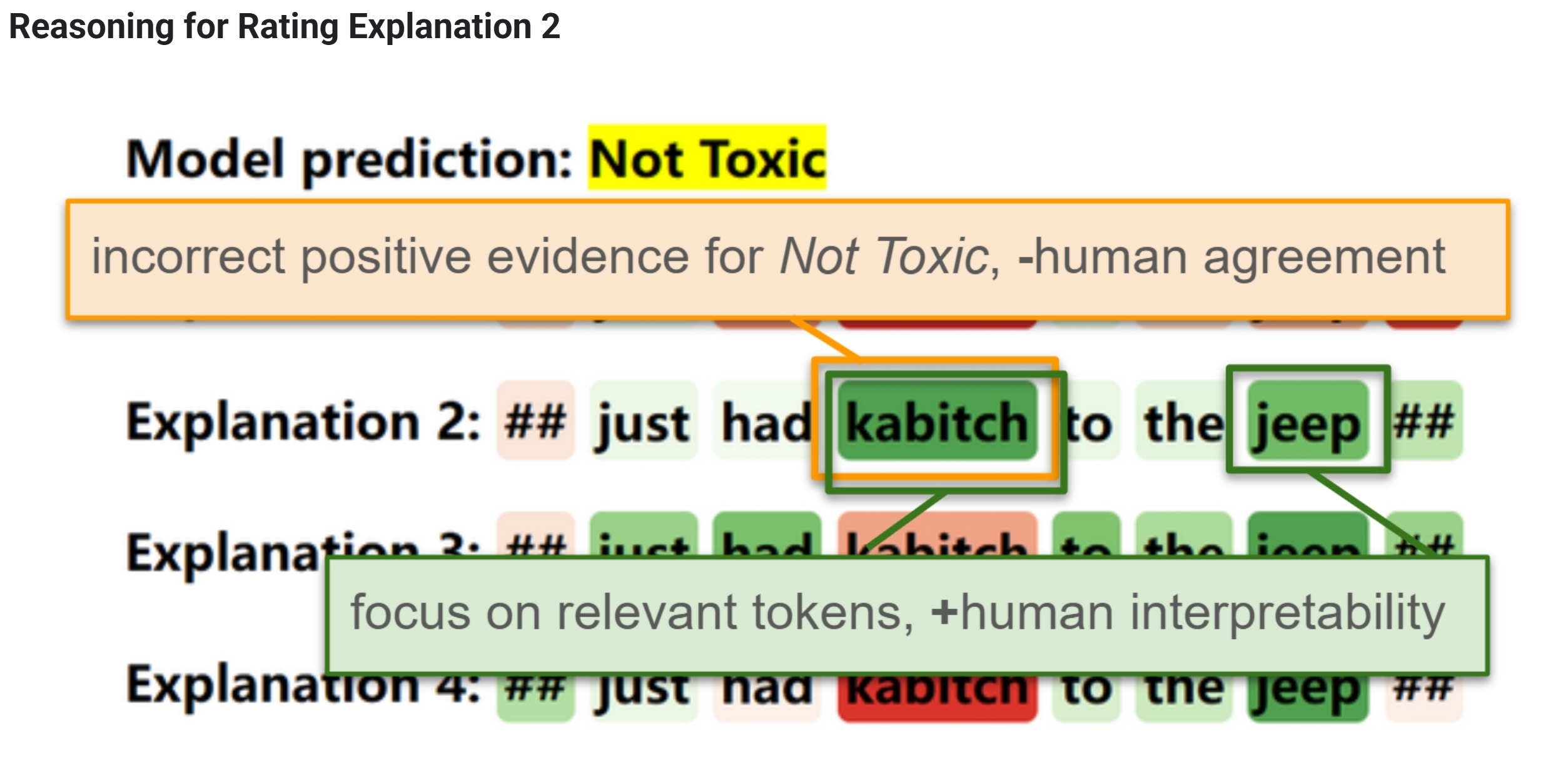}
    \caption{An example shown to participants that demonstrates how to rate explanations.}
    \label{fig:rationale}
\end{figure}

\begin{figure}[ht]
    \centering
    \includegraphics[width=0.8\linewidth]{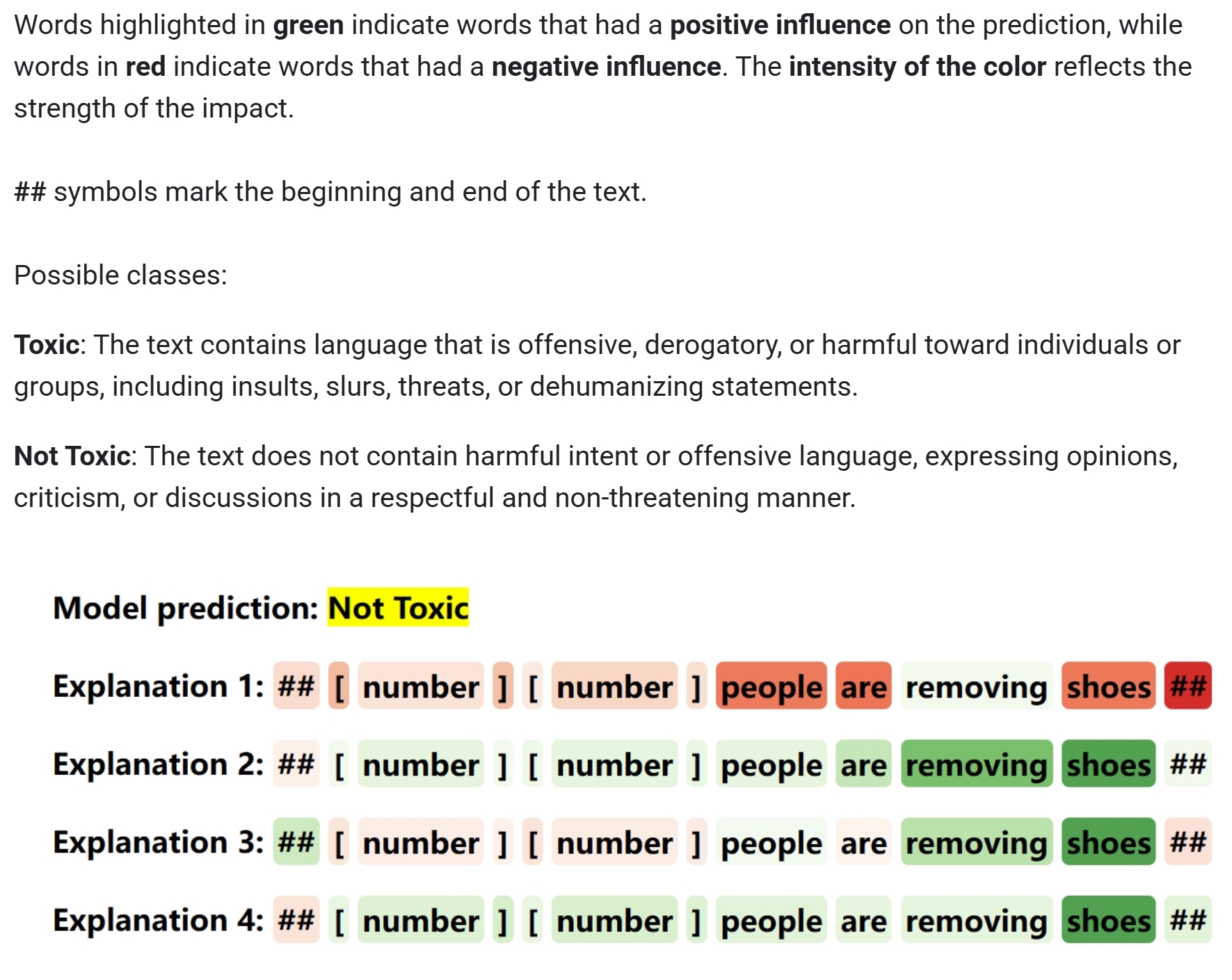}
    \caption{ An examples of a model prediction and its explanations presented to participants.}
    \label{fig:interface1}
\end{figure}

We also provide participants with examples to illustrate the reasoning behind rating explanations.
One such example is shown in Figure~\ref{fig:rationale}.
Additionally, Figure~\ref{fig:interface1} presents an example of a model prediction and its explanations as displayed to participants during the study.

\section{More Examples of B-cos Explanations}
\label{appendix:more examples}

\begin{figure*}[t]
    \centering
    \includegraphics[width=0.8\textwidth]{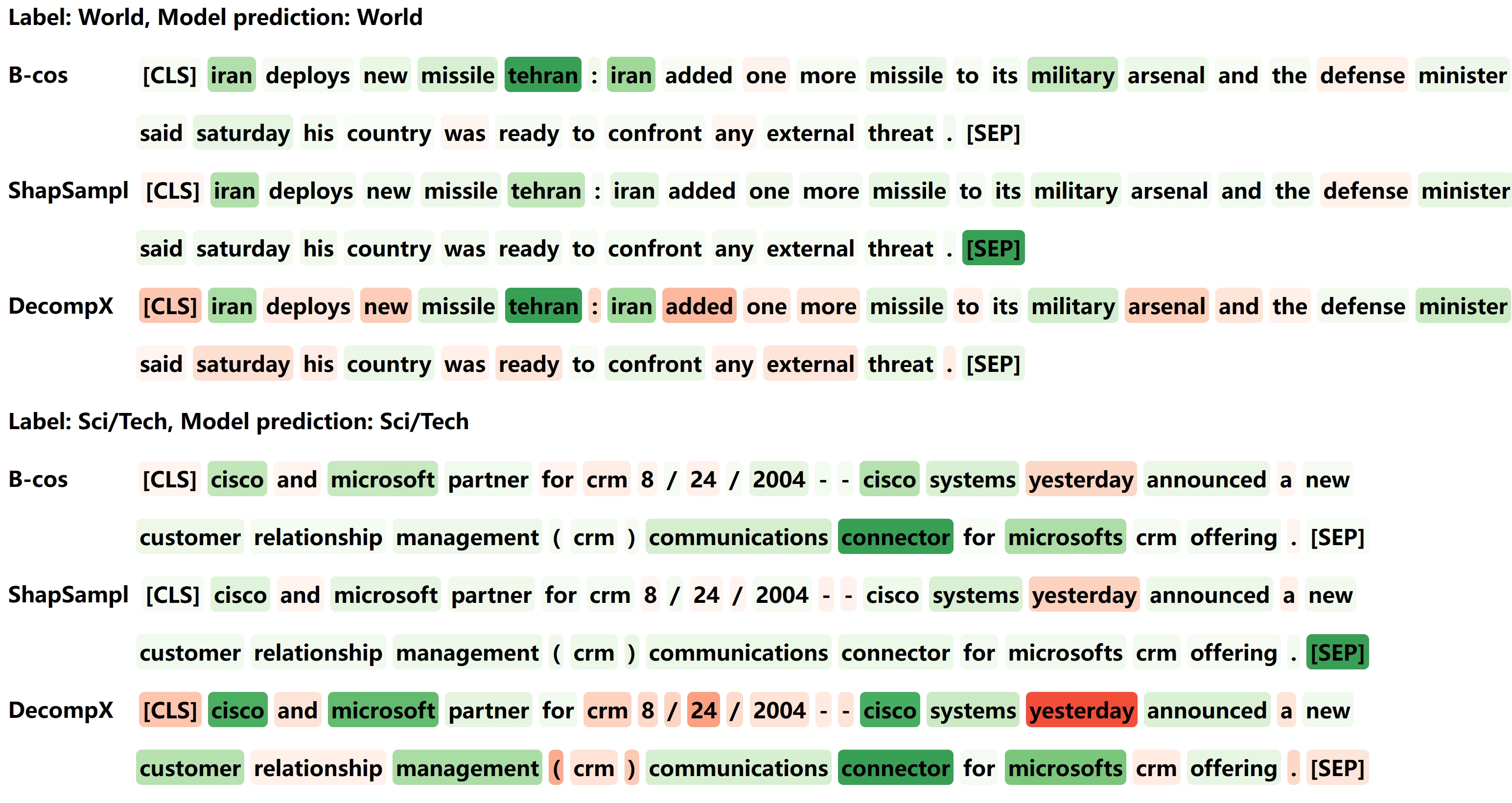}
    \caption{More examples of B-cos explanations (B-cos BERT) as well as ShapSampl and DecompX explanations (BERT) from the AG News dataset.
    \textcolor{forestgreen}{Green} (\textcolor{red}{red}) indicates the \textcolor{forestgreen}{positive} (\textcolor{red}{negative}) impact of tokens on the prediction. 
    As can be seen, the B-cos explanation highlights only relevant tokens and is more interpretable to humans.}
    \label{fig:more examples}
\end{figure*}

We provide two more examples of B-cos and other (post-hoc) explanations from AG News in Figure~\ref{fig:more examples}. Consistent with our findings in \S~\ref{sec:qualitative}, B-cos LMs provide more human interpretable explanations.

\begin{figure*}[h]
    \centering
    \includegraphics[width=0.8\textwidth]{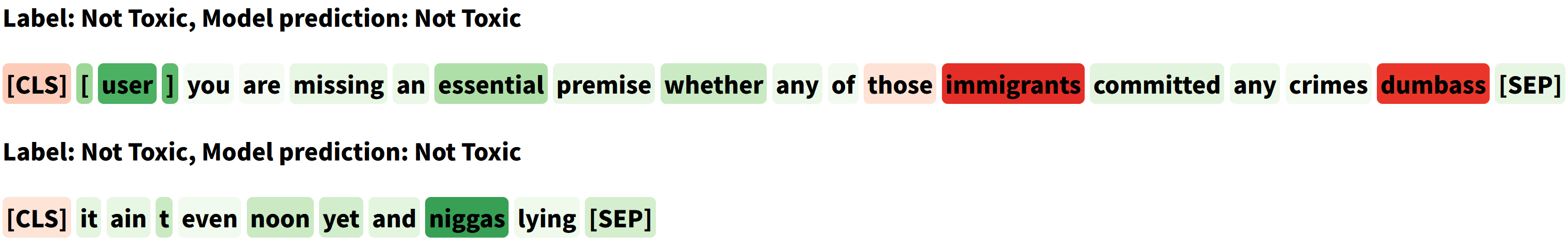}
    \caption{Two undesired B-cos explanations from B-cos BERT on HateXplain.
    \textcolor{forestgreen}{Green} (\textcolor{red}{red}) indicates the \textcolor{forestgreen}{positive} (\textcolor{red}{negative}) impact of tokens on the prediction. 
    B-cos explanations can be influenced by spurious correlations learned by the model and may occasionally misattribute the contribution of certain tokens.}
    \label{fig:failure cases}
\end{figure*}

Figure~\ref{fig:failure cases} presents two examples where B-cos explanations do not align with human expectations.
As B-cos explanations faithfully summarize model computations, they may also reveal spurious correlations from the training data and occasionally produce unreasonable rationales. 
For example, in the first case, the token "[user]" receives high attribution, likely due to recurring patterns in the data. 
Another observed error is polarity misinterpretation: in the second case, the token "niggas" is incorrectly highlighted as positive evidence for a non-toxic prediction. 
Similar types of undesired behavior have also been reported for other explanation methods.


\section{Impact of B on Input-weight Alignment}
\label{appendix:alignment}

\begin{figure}[h]
    \centering
    \includegraphics[width=0.8\linewidth]{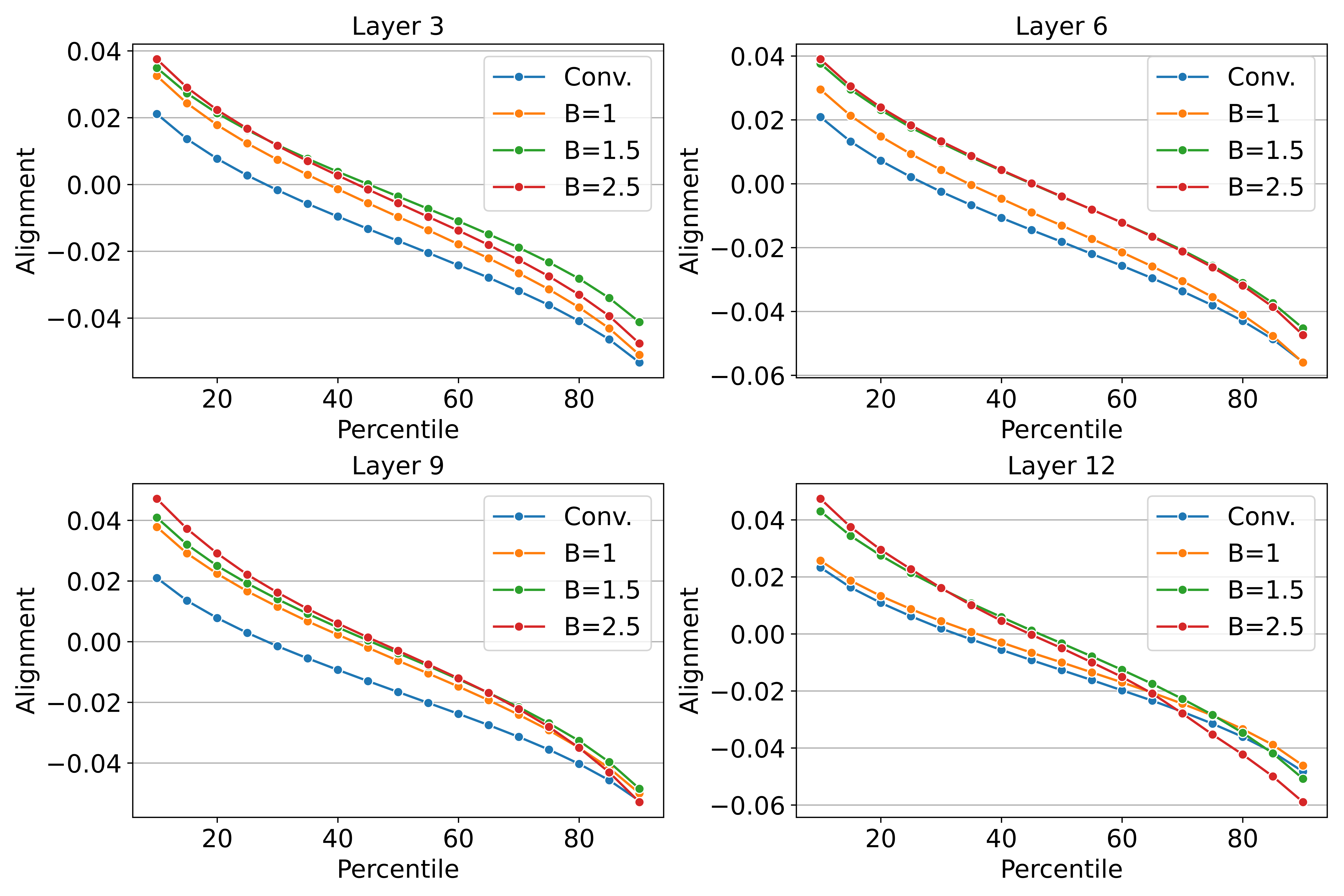}
    \caption{Percentiles of input-weight alignment in B-cos modules across selected layers of conventional and B-cos BERT models with different B values (HateXplain).}
    \label{fig:alignment}
\end{figure}

To analyze how B-cosification and alignment pressure influence the behavior of B-cos LMs, we compute the alignment (cosine similarity) between each input and its corresponding weight in B-cos modules across all layers. This analysis is performed on 100 examples from the HateXplain dataset. In Figure~\ref{fig:alignment}, we plot different percentiles of input-weight alignment for conventional and B-cos BERT models with varying B values. For better visualization, we display only the 10th to 90th percentiles.

Overall, larger B values generally lead to stronger input-weight alignment compared to smaller B and conventional models, as evidenced by the curves for B=1.5 and B=2.5 lying above those for the conventional model and B=1. However, the alignment pattern becomes more complex when comparing B=1.5 and B=2.5. Specifically, at B=2.5, the most aligned input-weight pairs exhibit higher alignment than in other models, but some pairs show very low alignment. This result may arise because certain weights are highly optimized for specific input patterns, leading to poor alignment with others, particularly in later layers where input features become more anisotropic \citep{ethayarajh-2019-contextual, li-etal-2020-sentence}. As a result, some outputs from the B-cos layers are highly negative. When these outputs are fed into GELU activation functions, their dynamic weights approach zero, making the explanations more sparse.

\section{Effects of B on Other Metrics}
\label{appendix:different b all metrics}
Table~\ref{tab:b values} presents the complete results on how B values affect task performance, explanation faithfulness and explanation entropy, as shown in Figure~\ref{fig:b analysis}. 
Similar to Comp, SeqPG scores also decline with higher alignment pressure. 
This could also be attributed to the high sparsity of explanations. 
As B increases, fewer tokens receive attribution scores that are not close to zero, and in some SeqPG examples, B-cos LMs may attribute predictions to a single segment. This can lead to numerical instability when computing the positive attribution ratio.

\begin{table}[h]
    \centering
    \begin{small}

        \tabcolsep=2pt
    \begin{tabular}{lrrrrrrr}
    \toprule
      \textbf{B}   & 1.00 & 1.25 & 1.50 & 1.75 & 2.00 & 2.25 & 2.50 \\
      \midrule
      Acc ($\uparrow$) & 78.57 & \textbf{79.23} & 78.10 & 77.41 & 77.48 & 70.44 & 73.55 \\
      Comp ($\uparrow$) & 55.09 & 58.99 & \textbf{59.64} & 59.23 & 54.44 & 35.80 & 27.11 \\
      Suff ($\downarrow$) & -4.25 & -5.71 & -5.47 & -5.84 & -6.69 & \textbf{-7.23} & -5.47 \\
      SeqPG ($\uparrow$) & 69.75 & 77.26 & \textbf{77.79} & 77.67 & 76.79 & 76.68 & 77.25 \\
      Entropy & 3.09 & 2.79 & 2.58 & 2.35 & 2.28 & 1.98 & 1.89 \\
      \bottomrule
    \end{tabular}
    
    \end{small}
    \caption{Task performance, explanation faithfulness, and explanation entropy of B-cos BERT models on HateXplain with different B values. Results are averaged over three runs. Similar to Figure~\ref{fig:b analysis}, task performance and explanation faithfulness peak around B=1.5, while explanation entropy correlates negatively with B.}
    \label{tab:b values}
\end{table}

\section{B-cos Explanations with Different B Values}
\label{appendix:examples different b}

\begin{figure*}
    \centering
    \includegraphics[width=0.8\textwidth]{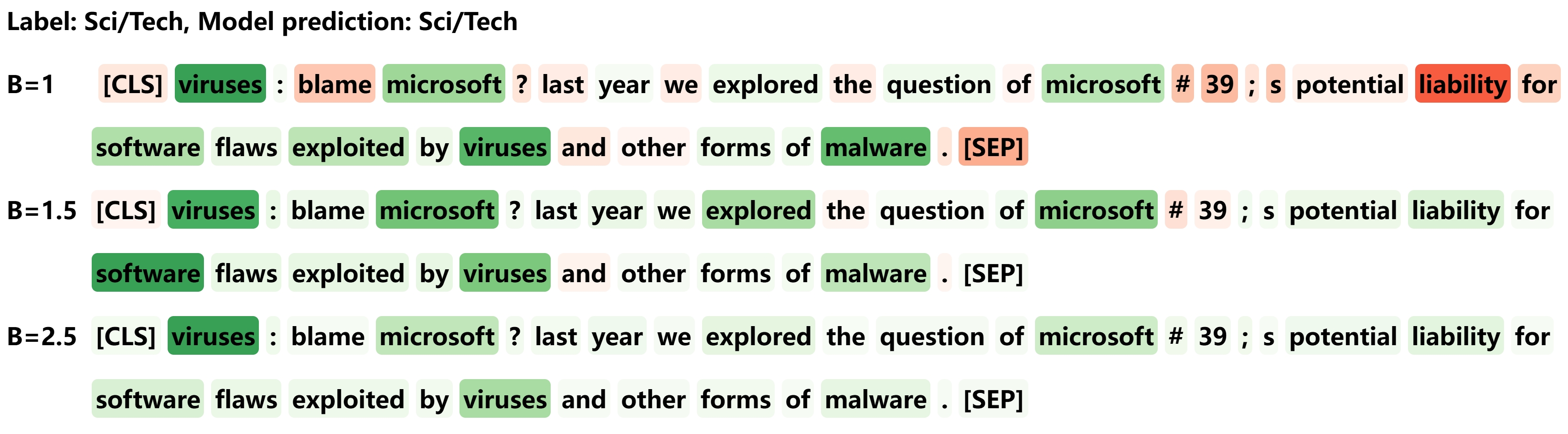}
    \caption{B-cos explanations (B-cos BERT) on AG News with different B values. \textcolor{forestgreen}{Green} (\textcolor{red}{red}) indicates the \textcolor{forestgreen}{positive} (\textcolor{red}{negative}) impact of tokens on the prediction. As B increases, B-cos LMs produce sparser explanations, with fewer tokens receiving significant attribution scores.}
    \label{fig:explanation different b}
\end{figure*}

Figure~\ref{fig:explanation different b} illustrates that with increased alignment pressure, B-cos LMs learn fewer but more task-relevant features. Consequently, they produce sparser explanations, with fewer tokens receiving significant attribution. This finding aligns with the statistics presented in \S~\ref{sec:B}.

\section{Example of Model Bias}
\label{appendix:bias}

\begin{figure*}[h]
    \centering
    \includegraphics[width=0.8\textwidth]{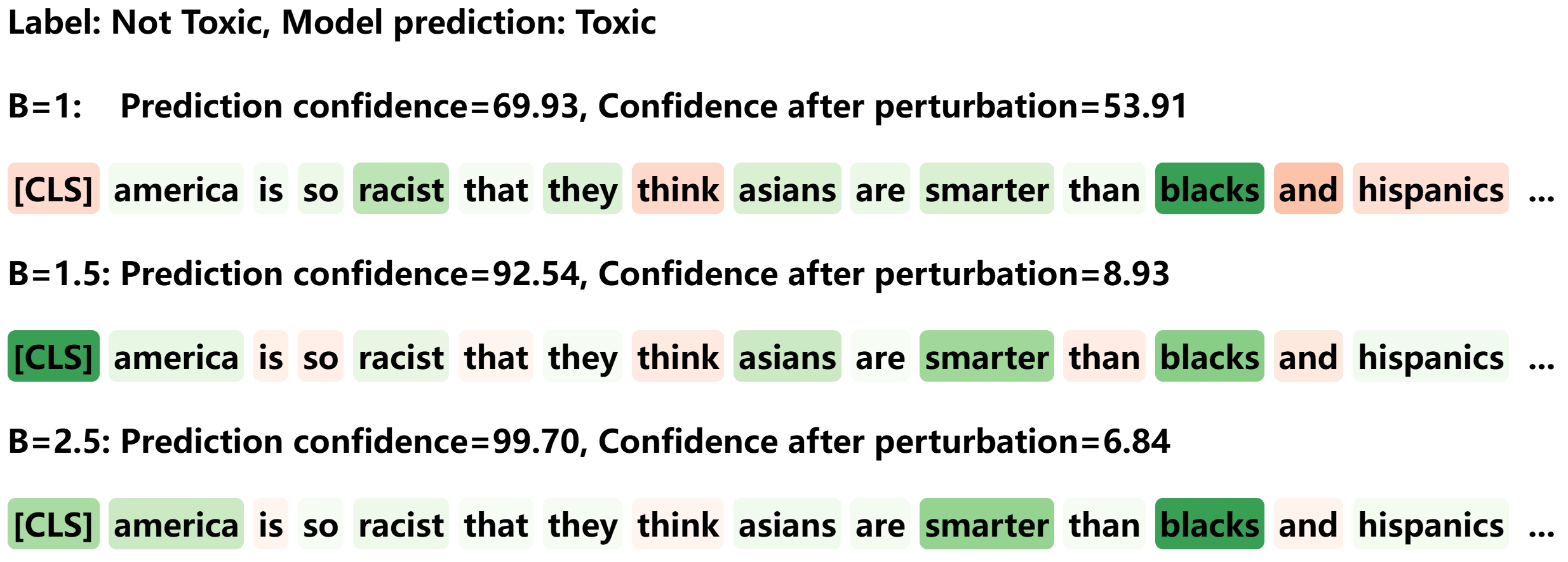}
    \caption{Example of how larger B values lead B-cos LMs to learn word-level spurious correlations. \textcolor{forestgreen}{Green} (\textcolor{red}{red}) indicates the \textcolor{forestgreen}{positive} (\textcolor{red}{negative}) impact of tokens on the prediction. Higher alignment pressure increases the reliance of B-cos LMs on spurious correlations in the data. In this example, perturbation involves changing ``blacks'' to ``whites''.}
    \label{fig:word bias}
\end{figure*}

In the example shown in Figure~\ref{fig:word bias}, models become increasingly confident in the incorrect prediction as B increases, with attributions primarily assigned to the word ``blacks''. Moreover, simply replacing ``blacks'' with ``whites'' results in a sharp drop in confidence, which demonstrates a growing reliance on spurious correlations with increased alignment pressure. The observation further confirms our findings in \S\ref{sec:B}.

\section{Decoder-Only Model Results}
\label{appendix:blimp}
Table~\ref{tab:blimp probability difference} presents the average probability gaps between target and foil predictions on every dataset from different vanilla and B-cos models. Table~\ref{tab:blimp gpt} and Table~\ref{tab:blimp Llama} contain MRR results on every dataset in BLiMP and IOI for GPT-2 and Llama-3.2 models, respectively.

\begin{table}[h]
    \centering
    \begin{tabular}{lcccc}
         \toprule
         \multirow{2}{*}{\textbf{Dataset}} & \multicolumn{4}{c}{\textbf{Probability Gap ($\uparrow$)}} \\
         \cmidrule{2-5}
         &\textbf{Vanilla GPT-2} & \textbf{B-cos GPT-2} & \textbf{Vanilla Llama-3.2} & \textbf{B-cos Llama-3.2}  \\
         \midrule
         aga & 0.0120 & 0.0170 & 0.0171 & 0.0196
               \\
          ana & 0.0152 & 0.0189 & 0.0140 &0.0170
                \\
        asp & 0.0007 &  0.0008 & 0.0016 &0.0009
                   \\
          dna & 0.0011 & 0.0011 & 0.0011 &0.0017
               \\
        dnai & 0.0021 & 0.0006 & 0.0011 &0.0013                   \\
          dnaa & 0.0014 & 0.0012 & 0.0012 &0.0017
          \\
          dnaai & 0.0091 & 0.0058 & 0.0077 &0.0072
          \\
          npi & 0.0015 & 0.0002 & 0.0002 &0.0001  \\
        darn & 0.0067 & 0.0078 & 0.0080 &0.0085     \\
          IOI & 0.3351 & 0.3265 & 0.4652 &0.5021
         \\
         \bottomrule 
    \end{tabular}
    \caption{Probability gaps between target and foil next token predictions from vanilla models and B-cos LMs on every dataset.}
    \label{tab:blimp probability difference}
\end{table}

\begin{table}[h]
    \centering
    \begin{tabular}{lccccccc}
         \toprule
         \textbf{Dataset} & \textbf{Random} &\textbf{Grad Norm} & \textbf{IxG} & \textbf{Occlusion} & \textbf{Logit} & \textbf{ALTI Logit} & \textbf{B-cos GPT-2} \\
         \midrule
          aga & 0.6875 & 0.7927 & 0.7910 & 0.7513
         & 0.827 & 0.964 &0.8764 \\
         ana & 0.7056 & 0.6753 & 0.7387 &0.5957
         &0.817& 0.976 &0.7532 \\
         asp & 0.3818 &  0.7512 & 0.4086 &0.4374
         &0.386& 0.499 &0.4939 \\
         dna & 0.4608 & 0.3629 & 0.3869 &0.9030
         &0.737& 0.646 &0.9308 \\
         dnai & 0.4626 & 0.4077 & 0.4317 &0.8395
         &0.711& 0.637 &0.8596 \\
         dnaa & 0.4103 & 0.2632 & 0.3214 &0.6557
         &0.951& 0.807 &0.7798 \\
         dnaai & 0.4074 & 0.2632 & 0.3392 &0.6167
         &0.9& 0.757 &0.7601 \\
         npi & 0.6121 & 0.7854 & 0.4948 &0.4775
         &0.445& 0.417 &0.4573 \\
         darn & 0.4888 & 0.6170 & 0.3627 &0.4247
         &0.802& 0.949 &0.8936 \\
         IOI & 0.2360 & 0.8599 & 0.1112 &0.8517
         &1.0& 1.0 & 1.0 \\
         \bottomrule 
    \end{tabular}
    \caption{MRR Alignment of different explanation methods on GPT-2 small predictions on every dataset. B-cos explanations are extracted from the B-cos GPT-2 model. Logit and ALTI Logit results are duplicated from~\citet{ferrando-etal-2023-explaining}.}
    \label{tab:blimp gpt}
\end{table}

\begin{table}[h]
    \centering
    \begin{tabular}{lccccc}
         \toprule
         \textbf{Dataset} & \textbf{Random} &\textbf{Grad Norm} & \textbf{IxG} & \textbf{Occlusion} & \textbf{B-cos Llama-3.2} \\
         \midrule
         aga & 0.6868 & 0.6030 & 0.5928 & 0.7811
          &0.8485 \\
         ana & 0.7037 & 0.5955 & 0.6432 &0.6072
          &0.8535 \\
         asp & 0.3842 &  0.7694 & 0.4537 &0.3670
          &0.6108 \\
         dna & 0.4615 & 0.4598 & 0.4898 &0.8352
          &0.6743 \\
         dnai & 0.4630 & 0.4542 & 0.5043 &0.7671
          &0.6652\\
         dnaa & 0.4112 & 0.4299 & 0.4750 &0.6115
          &0.6308 \\
         dnaai & 0.4075 & 0.4221 & 0.4498 &0.5758
          &0.5563 \\
         npi & 0.6123 & 0.6367 & 0.7062 &0.5154
          &0.6264 \\
         darn & 0.4884 & 0.5828 & 0.45787 &0.5210
          &0.8065 \\
         IOI & 0.2328 & 0.3637 & 0.1034 &0.4767
          & 0.9913 \\
         \bottomrule 
    \end{tabular}
    \caption{MRR Alignment of different explanation methods on Llama-3.2 predictions on every dataset. B-cos explanations are extracted from the B-cos Llama-3.2 model. As Llama models are not supported in~\citet{ferrando-etal-2023-explaining}, we do not include their results.}
    \label{tab:blimp Llama}
\end{table}
\end{document}